\documentclass[Afour,sageh,times]{sagej}
\usepackage{preamble}
\usepackage{moreverb,url}

\usepackage{subcaption}

\usepackage[colorlinks,bookmarksopen,bookmarksnumbered,citecolor=red,urlcolor=red]{hyperref}

\newcommand\BibTeX{{\rmfamily B\kern-.05em \textsc{i\kern-.025em b}\kern-.08em
T\kern-.1667em\lower.7ex\hbox{E}\kern-.125emX}}

\def\volumeyear{2016}

\begin{document}

\runninghead{Aoyama \textit{et~al.}}

\title{Beyond Pure Sampling:\\
Hybrid Optimization Mechanisms for Non-Convex Model Predictive Control}

\author{Yuichiro Aoyama\affilnum{1,2},
Minchan Jung\affilnum{1,3$\ast$}, 
Akash Ratheesh\affilnum{1 $\ast$},
and Evangelos A. Theodorou\affilnum{1}
}

\affiliation{\affilnum{1} School of 
Aerospace Engineering, Georgia Institute of Technology, Atlanta, GA, USA\\
\affilnum{2} Development Division, Komatsu Ltd., Tokyo, Japan\\
\affilnum{3} Department of Electrical and Computer Engineering, Inha University, Incheon, Republic of Korea \\
\affilnum{$\ast$}
Equally contributed. 
}

\corrauth{Yuichiro Aoyama, School of Aerospace Engineering, Georgia Institute of Technology, 270 Ferst Drive, Atlanta, GA, USA}

\email{yaoyama3@gatech.edu}

\begin{abstract}
This paper investigates the optimization mechanisms of non-convex Model Predictive Control (MPC) using the Maximum Entropy Differential Dynamic Programming (ME-DDP) framework. Navigating non-convex cost landscapes induced by 
nonlinear dynamics, multiple obstacles, etc. 
remains a fundamental challenge in robotics, where gradient-based methods frequently converge to suboptimal local minima. We demonstrate a dual-step optimization mechanism designed to overcome these traps. (1) an initial phase of using DDP to exploit the gradient of the cost landscape, followed by (2) disruption of the optimization via sampling from policies characterized by the inverse Hessian of the action-value function. We provide a rigorous analysis of this sampling mechanism of three ME-DDP variants: Unimodal Gaussian ME-DDP, Multimodal Gaussian ME-DDP, and Stein Variational DDP. Furthermore, with navigation tasks of four robotic systems under cluttered environments, we conduct extensive benchmarking of three variants of the ME-DDP, against deterministic DDP, and one of the most successful sampling-based schemes, Model Predictive Path Integral (MPPI) control with three policy parameterizations and update laws that correspond to those of ME-DDPs.

The results show that in low-dimensional systems where the cost landscapes are relatively simple and local information is sufficiently representative, our framework consistently outperforms MPPIs. In high-dimensional systems, MPPI can occasionally discover aggressive maneuvers that enable it to steer the systems faster than DDP-based methods, whereas our method maintains a higher, more stable success rate.

Finally, we validate the practical efficacy of the framework through hardware experiments with a quadrotor navigating a dense, non-convex obstacle field, confirming the robustness of the proposed framework for real-world deployment.

\end{abstract}


\maketitle

\section{Introduction}\label{sec:introduction}

Trajectory Optimization (TO) is a fundamental technique for enabling robotic and autonomous systems to navigate safely and efficiently \citep{Kuindersma2016OptAtlas, Tassa2014controllimited}. From an optimization perspective, TO in robotics is characterized by nonlinear dynamics, state and actuation constraints, and complex environmental geometries, such as cluttered environments with multiple obstacles. The interplay of these factors induces a non-convex cost landscape \citep{Schulman2014motion, deits2015computing}. A canonical example is found in navigation tasks in cluttered environments, where the presence of multiple local minima, i.e., different topological paths around obstacles, creates multiple solutions \citep{Osa2020MultiModal}. 

To make trajectory optimization computationally tractable, researchers have traditionally relied on local approximations of this landscape. Consequently, gradient-based families of algorithms, such as Differential Dynamic Programming (DDP) \citep{Jacobson1970ddp} and Sequential Quadratic Programming (SQP) for dynamical systems \citep{Wilson1963simplicialSQP, Gill2000SQPdynamical, Gill2002SNOPT}, have been widely used.  


DDP and its variants, such as the iterative Pontryagin Maximum Principle (iPMP) \citep{Pontryagin1964mathematical}, utilize a forward-backward sweep architecture to propagate value functions or costates, arising from quadratic or linear approximation of cost and dynamics. Although this structure offers scalability and speed, it relies on local information (gradient) and approximations around nominal trajectories. Similarly, SQP addresses TO by solving quadratic subproblems derived from quadratic approximation of the cost and linearized dynamics. 

Because both families rely strictly on local first- and second-order information, they are both vulnerable to being captured by sub-optimal solutions . In the context of robotic tasks, this may correspond to situations where the robot gets stuck at a dead end or finds an inefficient path even if it reaches the target. In addition, the local nature makes the algorithm sensitive to the initial condition and incremental \citep{ kalakrishnan2011stomp, Zucker2013HamiltonianChomp}.

On the opposite side of gradient-based approach, sampling-based Stochastic Optimization (SO) methods overcome the limitations of locality.
These algorithms, typically deployed in a Model Predictive Control (MPC) fashion, have been successfully used across diverse robotic systems. The most representative algorithms in this approach are Cross-Entropy (CE) and Model Predictive Path Integral control (MPPI) and its variations \citep{Williams2016MPPI, williams2017model, Williams2018MPPI, Williams-RSS-18, RobustMPPI, Wang2021Tsallis}.
The algorithms sample dynamics forward and with perturbed control and refine their policy by weighting the trajectories based on the cost associated with the samples.

Because these algorithms do not require derivatives of the cost and dynamics, they are easily applied to non-smooth systems. Furthermore, the stochastic sampling process enables broader exploration, allowing algorithms to explore multiple local solutions that gradient-based methods cannot. The drawback of MPPI is that the resulting optimal decisions are noisy, and this can result in undesirable stochastic behavior and slow convergence. 

Recent efforts to mitigate the negative effect of noise include sampling from colored noise \citep{Vlahov2024ColoredMPPI} to have temporal correlation for a smoother trajectory, and using Hessian of cost as sampling covariance for improving convergence \citep{yi2024covompc}. Furthermore, techniques such as spline parameterization of control can effectively reduce control jitter \citep{Miura2024SVMPPISpline, Haoru2025DialMPC}. Although they all alleviate the negative effect, 
often do so at the cost of expressive flexibility or by reintroducing differentiability requirements.

While the sampling aspect of SO provides a way to handle complex objective functions, this feature does not completely address the issue. Naive sampling can often be inefficient in exploring the state space. These limitations motivated the work on multimodality.
One common strategy involves a coarse global search followed by local refinement.
The method proposed by \cite{Osa2020MultiModal} utilizes a Gaussian Mixture Model (GMM) to coarsely fit multiple candidate solutions in a reduced basis-function space, followed by a gradient-based optimization to refine them. Although this approach works well in a robotic arm as presented,  it is fundamentally kinematic in nature. Because the first sampling phase is decoupled with the system dynamics, it would become a bottleneck for underactuated systems where the manifold of feasible trajectories is much more strict. 
\cite{Sundaralingam2023curobo} adapts this concepts for the GPU era by adding massive parallelization. It uses MPPI-like stage to seed a kinematic gradient-based optimizer.
\cite{10960717} runs MPPI to produce coarse but diverse  trajectory, then refines it with gradient based interior-point DDP solver.

Similar, but another way to combine the two optimization techniques, i.e., gradient and sampling, to handle multimodality is to perturb a single-seed gradient-based optimization via sampling. In \cite{Schulman2014motion}, SQP-based trajectory optimizer is perturbed by randomly sampled noise during optimization to find better local solutions. Authors in \cite{Zucker2013HamiltonianChomp} perturb gradient-based kinematic optimizer by sampling momentum from energy-like distribution induced by the cost of the trajectory to jump to a better local solution.

Beyond purely optimization-based methods, recent research has explored the use of generative models to provide multimodal priors for planning. For instance, \cite{urain2022learaningimplicitprior} utilize an energy-based model to represent multimodal distributions, which are then integrated into gradient-based and stochastic optimizers. Similarly, \cite{Carvalho2023Diffusionrobot} employ a diffusion model to capture complex trajectory priors from large-scale datasets.  Authors in \cite{huang2024diffusionseeder} leverage a generative prior to initialize the parallelized optimization mentioned earlier \citep{Sundaralingam2023curobo} in an informed way. While these methods excel at representing the multimodality present in historical data, they are fundamentally data-dependent.

\cite{le2023acceleratingOT} utilizes Optimal Transport to handle multimodality, they often rely on a simplified linear Gaussian transition to enforce smoothness. This formulation implicitly treats dynamics as a soft smoothness penalty rather than following the true nonlinear dynamics. While \cite{pan2024mbd} adopts the nomenclature of generative diffusion, its underlying mechanism is a multi-step adaptation of the CEM using a noise-annealing schedule.

Alternative algorithms  \cite{lambdert2021SVMPC,lambert2021entropyregularizedmotionplanning,power2024constSV} explore multimodality by making use of Stein Variational Gradient Descent (SVGD) \citep{Liu2016SVGD}. SVGD is a particle-based optimization method that actively maintains solution diversity by using a kernel-based repulsive force.
The work in  \cite{lambdert2021SVMPC, lambert2021entropyregularizedmotionplanning} can be seen as SV extensions of MPPI. 
The method is also applied in kinematic planners, such as \citep{Yin2025SVDiffusion}, and as a parallelized search algorithm \citep{Lee2024SVErgodic}. In \cite{power2024constSV, Lee2025SVGD_STAMP}, a constrained SV TO algorithm is proposed. However, the approach scales unfavorably with the dimensionality of control, state variables, time horizon, and the number of equality/inequality constraints. 
In \cite{Barcelos2024Pathsignature}, the authors proposed signature-based repulsion, which, unlike pointwise diversity metrics that often yield redundant temporal variations of a single path, ensures that exploration occurs over topologically distinct and geometrically meaningful trajectories.

Motivated by the complementary strengths of these methods, we propose a framework that couples sampling- and gradient-based optimization. Specifically, we investigate two mechanisms that combine sampling- and gradient-based optimization to address the challenges of MPC in non-convex landscapes. One of the primary objectives is to provide a more rigorous analytical understanding of how these hybrid structures prevent optimization from becoming trapped in sub-optimal local minima.

The first mechanism is a schedule of optimization steps consisting of several Newton/second-order steps followed by an iteration that relies on sampling. The motivation for this hybrid process is that while Newton steps can locally optimize the objective function, the sampling phase promotes the escape of the trajectory from poor local minima. In the context of TO for robotics, this mechanism was first introduced in Maximum Entropy DDP (ME-DDP) \citep{so2022maximum} with Shannon's entropy and then extended to the broader class of Tsallis entropy \citep{aoyama2024tsallis_me_ddp} in unconstrained settings. It is also interesting to note that this hybrid optimization mechanism was also proposed in the area of business administration and management science \cite{harford2016messy} as a mental model to enhance creativity and improve innovation. 

The second mechanism provides a further extension of ME-DDP through Stein Variational DDP (SV-DDP). Based on the Stein Variational Newton's Method (SVNM) \citep{detommaso2018steinNewton}, SV-DDP is a kernel-based extension of ME-DDP. In SV-DDP, the optimization process alternates between two phases as in ME-DDP.
Starting from different initial trajectories, the first phase consists of multiple DDP updates  performed in parallel fashion, resulting in optimized state and control trajectories. In the second phase, the control trajectories are updated with the rule based on SVNM to generate the new state trajectories. This phase facilitates exploration by pushing the trajectories apart using a kernel-based repulsion force.

A preliminary version of this work appeared in \cite{Aoyama2025SVDDP}. This journal version expands on the previous work by providing a unified mathematical derivation of the algorithm family, a rigorous interpretation of exploration via Hessian, extensive simulation-based comparisons, and hardware validation.

\section{Background}\label{sec:background}
In this section, we establish theoretical preliminaries for dynamic optimization and entropic regularization. 
We first formalize the entropic regularization of a dynamic optimization, followed by a review of DDP and ME-DDP. 
To align with standard conventions in information-theoretic literature, we denote the natural logarithm as $\ln$ throughout this section.

\subsection{Entropic-regularized Dynamic Optimization}
With the deterministic dynamics
$x_{t+1} = f(x_{t}, u_{t})$, with state $x\in \mathbb{R}^{n_{x}}$ and control $u\in \mathbb{R}^{n_{u}}$, we consider the following TO problem: 
\begin{align}\label{eq:ocp_cost}
\min_{X,U} J(X,U) &= \min_{X,U}  \sum_{t=1}^{T-1}l(x_{t},u_{t}) + \Phi(x_{T}),
\end{align}
subject to the dynamics. Here, we define state and control trajectories as $X=[x_{1}, \cdots, x_{T}]$ and $U=[u_{1}, \cdots, u_{T-1}]$.
The scalar-valued functions $l(\cdot,\cdot)$, $\Phi(\cdot)$, and $J(\cdot, \cdot)$ denote the running, terminal, and total cost of the problem, respectively.
There exist several second-order solvers that can solve the problem above. Typically, they solve Quadratic Programming (QP) with a quadratic approximation of the cost under constraints from dynamics. Two well-known classes of algorithms are DDP \citep{Jacobson1970ddp} and 
SQP \citep{Gill2000SQPdynamical}. 
With quadratic approximation of dynamics, DDP effectively splits the problem into a sequence of stage-wise subproblems and solves them efficiently. SQP solves a large QP over trajectories under linearized dynamics, although there exist methods to solve QP efficiently with LQR \cite{Rao1998IPMPC}. 
Both of these methods are successfully used in robotic applications. 
However, since they are relying on local information on the cost and dynamics, i.e., gradient and Hessian, they are vulnerable to being trapped at a poor local solution as mentioned in \nameref{sec:introduction}. 

To alleviate the issue, we consider a stochastic control policy $\Pi(U|X)$ and 
introduce expected entropy of the policy:
\begin{align*}
\notag
\mathcal{H}[\Pi] &= \mathbb{E}_{P(X)}[\mathbb{E}_{\Pi(U|X)}[-\ln\Pi(U|X)]] \\ &=- \mathbb{E}_{P(X,U)} [\ln \Pi(U|X)],
\end{align*}
where $P(X)$ is the marginal distribution of the state trajectory induced by the policy $\Pi$, and $P(X,U) = P(X)\Pi(U|X)$ is the resulting joint distribution of the state and control trajectories. Here, we use the term expected entropy because it represents the conditional entropy of the control trajectory, marginalized over the state trajectory distribution. We add it to the objective to promote exploration. Incorporating the dynamics and taking expectation over a trajectory, we have a new objective:
\begin{equation}\label{eq:entropy_reg_objective}
J_{\Pi}(x_{1}, \Pi)= \mathbb{E}_{P(X,U)} \Big[\sum_{t=1}^{T-1}l(x_{t},u_{t})  + \Phi(x_{T})\Big] - \tau \cH[\Pi].
\end{equation}
The temperature parameter $\tau$ acts as a thermodynamic scaling factor. The intuition here is that, at high temperatures, the control (thermodynamic particles) has high kinetic energy, spreading across the cost landscape to maximize entropy. This prevents the optimizer from collapsing into a local minimum. In contrast, when the temperature is low, the control acts like a crystal, which does not explore the landscape.

With the objective with the entropy, the TO problem seeks a stochastic policy that minimizes the original objective while maximizing the corresponding entropy. The resulting policy can explore multiple local solutions and is robust to being captured by poor ones. There exist multiple approaches to solve the problem. One of the simplest cases is with a feedforward control $\Pi(U|x_{1})$ which gives 
\begin{align}\label{eq:pi_star_general_entropic_dyn_opt}
\pi^{\ast}(U|x_{1}) =Z_{0}^{-1} \exp[-J(x_{1}, U)/\tau],  
\end{align}
where $Z_{0}$ is a partition function \cite{Yang2017SVpolicygrad}. In this work, we use DDP to effectively obtain and utilize the feedback policy.

\subsection{Differential Dynamic Programming}
In this section, we review DDP. Although this is a classic work, we highlight its inherent quadratic structure and treatment of its Hessian. This is because the Hessian plays a key role in the exploration strategies developed in the following sections.

We consider the cost-to-go at time step $t=i$, i.e., cost starting from $t=i$ to $N$,  denoted by ${J}_{i}(X_{i},U_{i})$. This is given by
\begin{equation*}
     {J}_{i}(X_{i},U_{i}) = \big[\sum_{t=i}^{T-1} l_{t}(x_t,u_t)\big] + \Phi(x_{T})
\end{equation*}
with trajectories starting from time step $i$ :  $X_{i} =[{x}_{i},\dots,{x}_{N}]$, $U_{i} =[{u}_{i},\dots,{u}_{N-1}]$. The value function is defined as the minimum cost-to-go in each state and time step $t$ via $V_t({x}_t):=\min_{{u}_t}J_{t}({X_{t}},{U_{t}})$.
Given Bellman's principle of optimality that provides the following rule
\begin{equation}\label{eq:bellman}
    V_t(x_t) =  \min_{{u}_t} [\underbrace{l({x}_t,{u}_t) + V_{t+1}({x}_{t+1})}_{Q_{t}(x_{t},u_{t})}], 
\end{equation}
where $Q_{t}(x_{t}, u_{t})$ is action-state, or simply $Q$ function,
DDP finds local solutions to the minimization of \eqref{eq:bellman} by expanding about nominal trajectories $\bar{X}$ and $\bar{U}$. 

The first step of the minimization is to perform quadratic expansions of $Q_t$ about nominal pair ($\bar{x}_{t}$, $\bar{u}_{t}$) with the deviation $\delta x_t=x_t -\bar{x}_{t}$, $\delta u_t=u_{t} - \bar{u}_{t}$, obtaining
\begin{align}\label{eq:Q_quad_expand}
 Q(x_{t},u_{t}) & \approx  Q(\bar{x}_{t}, \bar{u}_{t}) + Q_{u}^{\tr}\delta u_{t} + Q_{x}^{\tr}\delta x_{t} \\&+\frac{1}{2}\delta u_{t}^{\tr}Q_{uu}\delta u_{t} + 
\delta u^{\tr}Q_{ux}\delta x_{t}  \notag
 +\frac{1}{2}\delta x_{t}^{\tr}Q_{xx}\delta x_{t},
\end{align}
where we drop the time index $t$ for $Q$. For readability, we drop the time subscript $t$ where the time dependency can be recovered from the arguments or associated variables (e.g., writing $Q_{u}^{\tr} \delta u_{t}$ instead of $Q_{u,t}^{\tr} \delta u_{t}$), hereafter.

The quadratic approximation is a standard process in nonlinear optimization, as used in Newton's method \citep{Nocedal2006numerical}.

Assuming that the Hessian $Q_{uu}$ is Positive Definite (PD), we can explicitly optimize $Q$ approximated in \eqref{eq:Q_quad_expand} with respect to $\delta{u}_{t}$ by computing a partial derivative of \eqref{eq:Q_quad_expand} with respect to $\delta u_{t}$ and setting it zero. This minimization yields the following local optimal control law
\begin{align}\label{eq:delta-u-star}
    \delta {u}^{\ast}_t&={\kappa}_{t}+{K}_{t}\delta {x}_{t}, \\  \notag
    \text{with}\quad \kappa_{t}&= -{Q}^{-1}_{{uu}}{Q_{{u}}},\ {K}_{t} = -{Q}^{-1}_{{uu}}{Q_{{ux}}},
\end{align}
where $\kappa$ and $K$ are known as feedforward and feedback gains, respectively. 

DDP has backward and forward passes. In the backward pass, the derivatives of $Q$ functions which is later introduced in \eqref{eq:Qexpanded_derivative}, gains in \eqref{eq:delta-u-star} are computed backward in time. In the forward pass, the new control $\bar{u}_{t} + \delta u^{\ast}_{t}$ is propagated forward in time to give a new pair of nominal trajectories. 

\subsubsection{Backward pass}
To obtain the derivatives of $Q$ functions that are required to compute gains, we perform a similar expansion on the term $l(x_{t}, u_{t}) + V_{t+1}(x_{t+1})$ given in the definition of $Q$ function in  \eqref{eq:bellman}, and eliminate $\delta x_{t+1}$ using quadratic approximation of the dynamics:
\begin{align}\label{eq:linearized_dyn}
\delta x_{t+1}\approx  f_{x} \delta x_{t} + f_{u} \delta u_{t} + \frac{1}{2} 
\begin{bmatrix}
\delta x_{t} \\
\delta u_{t}
\end{bmatrix}^{\tr}
\begin{bmatrix}
f_{xx} & f_{xu}\\f_{ux}& f_{uu}
\end{bmatrix}
\begin{bmatrix}
\delta x_{t} \\
\delta u_{t}
\end{bmatrix},
\end{align}
where $f_x$ and $f_u$ denote the state and control Jacobians, while the block-matrix in the last term contains the Hessian tensors. Specifically, $f_{xx}$ and $f_{uu}$ represent the second-order sensitivities with respect to state and control, respectively, and $f_{xu}$ accounts for their coupling. Mapping the terms on both sides of quadratic approximation of $Q(x_{t}, u_{t}) = l(x_{t}, u_{t}) + V_{t+1}(x_{t+1})$, we obtain the derivatives of $Q$ (evaluated on $\bar{X}$ and $\bar{U}$) as follows. 
\begin{align}\label{eq:Qexpanded_derivative}
    {Q}_{{u},t}&={l}_{{u}}+{f_u}^{\tr}{V}_{{x},t+1}, \quad {Q}_{{x},t}={l}_{{x}}+{f_x}^{\tr}{V}_{{x},t+1}, \\\notag
    {Q}_{{uu},t} &={l}_{{uu}}+{f_u}^{\tr}{V}_{{xx},t+1}{f_u} + f_{uu} \cdot V_{x,t+1}, \\ \notag
    {Q}_{{xu},t}  &={l_{{xu}}}+{f_x}^{\tr}{{V}_{{xx},t+1}}{f_u} + f_{xu} \cdot V_{x,t+1},\\\notag 
    {Q}_{{xx},t}  &={l}_{{xx}}+{f_x}^{\tr}{V}_{{xx},t+1}{f_x} + f_{xx} \cdot V_{x,t+1},
\end{align}
where derivatives of the running cost and dynamics are evaluated at time $t$, and $\cdot$ for Hessians and ${V_{x,t+1}}$ is tensor contraction along the first (state) axis.

Now, $\delta {u}_{t}^{\ast}$ is computed using $V_{x,t+1}$ and $V_{xx,t+1}$, which are in one time step ahead. To propagate derivatives of $V_{t}(x)$ back in time, we consider the quadratic expansion of $V_{t}(x)$, that is, 
\begin{equation*}
V(x_{t}) = V(\bar{x}_{t}) + V_{x}(\bar{x}_{t})^{\tr}\delta x_{t} + \frac{1}{2}\delta x_{t}^{\tr}V_{xx}(\bar{x}_{t})\delta x_{t}, 
\end{equation*}
and equate the equation with the quadratic expansion of $Q$ in \eqref{eq:Q_quad_expand} through \eqref{eq:bellman}. Since we now have a solution of the $\min_{u_{t}}$ in the right-hand side of \eqref{eq:bellman}, by substituting $\delta u^{\ast}$ for $\delta u$, the $\min$ operator vanishes. This allows us to compare the coefficients of $\delta x_{t}$ by mapping the terms, giving the backward recursions: 
\begin{align}\label{eq:value_update_with_gain}
    V_{x,t} &= Q_{{x,t}} + {K_{t}}^{\tr}Q_{{uu,t}}{\kappa_{t}}
    +{K_{t}}^{\tr}Q_{{u,t}} + Q_{{ux,t}}^{\tr}{\kappa_{t}}, \\\notag
    V_{xx,t}&= Q_{{xx,t}} + {K_{t}}^{\tr}Q_{{uu,t}}{K_{t}}
    +{K_{t}}^{\tr}Q_{{ux,t}} + Q_{{ux,t}}^{\tr}{K_{t}},
\end{align}

with the terminal condition
\begin{equation}\label{eq:value_terminal_cnd}
V_{T}(x_{T}) = \Phi(x_{T}), \quad V_{x,T} = \Phi_{x}(x_{T}), \quad V_{xx,T} = \Phi_{xx}(x_{T}). 
\end{equation}

We note that in our implementation, we drop the second-order information of the dynamics and use a linear approximation, which corresponds to iterative LQR \citep{Li2004iLQR}.

\subsubsection{Forward pass}
In the forward pass, the new control $\bar{u}_{t} + \delta u^{\ast}_{t}$ is propagated forward in time to give a new pair of nominal trajectories, typically with a backtracking line search to absorb the discrepancy between the quadratic cost model, linear or quadratic dynamics model, and the actual ones.

\subsubsection{Regularization}

To compute the optimal gains by minimizing \eqref{eq:Q_quad_expand}, the Hessian $Q_{uu}$ must be PD. Furthermore, established convergence analysis relies on PD $Q_{uu}$ or on the convexity of the surrogate model formed with regularized $Q_{uu}$ \citep{liao1991convergence, liao1996global}. When the $Q_{uu}$ is not PD, it must be regularized. One of the well-known strategies is 
\begin{align}\label{eq:DDP_regularization}
    Q_{uu,t}^{\rm{reg}} = Q_{uu,t} + \mu_{\mathrm{reg}} I_{n_{u}},
\end{align}
which is equivalent to adding a cost penalizing large $\delta u_{t}$ via a quadratic trust-region penalty \citep{liao1991convergence}.

In practical implementations of DDP, if cost reduction is not achieved with a small step size in the forward pass, the backward pass is rerun with a larger regularization parameter \citep{TassaDDP2012}. 

This effectively restricts the optimization to a smaller, more reliable neighborhood of the nominal trajectory where the local approximation remains valid, and therefore, cost reduction is expected.

\subsection{Maximum Entropy Differential Dynamic Programming}
This section provides a review of ME-DDP and demonstrates its efficacy in solving the entropy-regularized dynamic optimization problem. We also introduce its unimodal and multimodal policies. We have a detailed derivation in the Appendix. 

Using the Markovian assumption \cite{puterman1994markov}, we decompose the policy \eqref{eq:pi_star_general_entropic_dyn_opt} defined over the trajectories into multiple of policies at each time step:  
\begin{align*}
\Pi(U|X) = \prod_{t=1}^{T-1} \pi(u_t|x_t).
\end{align*}
Then, the expected total entropy of the policy $\mathcal{H}[\Pi]$ becomes the sum of stage-wise entropy:
\begin{align*}
\mathcal{H}[\Pi] &= \mathbb{E}_{P(X,U)} \left[ -\ln \prod_{t=1}^{T-1} \pi(u_t|x_t) \right] \\ \notag
&= \sum_{t=1}^{T-1} \mathbb{E}_{P(x_t, u_t)} \left[ -\ln \pi(u_t|x_t) \right] \\ \notag
&= \sum_{t=1}^{T-1} \mathbb{E}_{P(x_t)} \left[ \mathbb{E}_{u_t \sim \pi(\cdot|x_t)} [ -\ln \pi(u_t|x_t) ] \right] \\ \notag
&= \sum_{t=1}^{T-1} \mathbb{E}_{P(x_t)} \left[ \mathcal{H}[\pi(\cdot|x_t)] \right],
\end{align*}
With this decomposition, we consider the stage-wise minimization of \eqref{eq:entropy_reg_objective} given state $x_{t}$,
\begin{align}\label{eq:max_entropy_with_cnst}
&\min_{\pi} \big\{\mathbb{E}_{u \sim \pi(\cdot|x)}[l(u_{t},x_{t}) + \regV_{t+1}(x_{t+1})] - \tau \cH [\pi(\cdot | x_{t})]\big\}, \\\notag
&\quad \quad\text{subject to} 
\int \pi(u_{t}|x_{t})\mathrm{d}u = 1,
\end{align}
which can be seen as an entropic regularized version of \eqref{eq:bellman}. Here, we use $\regV$ to denote the value function of the entropic regularized problem. The minimization in \eqref{eq:max_entropy_with_cnst} results in the optimal control policy 
$\pi_{t}^{\ast}$:
\begin{align} \label{eq:me_ddp_pi_star_uni}
\pi^{\ast}(u|x) &= {Z(x)}^{-1}\exp[ -\tilde{Q}(x,u)/{\tau}], \\
&\text{with}\ 
\tilde{V}(x) = -\tau \ln Z(x),
\end{align}
where  $Z(x)$ is the corresponding partition function and 
$\tilde{Q}_{t}(x_{t}, u_{t}) = l(x_{t}, u_{t}) + \tilde{V}_{t+1}(x_{t+1})$.
The relationship represents a smooth approximation of the Bellman optimality operator. The intuition here is that, by exponentiating the negative $Q$ function, the lowest value are magnified relative to higher ones. The subsequent operation with $\ln$ and negation returns the result to the original scale. This operation effectively acts as a Soft-min operator \citep{levine2018reinforcement, ziebart2010modeling}.

\subsubsection{Unimodal Gaussian Policy}
Combined with the quadratic approximation of $Q$ function \eqref{eq:Q_quad_expand} utilized in DDP,  \eqref{eq:me_ddp_pi_star_uni} indicates that the optimal policy has a form of unimodal Gaussian $\pi^{\ast}(\delta u|\delta x) \sim \mathcal{N}(\delta u^{\ast}, \tau Q_{uu}^{-1})$, where $\delta u^{\ast}$ is a solution of deterministic DDP in \eqref{eq:delta-u-star}.
\subsubsection{Multimodal Gaussian Policy}
In the multimodal case, we consider $N$ trajectories or modes:
$$
X^{(n)} = [x_
{1}^{(n)}, \dots, x_{T}^{(n)}], \quad 
U^{(n)} = [u_
{1}^{(n)}, \dots, u_{T-1}^{(n)}],
$$
with $n=1,\dots,N$ and the LogSumExp approximation of the value function 
$$\regV(x) = -\tau \ln \sum_{n=1}^N \exp(-\regV^{(n)}(x)/\tau),$$
where the superscript $(n)$ denotes the $n$ th trajectory. The exponential transformation $
\mathcal{E}_{\tau}(y) = \exp(
-y/{\tau})$ of the value function results in a control policy represented as a mixture of Gaussians whose categorical distribution is proportional to the value function of each trajectory. This multimodal policy is represented as follows: 
\begin{align}\label{eq:me_ddp_pi_star_multi}
 &\pi^{\ast}(u|x) 
= \sum_{n=1}^{N} \omega^{(n)}(x) \pi^{\ast (n)}(u|x)\\\notag
&\text{with} \quad \omega^{(n)}
\propto 
\mathcal{E}_{\tau}[
 \regV^{(n)}(x)], \quad \text{and} \quad \sum_{n=1}^{N}\omega^{(n)} =1. 
\end{align}
The intuition here is that the policy decides which modes to sample based on the value function and then samples from the corresponding Gaussian.
We note that while \eqref{eq:me_ddp_pi_star_multi} theoretically allows the policy to be multimodal, in practice these modes may collapse into a single dominant local minimum. This numerical collapse significantly diminishes the exploratory benefit of ME-DDP. 

To mitigate the issue, we introduce a heuristic in our implementation with a lower bound $\omega_{\mr{min}}$ on the weights. The weights are updated as $\hat{\omega}^{(n)} = \max(\omega^{(n)}, \omega_{\rm{min}})$ and subsequently renormalized.

\subsubsection{Exploration (sampling) vs Exploitation (gradient)}

In the ME-DDP framework, the entropic term facilitates exploration, while the deterministic DDP part governs exploitation. As the title suggests, excessive exploration can wash out the signal provided by the gradient. If the temperature $\tau$ is too high, the optimizer may not identify the underlying cost structure, leading to poor convergence. To mitigate this, the authors of \cite{so2022maximum} suggest that sampling should not happen at all iterations. Instead, perturbations are introduced periodically to allow the nominal trajectory to stabilize between exploratory phases. Although the original work was in the unconstrained setting, we observe that this is even more important when constraints are handled via DDP. Furthermore, following the spirit of \cite{Dong2021Replicaexchange}, we implement a heuristic that preserves the trajectory with the best metric without sampling. This approach ensures that the high-precision convergence properties of the deterministic DDP are maintained for the current mode.

\section{Stein Variational Dynamic Optimization}\label{sec:SV dyn_opt}
In this section, we introduce SV-DDP, a kernel extension of the ME-DDP framework. By incorporating a kernel-based repulsive force, the algorithm actively prevents particle collapse and preserves a diverse set of modes during the exploration process.

To provide the theoretical basis, we first review SVGD and its connection to functional gradient descent. We then discuss the SVNM method as a second-order extension. Subsequently, we apply these techniques to the DDP framework to derive SV-DDP. Finally, we discuss the essential components of the algorithm.

\subsection{Stein Variational Gradient Descent and Newton's Method}

SVGD minimizes the Kullback-Leibler (KL) divergence between a set of particles and a target distribution by performing functional gradient descent in a Reproducing Kernel Hilbert Space (RKHS). SVNM incorporates the Hessian to accelerate convergence and better capture the geometry of the underlying distribution. 

For comprehensive treatments, we refer the reader to \cite{Liu2016SVGD, Liu2017SteinGDFLOW} for SVGD and to \cite{detommaso2018steinNewton} for SVNM.
In this subsection, $x \in \mathbb{R}^{d}$ denotes a generic optimization variable rather than the system state, and $x_{n} \in \mathbb{R}^{d}$ represents a specific sampled point (particle).

\subsubsection{Stein Variational Gradient Descent}

Let $p$ on $\bR^{d}$ be a target distribution that we wish to approximate using a collection of samples. 
We specify the argument of $p(\cdot)$, when we evaluate it at a point $x_{n}$ as $p(x_{n})$.
With samples $\{x_{n}\}$ from a tractable reference distribution $q$ on $\bR^{d}$, SVGD iteratively computes a transport map $\cT:\mathbb{R}^{d} \rightarrow \mathbb{R}^{d}$ so that the transformed samples of $q$, i.e., $\{\cT(x_{n})\}$ can empirically approximate $p$. This map is obtained by solving the following optimization problem:
\begin{align}\label{eq:SVGD_problem}
\min_{\cT} \{ D_{\rm{KL}}( {q}^{l+1}\|p)\} &= \min_{\cT}\left\{\mathbb{E}_{\cT_{\#} {q}^{l}}\left[\ln\left(\frac{ \cT_{\#}{q}^{l}}{p}\right)\right]\right\},
\end{align}
where $\cT_{\#} q^{l} = q^{l+1}$ and $l$ stands for $l$-th iteration. Here, $D_{\rm{KL}}$ is KL divergence that measures the difference between the two distributions. To simplify this problem, SVGD considers the vector-valued Reproducing  Kernel Hilbert Space (RKHS) $\cF^{d} = \cF\times \cdots \times \cF$, where $\cF$ is a scalar-valued RKHS with kernel $k(x, x')$. This framework allows us to represent functions as weighted compositions of kernels centered at the samples. Furthermore, it also allows us to solve the optimization problem in functional space by solving the corresponding problem with the weights using standard optimization techniques. To solve \eqref{eq:SVGD_problem}, 
we define the map $\cT$ as a perturbation $\cP$ of the identity map $I$ in $\cF^{d}$ as
\begin{align*}
\cT^{l}(x) = I(x) + \cP(x), \quad \text{for} \quad \cP(x) \in \cF^{d}.
\end{align*}
With the map, we reformulate the objective of the problem \eqref{eq:SVGD_problem} as
\begin{align}
\hat{J}_{q^{l}}[\cP] =D_{\rm{KL}}\big((I + \cP)_{\#} q^{l} || p\big ).
\end{align}
To minimize the objective $\hat{J}_{q^{l}}[\mathcal{P}]$, we consider a perturbation $\mathcal{P}$ in the direction of the functional gradient. Specifically, we define the descent direction as the negative functional gradient of $\hat{J}_{q^{l}}$ evaluated at the identity map (represented by the zero perturbation $\mathbf{0}$):\begin{align*}\mathcal{T}^{l} = \mathcal{I} - \alpha \nabla_{\mathcal{P}} \hat{J}{q^{l}}[\mathcal{P}] \big|_{{\mathcal{P}=\mathbf{0}}}\end{align*}
where $\nabla_{\mathcal{P}} \hat{J}_{q^{l}}[\mathcal{P}]$ denotes the functional derivative of the objective with respect to $\mathcal{P}$. This formulation allows us to treat the transformation of the distribution as a steepest descent process in the space of maps. By selecting $\mathcal{P} = -\alpha \nabla_{\mathcal{P}} \hat{J}_{{q}^{l}}[\mathbf{0}]$, we shift the current distribution ${q}^l$ towards the target $p$ in the direction that yields the most rapid decrease in KL divergence. This construction provides a direct link between standard gradient descent on a point $x$ and the functional update of the entire distribution ${q}$.

The descent direction satisfies the following condition with $S\in\cF^{d}$:
\begin{align*}
D\hat{J}_{q}[S](\cV) = \langle \nabla \hat{J}_{q}[S], \cV \rangle_{\cF^{d}}, \quad \forall \cV \in \cF^{d}, 
\end{align*}
where we drop $l$. Here, the left-hand side is the first variation of $\hat{J}_{q}$ at $S$ along $V$ defined as follows:
\begin{equation*}
D\hat{J}_{q}[S](\cV) = \lim_{\zeta \rightarrow 0}(\hat{J}_{q}[S+\zeta \cV] - \hat{J}_{q}[S])/{\zeta}.
\end{equation*}
The authors of \cite{Liu2016SVGD} showed that the functional gradient at $\bm{0}$ is empirically approximated by $N$ particles:
\begin{align}\label{eq:SVGD_emp_func_gradient}
&-\nabla \hat{J}_{q}[\bm{0}](x_{i}) \\\notag =& \frac{1}{N}\sum_{n=1}^{N}
\Big[\big(\nabla_{x_{n}}\log p(x_{n})\big)k(x_{n}, x_{i}) + \nabla_{x_{n}}k(x_{n}, x_{i})\Big],
\end{align}
where the first term in the summation follows the gradient direction, and the second term spreads the particles apart from each other. Thus, it is known as the repulsive force. Finally, the update rule for samples is given by setting $\phi^{\ast} = -\nabla \hat{J}_{q}[\bm{0}]$ and with a step size $\alpha$ as follows:
\begin{equation*}
x_{n}^{l+1} \gets x_{n}^{l} + \alpha \phi^{\ast}(x_{n}^{l}).
\end{equation*}

\subsubsection{Stein Variational Newton Method}
SVGD constructs a vector field and evaluates it at a single point to determine how a particle moves.
SVNM additionally incorporates second-order information by approximating the Hessian of the KL functional, which is an operator rather than a vector.
Operators act on functions and therefore require two evaluation points. To reduce indices, we adopt the following notation: $z$ represents the point where the input function is evaluated (input location), and $y$ represents the location where the output of the operator is evaluated (output location).
Let us consider a function $f_{0}$ and an operator $H_{0}$, then applying the operator to the function gives
$$
(H_{0}f_{0})(y) = \int k(y,z) f_{0}(z) dz,
$$
where the kernel $k(y,z)$ defines the coupling of $z$ and $y$.

For second-order optimization in functional space, we define the second variation of $\hat{J}_{q}$ at $\bm{0}$ along the pair of directions $\cV, W \in \cF^{d}$ as:  
\begin{align*}
D^{2}\hat{J}_{q}[\bm{0}](\cV,W) = \lim_{\zeta \rightarrow 0} (D \hat{J}_{q}[\zeta W](\cV) - D\hat{J}_{q}[\bm{0}](\cV))/{\zeta}.
\end{align*}
The Newton direction $W$ is obtained by the optimality condition given by the following equation.
\begin{align*}
D^{2}\hat{J}_{q}[\bm{0}](\cV,W) =  - D \hat{J}_{q}[\bm{0}](\cV), \quad \forall \cV \in \cF^{d}, 
\end{align*}
which gives the transformation map of the Newton direction as a perturbation of the identity map as
$\cT = I + \alpha W.$
The authors in \cite{detommaso2018steinNewton} proved that the Newton direction $W = (w_{1}, \cdots, w_{d})^{\mathsf{T}}$ satisfies for all $\cV = (v_{1}, \cdots v_{d})^{\tr} \in \cF^{d}$,
\begin{align}\label{eq:SV_Newton_hessian}
\notag
&\sum_{i=1}^{d}\Bigl\langle\sum_{j=1}^{d} \langle h_{ij}(y,z),w_{j}(z) \rangle_{\cF} + \nabla \hat{J}[\bm{0}]_{i}(y), v_{i}(y) \Bigr\rangle_{\cF} = 0, 
\\\notag
&\text{with} \
h_{ij}(y,z) = \mathbb{E}_{x \sim q}\big[-[\nabla^{2}\log p(x)]_{i,j}k(x, y)k(x,z) \\ & \hspace{35mm}+ \nabla k{(x,y)}_{i}\nabla k(x,z)_{j}\big].
\end{align}
In the same work, the Galerkin approximation of the solution of $W$ is also proposed, where $W$ is expanded on $\cF^{d} = \mathrm{span}\{k(x_{1}, \cdot), \cdots, k(x_{N}, \cdot)\}$. The expansion leads to the following approximation with coefficients $\beta^{n}  \in \mathbb{R}^d$
\begin{align}\label{eq:SVNM_diag_direction}
w(z) = {\sum_{n=1}^{N}}\beta^{n}k(x_{n},z),
\end{align}
where the superscripts $n$ indicate the $n$-th particle. The coefficients $\beta$s are given as a solution of linear systems: 
\begin{equation*}
\sum_{n=1}^{N} H^{s,n} \beta ^{n} = -\nabla \hat{J}_{p}[\bm{0}](x_{s}), \quad \text{for} \ s = 1, \cdots ,N,
\end{equation*}
where the Hessian is denoted as $H^{s,n}_{i,j} = h_{ij}(x_{s}, x_{n})$. 

The authors also propose the block-diagonal approximation of the system for parallelization, by transforming the system as: 
\begin{align}\label{eq:SV_Newton_blkdiag_approx}
 H^{n,n} \beta^{n} = -\nabla \hat{J}_{p}[\bm{0}](x_{n}), \quad \text{for} \ n = 1, \cdots, N,
\end{align}
which means that the off-diagonal blocks are approximated to zero $H^{s,n} = O_{d}$ for $s \neq n$, where $O_{d} \in \mathbb{R}^{d\times d}$ is a zero matrix.
A detailed explanation with an example is provided in the appendix. 
SVNM repeats the process of solving the linear systems above and updating particles with 
\begin{align*}
x_{n}^{l+1} \gets x_{n}^{l} + \alpha w(x_{n}^{l}),
\end{align*}
to approximate the optimal distribution.
\subsection{ME-DDP with Functional Gradient and Hessian}
In this section, we apply SVNM to ME-DDP to derive a new algorithm SV-DDP. The motivation here is that, by using a kernel-based repulsive force, the algorithm can keep trajectories representing the modes diverse. This mechanism prevents mode collapse and maintains high exploration capability.

We consider $N$ trajectories optimized by DDP to compose policy as in the MG-ME-DDP. Here, we assume that the global $Q$ is approximated by $Q^{(n)}$ around the trajectories $[x^{(n)}, u^{(n)}]$. 

We initially considered a GMM to approximate $Q$. However, since each DDP trajectory is local in nature, treating the ensemble as a global PDF via GMM interpolation is physically inconsistent. Such an approach would require evaluating local policy in regions far from their nominal trajectories, where the underlying quadratic approximations are not valid.

By substituting the optimal policy $\pi^{\ast}(u)$ for $p$ in the functional gradient \eqref{eq:SVGD_emp_func_gradient}, we obtain
\begin{align*}
&-\nabla \hat{J}_{q}[\bm{0}]({u}^{(s)}) \\
=& \frac{1}{N}\sum_{n=1}^{N}
\Big[\nabla_{u^{(n)}}\Big( -\frac{Q({x}^{(n)},u^{(n)}) }{\tau} \Big)k( u^{(s)}, u^{(n)}) \\
&\hspace{40mm}+ \nabla_{u^{(n)}}k(u^{(n)}, u^{(s)})\Big],
\end{align*}
for $s = 1, \cdots,N$.
We consider the quadratic approximation of $Q$ function and deviation of trajectories. Due to the linearity of the deviation, derivative with respect to $u$ is now equivalent to that of $\delta u$. As in the case of ME-DDP, the optimizer alternates optimization and exploration.  Here, we assume that DDP refinement can bring the trajectory to a locally convex region with PD $Q_{uu}$, 
and consider the optimal control $u^{\ast}$ computed by DDP.  By substituting the optimal control back into the quadratic approximation of $Q$, we have 
\begin{align}\label{eq:SVDDP_grad}
&-\nabla \hat{J}_{q}[\bm{0}](u^{\ast(s)}) \\\notag
&= 
\frac{1}{N}\sum_{n=1}^{N}
\Big[-\frac{1}{\tau}\underbrace{(Q_{u}^{(n)}+ Q_{ux}^{(n)}\delta x^{(n)} +  Q_{uu}^{(n)} \delta u^{\ast(n)})}_{\approx 0}\\\notag & \hspace{20mm} \times  k(u^{\ast (s)}, u^{\ast (n)}) + \nabla_{u^{\ast( n)}}k(u^{\ast (n)}, u^{\ast (s)})\Big],
\end{align}
where the derivatives of $Q$ are evaluated at $\bar{u}^{(n)}$.
The coefficient of the kernel $k(\cdot, \cdot)$ is zero because of the optimality condition in the derivation of DDP. 
Thus, we are left with the repulsive force terms. This seems attractive for spreading trajectories. However, since the temperature $\tau$ is lost,  this formulation cannot capture the relative importance of the entropy in the objective.

To recover the temperature, we use SVNM. By substituting $\pi^{\ast}(u)$ for $p(x)$ in the approximated Hessian \eqref{eq:SV_Newton_blkdiag_approx} with the terms in \eqref{eq:SV_Newton_hessian}, and performing empirical approximation, the Hessian for SV Newton's method is obtained as: 
\begin{align}\label{eq:Hss_SVDDP}
H^{s,s} &= \frac{1}{N}\sum_{n=1}^{N}\Big[ \frac{Q^{(n)}_{uu}}{\tau} k (u^{ (n)},  u^{(s)})^{2} \\ \notag & \quad + \nabla_{ u^{(n)}} k(u^{(n)}, u^{(s)})\nabla_{u^{(n)}} k( u^{(n)},  u^{(s)})^{\tr}\Big],
\end{align}
where we drop optimality $\ast$ for readability. By plugging this into \eqref{eq:SV_Newton_blkdiag_approx} and solving the equations, we get the SV update (or sample) of control:
\begin{align}\label{eq:SV_ddp_update}
&u^{(s)} \gets u^{(s)} +\alpha w(u^{(s)}) + K^{(s)}\delta x^{(s)} \\\notag
&\text{with} \ w = \sum_{n=1}^{N} \beta^{(n)}k( u^{(n)},  u^{(s)}),
\end{align}
where $K$ is feedback gain of DDP \eqref{eq:delta-u-star}.

Consequently, the temperature $\tau$ is naturally recovered within the Hessian $H^{s,s}$. We therefore propose that this approach is the consistent formulation for incorporating SV methods into the DDP framework. In \eqref{eq:Hss_SVDDP}, the second term is an expected outer product of the gradient, which captures information on the variation in all directions like its Hessian \cite{Trivedi2014expectoutergradient}. Thus, $H^{s,s}$ can be seen as a sum of the Hessians of the objective and kernels. We analyze how the Hessian interacts with the repulsive force term in section \nameref{sec:invH as covariance}. 
We note that the SV framework is deterministic. In this method, stochasticity enters the system only via the random sampling used for initialization. Nevertheless, we conceptualize and refer to this behavior as an exploration mechanism, maintaining conceptual continuity with the ME-DDP framework.

\begin{figure*}[!h]
\centering
\includegraphics[width=\linewidth]{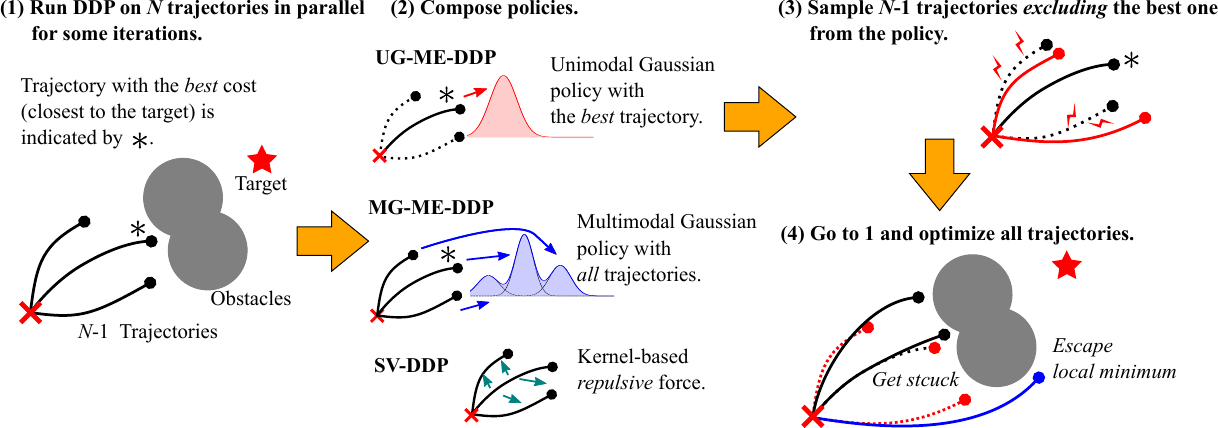}
\caption{Schematic of algorithms in a trajectory optimization setting. (1) The optimizer initializes $N$ trajectories, optimizing each via DDP for several iterations in parallel. The current best trajectory is marked ($*$). For illustrative clarity, we assume that the cost function is dominated by the distance to the target. 
(2) Policies are composed based on the specific algorithm: UG-ME-DDP uses a unimodal Gaussian centered on the best trajectory, MG-ME-DDP captures all trajectories in a multimodal distribution, and SV-DDP applies kernel-based repulsive forces to maintain diversity. Note that although the policies are constructed in control space, they are visualized here in state space for simplicity.  (3) New trajectories are sampled from these policies for exploration, excluding the current best. (4) The cycle repeats. Although the best trajectory may get stuck at a poor local solution, the sampling mechanism 
and trajectories ran in parallel  
allows trajectories to be re-initialized into more promising areas to reach the target. }
\label{fig:schematic_algs}
\end{figure*}

We provide the schematic of ME-DDP variants in Fig.\ref{fig:schematic_algs}. We note that the update is not applied to the best trajectory, ensuring that at least one mode shows a monotonic cost reduction to preserve the convergence property of DDP as in \cite{so2022maximum, Dong2021Replicaexchange}.
The proposed optimization framework, detailed in Algorithm \ref{alg:SVDDP_Main}, where an exploration mechanism based on the SV method is provided in \ref{alg:SV_Exploration}. We have alternative mechanisms, i.e., UG- and MG-ME-DDPs in appendix.

\begin{algorithm}[t!]
\SetKwInput{KwNotation}{Notation}
\DontPrintSemicolon
\SetAlgoLined
\SetCommentSty{textnormal}
\caption{Entropic Regularized DDP}\label{alg:SVDDP_Main}

\KwIn{
    $x_{1}$: Initial state, $\bar{u}$: Initial nominal sequence \\
    $\Sigma_{0}$: Initial covariance, $N$: Number of modes \\
    $m$: Sampling frequency, $I$: Max iterations \\
    $I_{\text{DDP}}$: DDP iterations per mode
}
\KwNotation{\\
    $\batch{U}$: Batched control trajectory $\{U^{(n)}\}_{n=1}^N$ \\
    $\batch{X}$: Batched state trajectory $\{X^{(n)}\}_{n=1}^N$\\
    $\batch{Q}_{uu}$: Batched Hessian sequences $\{Q_{uu, 1:T-1}^{(n)}\}_{n=1}^N$ \\
    $\batch{K}$: Batched feedback gain sequences $\{K_{1:T-1}^{(n)}\}_{n=1}^N$
}

\BlankLine
\tcp{Initialization}
$U^{(1)} \gets \bar{u}$\;
\For{$n = 2$ \KwTo $N$ \textbf{in parallel}}{
    $U^{(n)} \gets \{u_t^{(n)} \sim \mathcal{N}(\bar{u}_{t}, \Sigma_{0})\}_{t=1}^{T-1}$\;
    $J^{(n)} \gets \text{ComputeInitialCost}(X^{(n)}, U^{(n)})$\;
}
$b \gets \arg\min_{n} \{J^{(n)}\}_{n=1}^N$ \tcp*{Best mode index}

\BlankLine
\tcp{Optimization Loop}
\For{$i = 1$ \KwTo $I$}{
    \If{$i \pmod m = 0$}{
        \BlankLine
        $\{\boldsymbol{X}, \boldsymbol{U}\} \gets \textbf{Exploration}(\batch{U},\batch{X},  \batch{Q}_{uu}, \batch{K},b)$\;
        UG-ME-DDP: Algorithm \ref{alg:UG_Exploration}\\
        MG-ME-DDP: Algorithm \ref{alg:MG_Exploration}\\
        SV-DDP: Algorithm \ref{alg:SV_Exploration}
    }
    
    \BlankLine
    \tcp{Exploitation via Parallel DDP}
    \For{$n = 1$ \KwTo $N$ \textbf{in parallel}}{
        $X^{(n)}, U^{(n)}, K_{1:T-1}^{(n)}, Q_{uu,1:T-1}^{(n)}, J^{(n)} \gets \text{RunDDP}(X^{(n)}, U^{(n)}, I_{\text{DDP}})$\;
    }
    $b \gets \arg\min_{n} \{J^{(n)}\}_{n=1}^N$\;
}
\end{algorithm}

\begin{algorithm}[hbt!]
\DontPrintSemicolon
\SetAlgoLined
\SetCommentSty{textnormal} 
\caption{Stein Variational Exploration}\label{alg:SV_Exploration}
\KwIn{
  \batch{U},
  \batch{X},
  $\batch{Q}_{uu}$,
  \batch{K}, $b$
  }
\KwOut{Updated trajectories \batch{X}, \batch{U}} 
Compute Newton coefficients via \eqref{eq:SVDDP_grad} and\eqref{eq:Hss_SVDDP}

\For{$n = 1$ \KwTo $N$ \textbf{in parallel}}{
    \For{$t = 1$ \KwTo $T-1$ \textbf{in parallel}}{
        Solve $H_{t}^{n,n} \beta_{t}^{(n)} = -\nabla \hat{J}_{q}[\bm{0}]( u_{t}^{(n)})$
        $w_t^{(n)} \gets \sum_{n'=1}^{N} \beta_{t}^{(n')} k(u_{t}^{(n)},u_{t}^{(n')})$
    }
}
    \BlankLine
    \tcp{Apply update to all modes except the current best}
\For{$n = 1$ \KwTo $N, n \neq b$ \textbf{in parallel}}{
        $X^{(n)}, U^{(n)} \gets \text{LineSearch}(X^{(n)}, U^{(n)}, w_{1:T-1}^{(n)}, K_{1:T-1}^{(n)})$
    }
\Return $\{\batch{X}, \batch{U}\}$;
\end{algorithm}

\subsection{Essential Components in SV-DDP}\label{sec:components_SVDDP}
In this subsection, we first discuss the choice and properties of the kernel function, which serves as the core of the SV method and governs the behavior of the repulsive force. We then proceed with the selection of the step size for the SV update rule.

\subsubsection{Kernel and resampling}
\begin{figure}[h]
    \centering
    \includegraphics[width=\linewidth]{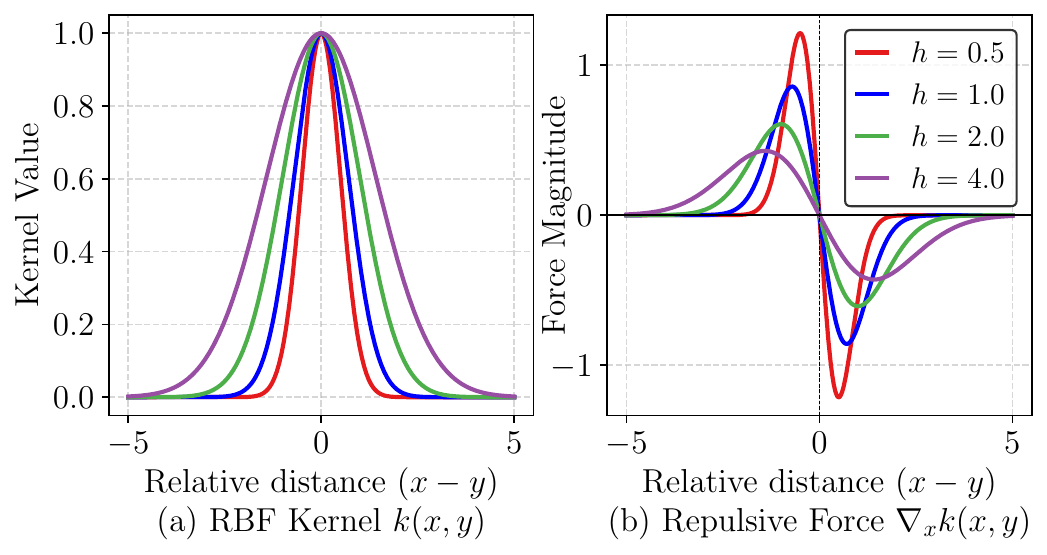}
    \caption{RBF kernel (a) and its gradient (b) across different bandwidths. The gradient represents the repulsive force of SV methods.}
    \label{fig:kernel}
\end{figure}

As in other SV literature in robotics, we use the RBF kernel $k(x, x') = \exp(-\|x - x'\|^2 / h)$ and choose its length parameter by median heuristic 
\begin{align*}
h = \frac{\text{median}(\{\|u^{(i)} - u^{(j)}\|^2\}_{i<j})}{\log N}
\end{align*}
\citep{garreau2018large_median_heuristic}. 
We note that while kernels such as the Inverse Quadratic (IQ) kernel are frequently used in static optimization and sampling to leverage heavy-tailed characteristics that induce repulsion at greater distances \citep{Gorham2017IQkernel}, the RBF kernel remains dominant in robotics. Fig. \ref{fig:kernel} shows the RBF kernel and its derivatives. We argue that in high-dimensional robotics problems, local repulsion of RBF is more effective than far-field interaction of IQ.

Although the exponential decay of the RBF kernel is a fundamental property, it poses a challenge in high-dimensional spaces where the concentration of measure increases the average Euclidean distance between particles. In such regimes, the kernel value $k(u^{(i)}, u^{(j)})$, and consequently its gradient, tends to vanish, as the exponential decay outpaces the linear growth of the distance vector. This leads to a vanishing repulsive force, making the SV method less effective \citep{zhuo2018messageSV}. The sum of local kernels alleviates this problem and is used in robotic applications \citep{Wang2021Tsallis, lambdert2021SVMPC}. This is because, in these works, the input of the kernel is a full horizon of control sequence whose dimension is $n_{u}(T-1)$.   In our work, due to the stage-wise formulation of DDP, the input dimension is only $n_{u}$.  Thus, it works well without the technique mentioned above.

The derivative of the kernel drives the repulsive force term. It reaches local extremum before decaying as the distance between points approaches zero. 
Consequently, while the force pushes trajectories apart at moderate distances, it vanishes when trajectories nearly coincide, potentially leading to mode collapse.
To counteract this issue, we monitor the distance of the trajectories normalized by the time horizon and dimension $\norm{u_{i}-u_{j}}/\sqrt{Tn_{u}}$. If the distance falls below a threshold, we keep the elite trajectory and resample the remaining redundant ones to ensure continuous exploration.

\subsubsection{Step size selection} Since the step size $\alpha$ determines the extent of the exploration, its choice is critical. While the SV literature often employs backtracking line search based on the Armijo condition \citep{Chen2019ProjectedSVNewton, Chen2020ProjectedSVGD}, robotic applications frequently tune task-specific fixed sizes \citep{Wang2021Tsallis, lambdert2021SVMPC}.

In our implementation, we observe that the repulsive force naturally pushes trajectories toward higher-cost regions, especially near obstacle boundaries. Consequently, a standard line search often prefers near zero $\alpha$. To address this, we implement a cost-bounded heuristic: we accept the largest $\alpha \in (0, 1]$ that keeps the trajectory cost within a factor $c_c$ (e.g., say 10-20) of the current minimum. This approach prioritizes global exploration while maintaining numerical stability.

\section{Geometric Analysis and Algorithmic Synthesis}
\label{sec:geom_analysis}
This section provides a formal justification for using the inverse Hessian as a covariance in sampling. Although utilizing $Q_{uu}^{-1}$, as a sampling covariance has been previously proposed in policy search literature \citep{Levine2016GuidedPolicy}, we present a more rigorous analysis to demonstrate why this choice is theoretically principled in the context of DDP. Subsequently, we provide how constraints are incorporated into our framework. Finally, we conclude with the implementation details for Model Predictive Control (MPC).

\subsection{Inverse Hessian as covariance}\label{sec:invH as covariance}
To understand the efficacy of the sampling scheme, i.e., the inverse of the Hessian as the covariance, we first examine a simple quadratic function and its Hessian to establish the mathematical relationship. Then, we extend the intuition to a more general static optimization problem. The static examples here are highly relevant because DDP can be seen as a sequence of local static sub-problems. Therefore, the insights gained here apply directly to ME-DDP. 
Using the same example, we analyze the effect of Hessian regularization on the sampling covariance. This analysis is essential for practical applications. Because the underlying problem is non-convex, the optimizer must robustly handle non-PD Hessians while maintaining meaningful exploration.
Finally, we discuss regularization itself.

\begin{figure}[H]
    \centering
    \includegraphics[width=\linewidth]{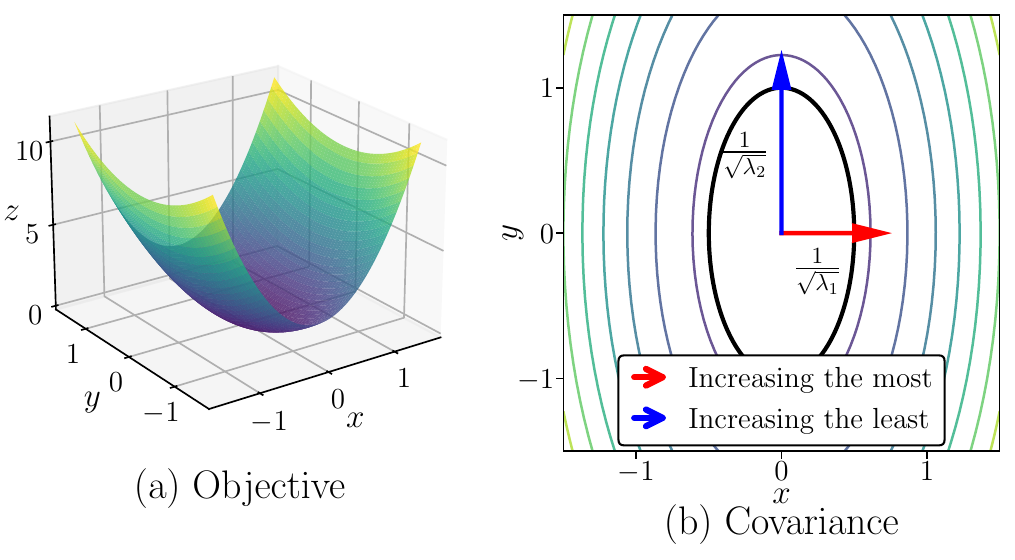}
    \caption{Surface plot of the quadratic function (a) and the resulting covariance ellipse overlaid on the contour plot (b). The blue and red arrows denote the principal directions of the covariance ellipse induced by the inverse of the Hessian $H$.}
    \label{fig:simple covariance}
\end{figure}

\subsubsection{Property of Hessian}
We consider a two-dimensional function with a diagonal Hessian:
\begin{equation*}
X^{\tr}HX, \ \text{with}\quad X = [x,y]^{\tr}\in \mathbb{R}^{2}, \ H = \mathrm{diag} \{ 4, 1\}.
\end{equation*}
The surface plot and the covariance ellipse induced by the inverse of the Hessian ($H^{-1}$) are shown in Fig. \ref{fig:simple covariance}. 
The Hessian has eigenvalues $\lambda_{1}=4$,\ $\lambda_{2} =1$ with corresponding eigenvectors $e_{1} =[1,0]^{\tr}, \ e_{2} =[0,1]^{\tr}$.
This means that the function changes the most rapidly along $e_{1}$ and the least rapidly along $e_{2}$.
In the contour plot, the levels are denser along the $x$-direction, reflecting the higher curvature in that dimension.

When the Hessian is inverted, the eigenvalues become the reciprocals of the original ones, while the eigenvectors are preserved. Consequently, the eigenpairs are $(1/\lambda_{1}=0.25, e_{1})$ and $(1/\lambda_{2}=1.0, e_{2})$. As illustrated by the covariance ellipse, using $H^{-1}$ as covariance, the sampling scheme samples more in directions where the function increases the least and less in directions where the objective increases the most, which is important when combined with optimization.

\subsubsection{Static problem}
In the previous example, we only analyzed the Hessian in a simple setting. Here, we consider the interaction between the local gradient and the geometry defined by the Hessian, and examine the sampling scheme in a more general setting.

We consider an optimization problem with an objective function $f_{0}(x)$ with $x \in \mathbb{R}^{n}$ and solve it with Newton's method by iteratively applying quadratic approximation and solving the subproblem. This process is fundamentally aligned with DDP. The subproblem with the approximation is given by 
\begin{align*} 
\min _{\delta x} \delta x^{\tr}\nabla f_{0}(\bar{x}) + \frac{1}{2} \delta x^{\tr}H \delta x,  \ \text{with} \ H = \nabla_{xx} {f}_{0}(\bar{x}), 
\end{align*}
where $\bar{x}$ is the current point, and we assume that the Hessian $H$ is PD.
Although a gradient-based step ($\delta x^{\ast}=H^{-1}\nabla f_{0}(\bar{x})$) reduces the cost of the quadratic model of the term $\delta x^{\tr}\nabla f_{0}(\bar{x})$, the PD Hessian serves as a local metric of cost sensitivity. It characterizes the sensitivity of the objective to deviations and how much step the optimizer can take. For example, when the Hessian has a large eigenvalue, the quadratic term ($\delta x^{\tr}H\delta {x}$) provides a large (positive) penalty that may counteract the descent term from the gradient.

To facilitate exploration, we consider the situation where we perturb the solution by $\delta x + \xi$, where $\xi \sim \mathcal{N}(0, H^{-1})$ once in a few iterations as we do in ME-DDP. 
As we examined in the previous example, the eigenvectors of the Hessian indicate the principal axes of curvature. When the inverse Hessian is used as the covariance matrix, the sampling scheme becomes anisotropic: it prioritizes exploration along the axis of minimal sensitivity, where the quadratic model is flat and therefore safe to explore. Furthermore, it restricts sampling along the axis of maximal sensitivity, where even small perturbations would drive the cost up, making exploration unsafe. Here, we use the term ``safety'' to refer to the preservation of the optimization signal. When exploration imposes too high a cost, it essentially washes out the information of the gradient, which makes the update pure random sampling. The choice of covariance allows the optimizer to exploit information from the gradient without being perturbed by overly costly samples. Therefore, the choice provides a rigorous framework for broad exploration while preserving the efficiency of the optimization via gradient.

Another interpretation can be obtained by investigating the change of cost induced by sampling. We consider the perturbation
$\delta x + \xi$, with noise $\xi \sim \mathcal{N}(0, \Sigma)$, where we specify the covariance $\Sigma$, later.  By substituting the perturbed $\delta x $ in the quadratic approximation and taking expectation of the stochastic components with $\xi$, we obtain:
\begin{align*}
\mathbb{E}\Big[\frac{1}{2}\xi^{\tr}H\xi + \xi^{\tr}(\nabla f(\bar{x})+H\delta x)\Big] & = \frac{1}{2}\mathbb{E}\big[\mathrm{Tr}(H\xi\xi^{\tr})\big]\\
&=\mathrm{Tr}(H\mathbb{E}[\xi\xi^{\tr}])/2\\
&= \mathrm{Tr}(H\Sigma)/2\\
&= \tau n/2,
\end{align*}
where we take the covariance as a scaled inverse Hessian $\Sigma=\tau H^{-1}$ and use the fact that the trace of a scalar is the scalar itself.
Generally, stochastic exploration imposes a penalty on the objective, particularly in regions of high curvature (large eigenvalue), where small perturbations can cause significant spikes. By this choice of covariance, we cancel the local geometry of the cost landscape. The optimizer explores more in the safe region and less in the unsafe region. This ensures that the expected cost of exploration remains constant and independent of the Hessian's eigenvalues, effectively normalizing the risk of sampling high-cost regions.

\begin{figure*}[t]
    \centering
    \includegraphics[width=\linewidth]{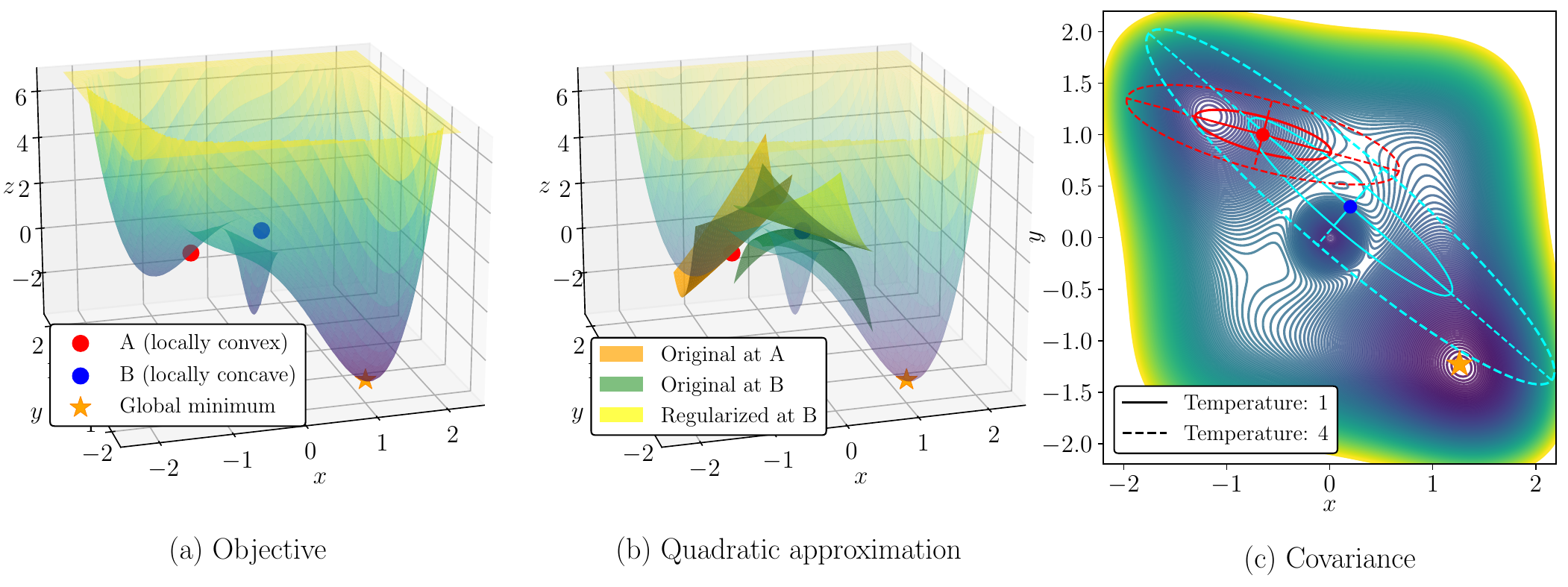}
    \caption{Surface plot of the objective function (a). Quadratic approximation of the objective at a locally convex point A and the approximation at a locally concave point B with original (not PD) and regularized (PD) Hessian (b). A contour plot of the objective function and the sampling covariance obtained at the two points A and B (c). The covariance ellipses, scaled by two temperatures, are drawn. 
    }
    \label{fig:sampling covariance}
\end{figure*}

\subsubsection{Regularization and covariance}
In the previous example, we assume that the Hessian is PD. In a practical setting, however, the Hessian may become indefinite or negative definite. To illustrate this, let us consider the two-dimensional example with three local minima whose surface plot is shown in Fig. \ref{fig:sampling covariance} (a).

As illustrated in Fig. \ref{fig:sampling covariance} (b), when the function is locally convex, the quadratic approximation for Newton's method is obtained with the original Hessian. Observe that the approximation yields a convex surface. In the same figure, we also perform a quadratic approximation at a locally concave point, where the Hessian is negative definite. At this point, we regularize the Hessian to obtain a valid quadratic model for minimization. Because regularization preserves the original eigenvectors and shifts the eigenvalues, it maps the most negative eigenvalues of the original Hessian to the smallest positive eigenvalues of the regularized PD surrogate. Consequently, the resulting inverse covariance prioritizes exploration along the directions of the steepest descent. This ensures that sampling is performed more extensively along the primary axes of the manifold, facilitating escape from saddle points, local maxima, and landscapes similar to them. 

Fig. \ref{fig:sampling covariance} (c) illustrates the covariance ellipsoids induced by the approximations in Fig. \ref{fig:sampling covariance} (b). To demonstrate the effect of temperature as a scaling factor, covariances at two different temperatures are projected onto the objective's contour plot. The higher temperature corresponds to broader exploration, and therefore, the corresponding covariance is large.

\subsubsection{Regularization with PD Hessian}
In \nameref{sec:background} section, we presented the importance of the regularization in DDP, however, with the first-order approximation of the dynamics and a proper choice of the cost function, i.e., PD Hessian with respect to state and control, $Q_{uu}$ is theoretically guaranteed to be PD (see Appendix for detailed analysis).

Even in such cases, regularization is still triggered to achieve cost reduction or to make $Q_{uu}$ well-conditioned for computing gains \citep{TassaDDP2012}. In these cases, since $Q_{uu}$ is originally PD, covariance samples within a tighter region around the nominal trajectory effectively serve as an implicit trust-region mechanism. This mechanism ensures that exploration remains within the vicinity where the approximations are valid.

\subsubsection{Inverse Hessian and Repulsive Force}
One can observe that the underlying principle of how the Hessian influences the repulsive force in SVNM and, by extension, SV-DDP is the same as the inverse Hessian acting as a covariance matrix. 

When solving \eqref{eq:SV_Newton_blkdiag_approx}, the repulsive force term is scaled by the inverse of the sum of Hessians of the $Q$ function and that of the kernel. Consequently, the repulsive force is amplified in directions where the joint curvature is small. This ensures that the rescaling does not impede the primary optimization objective, nor does it obstruct the spreading of trajectories. Recalling that the kernel value is highest when trajectories are in close proximity, this scaling mechanism steers the particles into low-sensitivity subspaces, facilitating efficient exploration.

\subsection{Constrained handing via Relaxed log- barrier Function}
This subsection details the integration of constraints into the ME-DDP framework (DDP and sampling). We begin by reviewing the standard $\log$-barrier approach and its application in optimal control, followed by the introduction of the relaxed log-barrier formulation. Finally, we discuss the limitations of the method.

One common approach to incorporating constraints into an objective function is through the $\log$-barrier function \citep{fiacco1968nonlinearbarrier}. These functions are powerful tools for static optimization and have been widely employed \cite{MurrayLinesearch1994Barrier, CHO2005BarrierStoch, ONeil2020NewtonCGBarrier}. The function is also used in the field of optimal control, specifically, in trajectory optimization \citep{BrysonHo1975AppliedOptCtrl, Betts2001OptCtrl} and MPC \citep{WILLS2004BarrierMPC} under constraints.

A key limitation of standard log-barrier methods is that they cannot handle infeasible trajectories due to the domain restrictions of the $\log$ function. This is problematic in our setting because the exploration is only implicitly aware of constraints. To address this issue, we utilize relaxed-$\log$ barrier function \citep{Hauser2006CDCbarrierfunc}.  
We define the scalar relaxed barrier $\mathcal{B}_{\mr{r},i}(g_i; \mu, \delta_{\mu})$ for a single constraint $g_i(x)<0$ as
\begin{align}\label{eq:def_relaxed_log}
&\mathcal{B}_{\mr{r},i}(g_i(x); \mu) \\ \notag
=& 
\begin{cases} 
-\mu \log (-g_i(x)), & g_i(x) \leq -\delta_{\mu} \\
\mu\! \left[ \dfrac{1}{2} \! \left( \!\left( \dfrac{g_i(x)+2\delta_{\mu}}{\delta_{\mu}} \right)^{2} \!\!\!- 1\! \right) \!- \log{\delta_{\mu}} \right]\!, \!\!& g_i(x) > -\delta_{\mu}.
\end{cases}
\end{align}
When $g(x) \in \bR^{w}$, the total penalty is then given by the sum $\mathcal{B}_{\mr{r}}(g(x)) = \sum_{i=1}^{w} \mathcal{B}_{\mr{r},i}(g_i(x))$. In this formulation, the function smoothly transitions from a logarithmic form to a polynomial approximation while maintaining 
$C^2$ continuity, allowing for infeasible trajectories. 

The relaxed formulation has found applications in MPC and robotics. For instance, \cite{Feller2017BarrierLinSys, Feller2017BarrierLinSysStable} applies it to a linear MPC problem and provides a detailed performance analysis. In \cite{Aguiar2017PRONTO}, the function is incorporated into a projection-based motion planning algorithm \cite{HAUSER2002prototypePRONTO}, achieving constrained trajectory optimization. The efficacy of the formulation with DDP/LQR is validated through hardware experiments on a quadruped in \cite{Grandia2019BarrierDDP}. Since it can handle infeasible trajectories, the formulation can be used that requires exploration that may cause constraint violation, such as reinforcement learning as proposed in \cite{Zhang2025constrainedBarrierRL}.

\subsubsection{Properties of the function}

\begin{figure}[H]
    \centering
    \includegraphics[width=\linewidth]{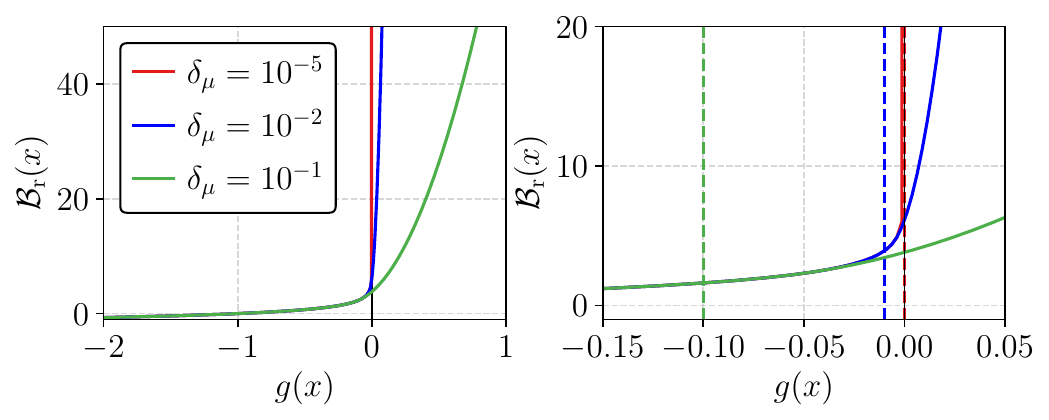}
    \caption{Relaxed log-barrier function $\mathcal{B}_{\mr{r}}(x)$ for various switching points $\delta_{\mu}$. The right panel provides a magnified view near the origin to highlight the behavior around the boundary ($g(x)=0$). Dashed vertical lines indicate the respective switching points $\delta_{\mu}$ where the barrier transitions from a logarithmic to a quadratic form.}
\label{fig:barrier_diffetent_deltas}
\end{figure}
Fig. \ref{fig:barrier_diffetent_deltas} provides the function with different relaxation parameters $\delta_{\mu}$. The parameter governs how closely the relaxed barrier approximates the exact logarithmic barrier and, consequently, the degree of constraint satisfaction. As $\delta_{\mu} \to 0$, the relaxed formulation converges to the exact $\log$-barrier, yielding increasingly strict constraint satisfaction. However, this increased accuracy comes at the cost of numerical conditioning: steep gradients and large curvature arise near the constraint boundary, leading to ill-conditioned gradient and Hessian, which is critical in second-order optimization methods \citep{Sleiman2021ETHDDPstyle}. Consequently, excessively small values of $\delta_{\mu}$ recover the numerical instabilities of the exact $\log$ barrier. In practice, $\delta_{\mu}$ is chosen to balance constraint satisfaction and numerical robustness \citep{Feller2017BarrierLinSys}. We note that even with a small $\delta_{\mu}$, the resulting conditioning issues are mitigated by appropriate scaling and regularization of gradient and Hessian terms \citep{Nocedal2006numerical, Wachter2006IPOPT}.

In the original formulation for static problems, the penalty parameter $\mu$ is reduced towards zero as optimization proceeds. In practice, however, a small fixed $\mu$ often provides sufficient accuracy while improving numerical stability \citep{Grandia2019BarrierDDP}. We present how the barrier function is incorporated into DDP in appendix.


\subsubsection{Limitations}
Numerical instability can be mitigated by the method described above, but numerical sensitivity, i.e., the amplification of small differences near the constraint boundary and hard branching, cannot be fully removed. We provide a detailed analysis in the appendix. We consider this to be an inherent limitation of the barrier approach compared with other constrained optimization methods, such as the Augmented Lagrangian (AL) method \citep{Hestenes1969MultiplierAG,powell1969method, rockafellar1973multiplier}, which incorporates Lagrangian multipliers for constraint satisfaction. The use of multipliers enables the algorithm to satisfy constraints and optimality conditions without requiring excessively large penalty parameters, which are a known source of numerical ill-conditioning and instability \citep{Nocedal2006numerical}. 
Nevertheless, we employ the barrier function because it can rapidly divert trajectories away from infeasible regions during the early stages of optimization.

\subsection{Receding Horizon Implementation}
To meet real-time requirements in highly nonlinear environments, as we will see in our experiments, we adopt an MPC formulation. A critical observation in our implementation is the necessity of consecutive optimization iterations for constraint satisfaction. Because the exploration process is not explicitly aware of the constraints, raw samples often violate constraints. We observe that overly frequent sampling is counterproductive, as it disrupts the optimizer's ability to recover feasibility. By restricting exploration to the initiation of the MPC cycle and prioritizing successive DDP iterations, we ensure the control sequence is both diverse and respects constraints.

A challenge in warm-starting is the coupling between the sampling covariance $\Sigma (Q_{uu}^{-1})$ and the nominal trajectories. While the control sequence $U^{(n)}$ can be shifted for warm-starting, the state sequence $X^{(n)}$ obtained by shifting inevitably diverges from those obtained by applying $U^{(n)}$ to the system with initial state $x_{t}$. Because feedback gains $K$ and sampling covariances $\Sigma$ are local approximations, this mismatch in state trajectory yields the simple temporal shift of feedback gain and covariance invalid. Consequently, a re-evaluation of the system's sensitivity via a backward pass is necessary at each time step to ensure the exploration remains aligned with the current local geometry.

We perform an initial backward pass at the start of each MPC loop to generate feedback gain and covariance. When the backward pass fails due to an ill-conditioned $Q_{uu}$, the algorithm samples with a pre-specified fixed covariance used in the initialization.

\section{Experiments}
In this section, we provide two experimental results.

First, we conduct extensive simulations across four robotic systems, comparing seven algorithms, including DDP, ME-DDP variants, and MPPI variants. Second, we present a hardware experiment with a quadrotor to validate the performance and real-time feasibility of the ME-DDP variants.

\subsection{Numerical Experiments}
The purpose of the experiment is to observe the properties of gradient-based (DDP), sampling-based (MPPI), and hybrid (ME-DDP) algorithms. 
We have comparison in performance and computational efficiency.

\subsubsection{Dynamics and Tasks}
\begin{figure}[H]
    \centering
    \includegraphics[width=1.0\linewidth]{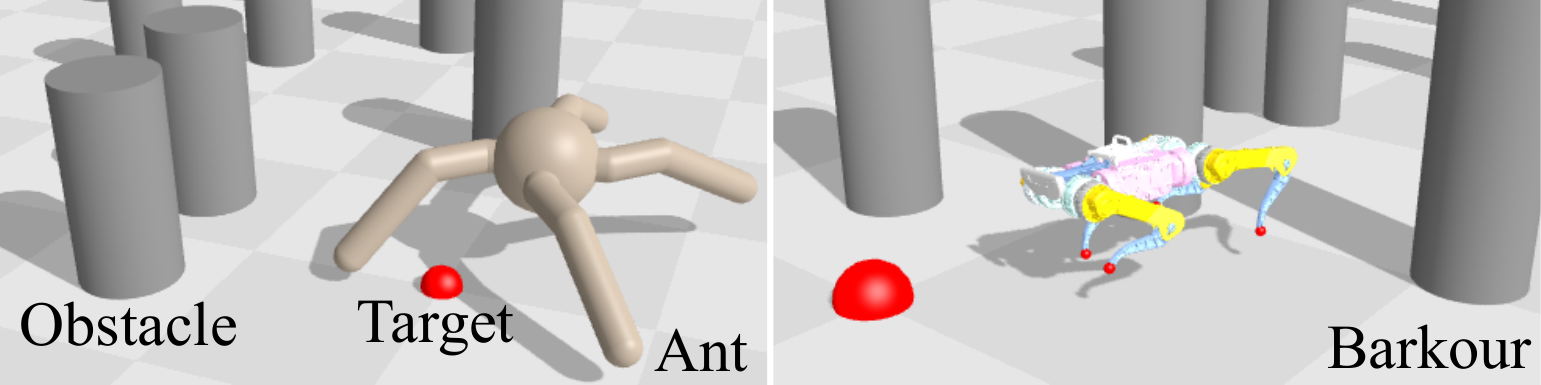}
    \caption{Experimental overview of Ant and Barkour. The task here is to reach the target while avoiding cylindrical obstacles. }
\label{fig:ant_barkour}
\end{figure}

We tested four dynamics: 2D car, quadrotor, Ant, and Barkour (quadruped) \citep{google2023barkour}. The Ant and Barkour are simulated in the Jax-based physics engine Brax \citep{brax2021github}. 
The task for each system is to reach targets while navigating obstacle fields. We use circular (2D) or cylindrical (3D) obstacles, generated in a pseudo-random fashion. To ensure rigorous testing, we manually introduce additional obstacles in certain configurations to create challenging non-convex landscapes, including local-minima traps. Here, we provide brief descriptions of the systems. Ant and Barkour are shown in Fig. \ref{fig:ant_barkour}.

\paragraph{2D car:}
The state $x\in \bR^{3}$ comprises 2D positions and heading angle. The control $u\in \bR^{2}$ consists of transitional and angular velocities. The discretization interval is 0.02 s. 
To investigate the effect of the MPC look-ahead horizon, we tested short (50 steps) and long (70 steps) horizons.

\paragraph{Quadrotor:}
The state $x\in\bR^{12}$ includes 3D positions, Euler angles, and their time derivatives.  The control $u\in\bR^{4}$ is the thrust force generated by four rotors. We follow the dynamics established in \cite{Luukkonen2011}. The discretization interval is 0.01 s. We have short (50 steps) and long (70 steps) horizons as in the case of the 2D car. 

\paragraph{Ant:}
The Ant has four legs, with two joints. The state $x\in \mathbb{R}^{29}$ contains the 3D position and orientation of the body in the quaternion, the angles of the legs, and their time derivatives. The control $u \in \mathbb{R}^{8}$ is torque applied to the joints. The control loop runs at  0.01 s with a simulation step of 0.002 s, integrating 5 substeps per control cycle.

\paragraph{Barkour:}
The quadruped has state $x \in \mathbb{R}^{37}$ comprising the 3D position, the body quaternion, and the three joint angles for each of the four legs, including the abductors, hips, and knees. The control $u \in \mathbb{R}^{12}$ is the command for the joint angles of the legs relative to the default angles that correspond to the standing posture. The control loop runs at  0.02 s with a simulation step of 0.006 s, integrating 3 substeps per control cycle.

We note that despite the higher dimensionality of the quadruped compared to the Ant, the use of a position-based (or angle-based) control interface provides a more structured search space, especially for MPPI. Position-level actions effectively abstract away the high-frequency dynamics and stability issues of torque-level control. This allows MPPI to discover task-relevant trajectories more efficiently. Typically, with torque-based control, many trajectories are wasted because random torque cannot even let robots stand stably. With position-based control, the optimizer can search for candidate trajectories within quasi-static equilibrium poses; therefore, the search space is significantly reduced to a region of low-cost configurations.

\subsubsection{Algorithms}
We compare seven algorithms across three families: normal DDP, three ME-DDP variants, and three MPPI variants. For the Ant and Barkour tasks, the DDP-based algorithms are modified to operate on the manifold $S^3$ to handle quaternions. We project quaternion deviations into the tangent space. This ensures that the quadratic approximations remain valid while the unit-norm constraint is maintained via retraction back to the manifold \citep{Boutselis2021DDPLiegroup}.

For MPPIs, we have three different policy parameterizations and update rules. They are Unimodal Gaussian MPPI (UG-MPPI), Multimodal Gaussian MPPI (MG-MPPI), and UG policy with SV update (SV-MPPI) \citep{Wang2021Tsallis}.
Here, the policy parameterizations correspond to UG-, MG-ME-DDP, and SV-DDP.
The experiment is performed in MPC fashion.

A key distinction between these DDP and MPPI families lies in their mechanism for constraint satisfaction. 
Due to the requirement of differentiability, DDP-based methods incorporate constraints into the cost function using barrier functions explained in section \nameref{sec:components_SVDDP}. 
While this allows for gradient-based optimization, it can permit the algorithm to violate constraints. In contrast, MPPI utilizes indicator functions with a prohibitive cost to encode constraints. The control update in MPPI is a weighted average of the sampled control trajectories based on their associated costs. Consequently, infeasible trajectories are assigned negligible weights. This mechanism effectively excludes unsafe trajectories from the control computation. Although it does not theoretically guarantee constraint satisfaction, our experiments show that it works well in practice to maintain feasibility given that MPPIs find a path to the targets.

\subsubsection{Parameter Tuning and Optimization}\label{subsec:param_tune_opt}

The performance of both ME-DDP and MPPI is affected by hyperparameter selection, such as the sampling covariance in MPPI or the temperature parameter $\tau$ in ME-DDP. Although the sampling covariance in the ME-DDP variants arises naturally from the inverse Hessian of the $Q$ function, the temperature must still be tuned to balance exploration and exploitation. To ensure a fair comparison, we acknowledge that different algorithms may require different cost weightings to achieve peak performance, even when the underlying structure (e.g., quadratic in position error to goal) is shared.

Consequently, we utilize a Bayesian hyperparameter optimization framework, specifically the Tree-structured Parzen Estimator (TPE), to tune both the policy parameters and the cost weights for every algorithm using an open-source Neural Network Intelligence (NNI) toolkit \citep{nni2021}.

For the tuning process, we define a high-level objective that prioritizes early entry into the target region and remaining within it for a prescribed duration, while penalizing the failure to reach the target within the specified time steps.

We use a fixed number of modes/trajectories $N=8$ in ME-DDP variants. The samples/trajectories of MPPIs vary with the complexity of the system. For multimodal policies of MPPI, i.e., MG- and SV-MPPI, we use the same number of modes $N=8$ and distribute an equal number of particles per mode.

\paragraph{2D Car and Quadrotor:} We utilize two sparse and two dense obstacle environments for tuning. This diversity is necessary because algorithms tuned solely on simple environments tend to converge toward low temperatures or narrow sampling covariances, which fail to generalize to complex, non-convex landscapes. The number of samples used in MPPI is 2048 for the 2D car and 8000 for the quadrotor.

\paragraph{Ant:} Due to the higher computational cost of these high-dimensional systems, we utilize a single representative environment for the tuning process. The number of samples in MPPI is 8192. 

\paragraph{Barkour:}
As in the case of Ant, we use one environment for tuning, and the number of samples in MPPI is 8192.

\begin{figure}[!ht]
    \centering
    \includegraphics[width=0.9\linewidth]{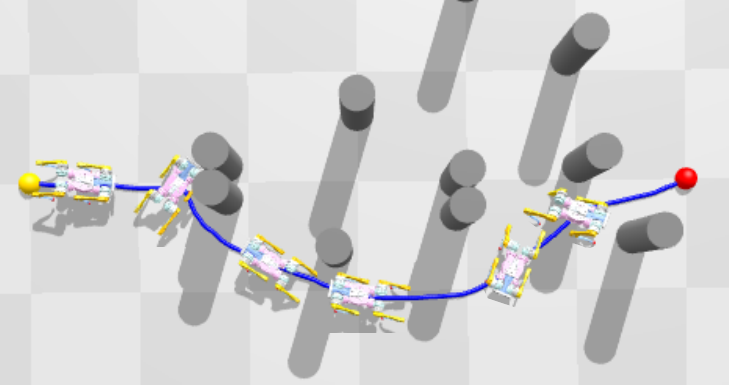}
    \caption{Impact of independent coordinate tuning on Barkour. The absence of yaw-position coupling leads to excessive angular excursions and rotational instability, despite the optimizer, UG-MPPI, successfully hitting the target. }
\label{fig:barkour_mppi_rotate}
\end{figure}

A primary challenge in tuning the cost function for the quadruped Barkour is to balance its performance with the requirement for gait realism. When the yaw rotation has less penalty compared to the position, the quadruped rotates while avoiding obstacles and moving forward. Although it can successfully reach the target without colliding, we prefer a realistic gait to spinning as provided in Fig. \ref{fig:barkour_mppi_rotate}. 

To mitigate the issue, we introduce coupled weights of position and yaw during the parameter search. This coupling ensures that when the tuner commands a high penalty on the position, the angle is also penalized to prevent spinning.  

In our experiments, we observed a divergence in how optimizers responded to this coupling. While the coupling significantly improved the qualitative trajectory quality (realistic gait) of the DDP-based solvers, it led to a degeneration in the success rate of  MPPIs. 
Consequently, while MPPI can achieve comparable or even higher performance in unconstrained scenarios, it produces physically controversial behaviors that are less suitable for realistic hardware deployment than DDPs.

\begin{table*}[t]
\small\sf\centering
\begin{threeparttable}
\caption{Benchmark results comparing optimization-based ME-DDP variants and sampling-based MPPI baselines. The results are categorized into feasible and all (including infeasible) trajectories. {\small\sf{Success rate}} shows the percentage of the trajectories that reach the goal. {\small\sf{Time}} is the average first time step when the trajectory enter the goal region and stays there for ten time steps. {\small\sf{Path}} is the average trajectory length. {\small\sf{Violation}} is constraint violation computed over infeasible trajectories.  The arrow $\uparrow$ indicates a higher value is preferred, while $\downarrow$ indicates a lower value is preferred. }
\label{tab:comprehensive_results}
\begin{tabular}{l l ccc cccc} 
\toprule
& & \multicolumn{3}{c}{\textbf{Feasible Trajectories}} & \multicolumn{4}{c}{\textbf{All (Including Infeasible Trajectories)}} \\
\cmidrule(lr){3-5} \cmidrule(lr){6-9} 
\textbf{Dynamics} & \textbf{Algorithm} & \makecell{\textbf{Success} \\ { \ \textbf{rate [\%]} $\uparrow$}} & \textbf{Time $\downarrow$} & \textbf{Path $\downarrow$} & \makecell{\textbf{Success} \\ \textbf{\ rate [\%] $\uparrow$}} & \textbf{Time $\downarrow$} & \textbf{Path $\downarrow$} & \textbf{Violation $\downarrow$} \\
\midrule
\textbf{2D Car}        &DDP         & 40.0          & \textbf{118} & \textbf{11.6}  & 40.0          & \textbf{118} & \textbf{11.6}      & \checkmark  \\
$n_{x} \in \bR^{4}$    & UG-ME-DDP  & 66.3          & 149          & 12.4           & 75.6          & 153          & 12.6               & \num{1.96e-4} \\
$n_{u} \in \bR^{2}$    & MG-ME-DDP  & 81.2          & 201          & 12.5           & 89.4          & 201          & 12.5               & \num{2.16e-4}  \\
$T_{\mr{mpc}} = 50$    & SV-DDP     & \textbf{81.9} & 136          & 12.5           & \textbf{93.8} & 138          & 12.6               & \num{2.44e-4}  \\
\addlinespace 
                        &UG-MPPI    & 50.0          & 157           & 13.6          & 50.0          & 157          & 13.60 & \checkmark\\
                        &MG-MPPI    & 45.6          & 156           & 13.6          & 45.6          & 156          & 13.56 & \checkmark\\
                        &SV-MPPI    & 78.1          & 175           & 13.7          & 78.1          & 175          & 13.65 & \checkmark \\
\midrule
\textbf{2D Car}        & DDP        & 31.3          & \textbf{127}  & \textbf{12.4} & 31.3          & \textbf{127}  & \textbf{12.4}&\checkmark\\
$n_{x} \in \bR^{3}$    & UG-ME-DDP  & 75.6          & 142           & 13.2          & 80.0          & 143          & 13.3           & \num{4.74e-4} \\
$n_{u} \in \bR^{2}$    & MG-ME-DDP  & 74.4          & 145           & 13.5          & 81.3          & 147          & 13.6           & \num{1.56e-4} \\
$T_{\mr{mpc}}=70$      & SV-DDP     & \textbf{85.0} & 135           & 12.8          & \textbf{95.0} & 138          & 12.9           & \num{2.68e-4} \\
\addlinespace
                        & UG-MPPI   & 62.5          & 151           & 12.9          & 62.5          & 151        & 12.86 & \checkmark \\
                        & MG-MPPI   & 63.1          & 162           & 13.6          & 63.1          & 162        & 13.62 & \checkmark \\
                        & SV-MPPI   & 82.5          & 153           & 13.2          & 82.5          & 153        & 13.21 & \checkmark \\
\midrule
\textbf{Quadrotor}      & DDP       & 37.5          & 145           & 9.55          & 43.8          & 155           &  10.2   & \num{5.28e-6} \\
$n_{x} \in \bR^{12}$    &UG-ME-DDP  & 77.5          & 140           & 9.20          & 83.1          & 141           &  9.26   & \num{1.32e-2} \\
$n_{u} \in \bR^{4}$     &MG-ME-DDP  & 76.9          & 139           & 9.00          & 80.0          & 141           &  9.05   & \num{1.10e-4}\\
$T_{\mathrm{mpc}} = 50$ &SV-DDP     & \textbf{83.1} & \textbf{132}  & \textbf{8.91} & \textbf{89.4} & \textbf{132}  &  \textbf{8.89}   & \num{1.19e-4}  \\
\addlinespace
                        & UG-MPPI   & 43.8          & 137.7         & 8.92          & 43.8          & 138         & 8.92  & \checkmark \\
                        & MG-MPPI   & 62.5          & 155.0         & 9.16          & 62.5          & 155         & 9.16 & \checkmark \\
                        & SV-MPPI   & 75.0          & 152.7         & 9.14          & 75.0          & 153         & 9.14 & \checkmark \\
\midrule
\textbf{Quadrotor}
                        & DDP        & 3.13         & 189           & 11.0          & 5.00          & 193           & 11.1 & \num{7.71e-2}  \\
$n_{x} \in \bR^{12}$    & UG-ME-DDP  & \textbf{91.3}& 141           & 9.36          & \textbf{95.6} & 141           & 9.32 & \num{4.20e-5} \\
$n_{u} \in \bR^{4}$     & MG-ME-DDP  & 83.8         & 153           & 9.20          & 85.0          & 153           & 9.19 & \num{1.75e-4} \\
$T_{\mr{mpc}}=70$       & SV-DDP     & 87.5         & \textbf{136}  & \textbf{8.91} & 94.4          & \textbf{136}  & \textbf{8.88} & \num{7.66e-5}  \\
\addlinespace
                        & UG-MPPI    & 69.4         & 148.2         & 9.17          & 69.4          & 148         & 9.17 & \checkmark \\
                        & MG-MPPI    & 69.4         & 143.9         & 9.18          & 69.4          & 144         & 9.18 & \checkmark \\
                        & SV-MPPI    & 67.5         & 186.4         & 9.39          & 67.5          & 186         & 9.39 & \checkmark \\
\midrule
\textbf{Ant}            & DDP        & 20.0         & 174           & 10.8          & 20.0          & 174           & 10.8     & \checkmark  \\
$n_{x} \in \bR^{29}$    & UG-ME-DDP  & 55.0         & 134           & 11.1          & 57.0          & 138           & 11.1    &\num{6.49e-2} \\
$n_{u} \in \bR^{8}$     & MG-ME-DDP  & \textbf{82.0}& 112           & 11.0          &\textbf{83.0}  & 113           & 10.6    & \num{2.50e-4}  \\
$T_{\mathrm{mpc}} = 40$ & SV-DDP     & 80.0         & 119           & 11.5          & 80.0          & 119           & 11.5     & \checkmark \\
\addlinespace
                        &UG-MPPI     & 53.0         & 73.2          & 9.79          & 53.0          & 73.2          & 9.79   & \checkmark \\
                        &MG-MPPI     & 54.0         & \textbf{65.6} & \textbf{9.34} & 54.0          & \textbf{65.6} & \textbf{9.34} & \checkmark\\
                        &SV-MPPI     & 48.0         & 66.0          & 9.39          & 48.0          & 66.0          & 9.39   & \checkmark\\
\midrule
\textbf{Barkour}        &  DDP      & 10.0          & 448           & \textbf{6.02} & 30.0          & 458           &6.40    & \num{1.21e-4}\\
$n_{x} \in \bR^{37}$    & UG-ME-DDP & 80.0          & 311           & 6.24          & 81.0          & 311           &6.25    & \num{9.13e-6} \\
$n_{u} \in \bR^{12}$    & MG-ME-DDP & 60.0          & 282           & 6.36          & 76.0          & 288           &6.38    & \num{1.56e-4} \\
$T_{\mathrm{mpc}} = 50$ & SV-DDP    &\textbf{91.0}  & 323           & 6.31          & \textbf{92.0} & 324           &6.32    & \num{7.95e-5} \\
\addlinespace 
                        & UG-MPPI   & 35.0          & 291           & 6.14          & 35.0          & 391           &6.14    & \checkmark \\
                        & MG-MPPI   & 30.0          & \textbf{245}  & 6.08          & 30.0          & \textbf{245}  &\textbf{6.08}& \checkmark \\
                        & SV-MPPI   & 43.0          & 343           & 6.29          & 43.0          & 334           &6.29    & \checkmark    \\

\midrule 
\textbf{Barkour-}        & UG-MPPI   & \textbf{95.0} & \textbf{237}  & 6.29          & \textbf{95.0} & \textbf{237}  & 6.29              & \checkmark\\
\textbf{Decouple-}       & MG-MPPI   & 58.0          & 250           & 6.26          & 58.0          & 250           & 6.26              & \checkmark\\
\textbf{Position-Yaw\tnote{a}}   & SV-MPPI   & 71.0          & 276           & \textbf{6.25} & 71.0          & 276           & \textbf{6.25}     & \checkmark\\
\bottomrule
\end{tabular}
\begin{tablenotes}
\item[a] \small{Decouple position and yaw weights of the cost during tuning.}
\end{tablenotes}
\end{threeparttable}
\end{table*}

\begin{figure*}[t]
    \centering
    \begin{subfigure}[b]{0.49\linewidth}
        \centering
        \includegraphics[width=\linewidth]{fig/simulation_envs/2dcar/twod_car_env_8.pdf}
        \caption{Sparse Environment for 2D Car}
    \end{subfigure}
    \begin{subfigure}[b]{0.49\linewidth}
        \centering
        \includegraphics[width=\linewidth]{fig/simulation_envs/2dcar/twod_car_env_19.pdf}
        \caption{Dense Environment for 2D Car}
    \end{subfigure}
    \begin{subfigure}[b]{0.49\linewidth}
        \centering        \includegraphics[width=\linewidth]{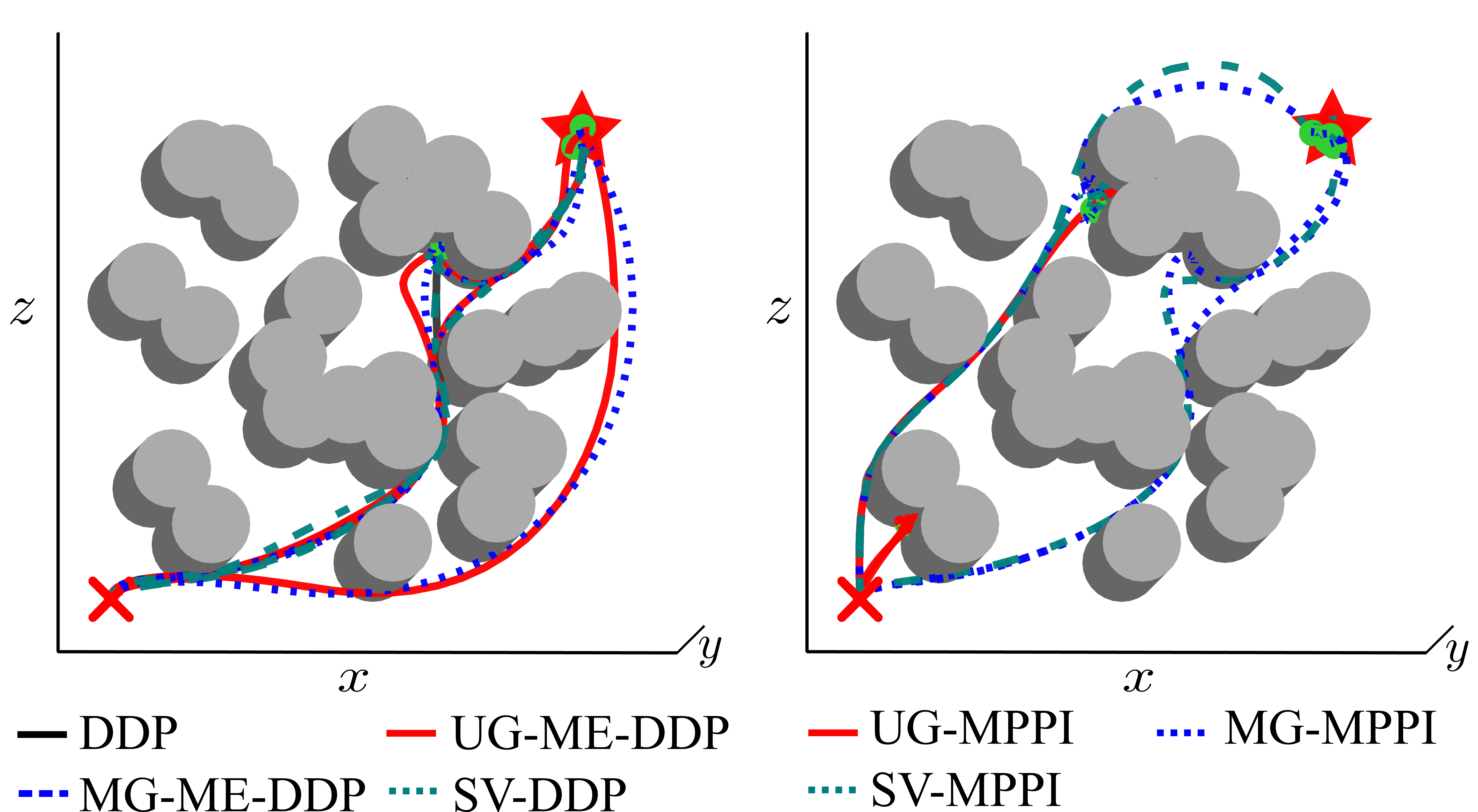}
        \caption{Sparse Environment for Quadrotor}
    \end{subfigure}
    \begin{subfigure}[b]{0.49\linewidth}
        \centering
        \includegraphics[width=\linewidth]{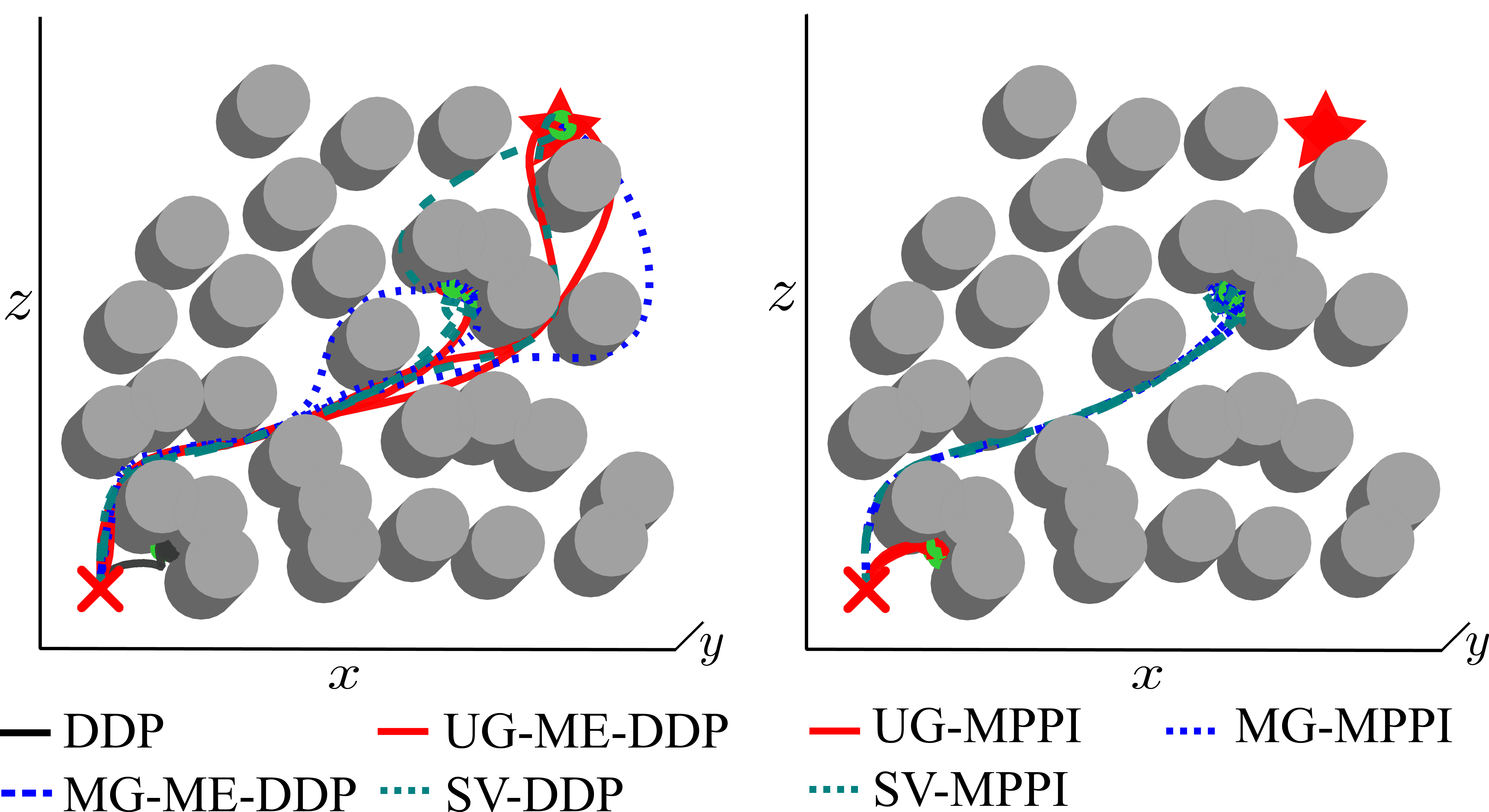}
        \caption{Dense Environment for Quadrotor}
    \end{subfigure}
    \caption{
    Trajectory comparisons for 2D Car and Quadrotor environments.  
    Within each panel, the left image shows DDP, and the right shows MPPI.}
\label{fig:comparison_simulator_2d3d}
\end{figure*}

\begin{figure*}[t]
    \centering
    \begin{subfigure}{1.0\textwidth}
        \centering
        \includegraphics[width=\linewidth]{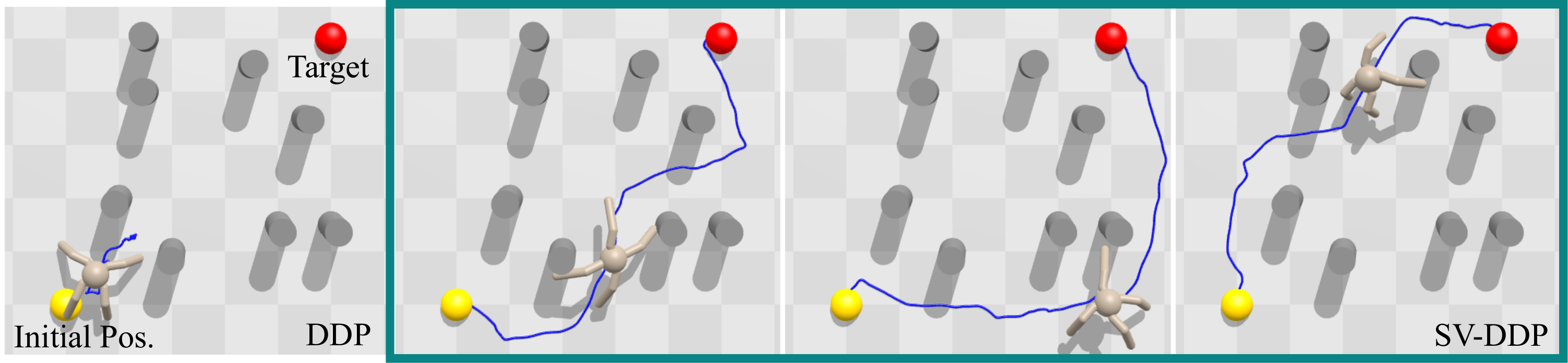}
        \caption{Environment 1.}
        \label{fig:ant_env1}
    \end{subfigure}

    \begin{subfigure}{1.0\textwidth}
        \centering
        \includegraphics[width=\linewidth]{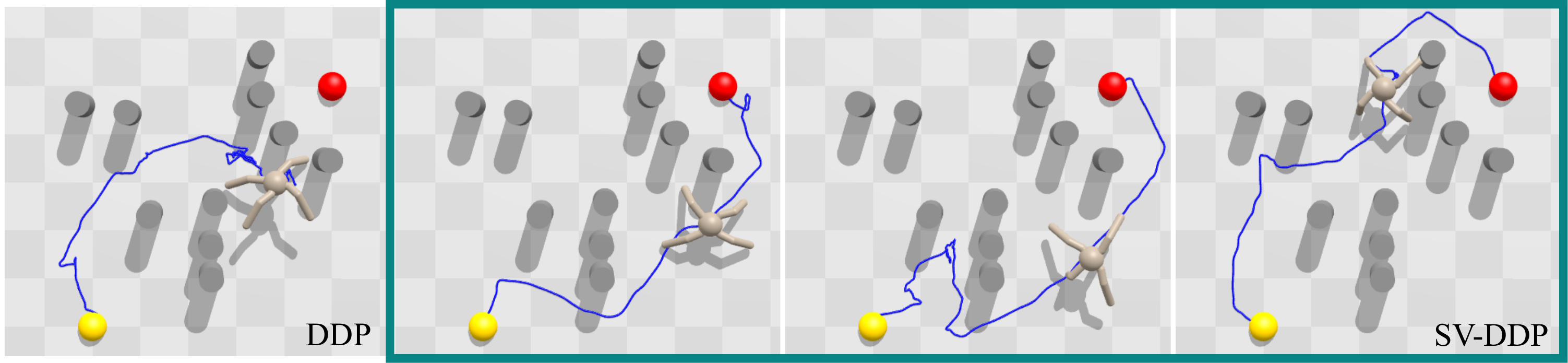}
        \caption{Environment 2.}
        \label{fig:ant_env2}
    \end{subfigure}
    
    \caption{Result of Ant experiments. 
    Trajectories get stuck at poor local minima, and three distinctive trajectories that successfully reach the target via SV-DDP are presented. 
    }
    \label{fig:ant_experiment_overview}
\end{figure*}

\begin{figure*}[t]
    \centering
    \begin{subfigure}{1.0\textwidth}
        \centering
        \includegraphics[width=\linewidth]{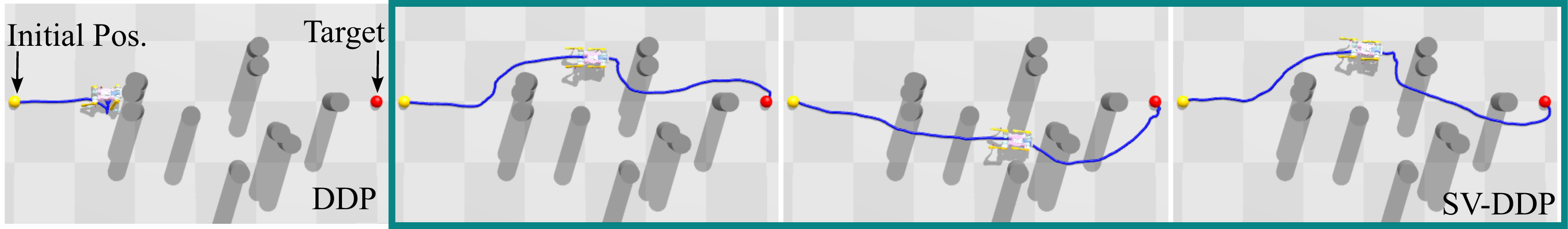}
        \caption{Environment 1.}
        \label{fig:barkour_env1}
    \end{subfigure}
    \begin{subfigure}{1.0\textwidth}
        \centering
        \includegraphics[width=\linewidth]{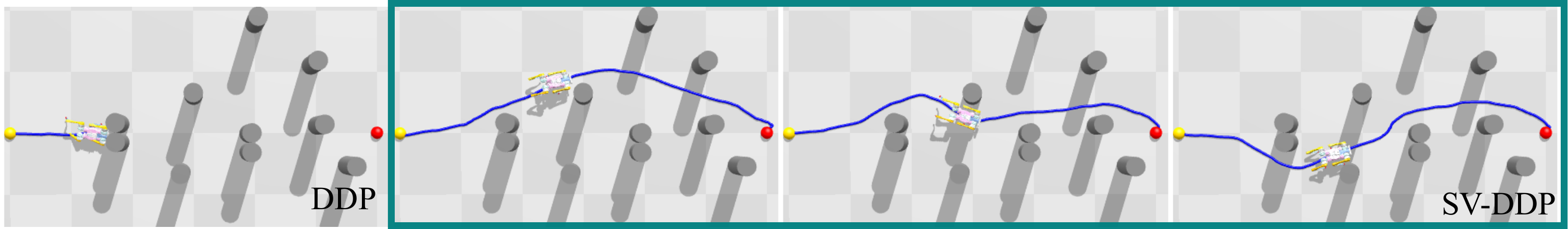}
        \caption{Environment 2.}
        \label{fig:barkour_env2}
    \end{subfigure}
    \caption{Result of Barkour experiments. 
    Trajectories get stuck at poor local minima, and three distinctive trajectories that successfully reach the target via SV-DDP are presented. 
    }
    \label{fig:barkour_experiment_overview}
\end{figure*}

\subsubsection{Results}
To capture the stochastic nature of the algorithms, we run ten trials per environment except for the normal deterministic DDP. We have 16 environments for 2D car and quadrotor (comprising eight sparse and eight dense configurations), and ten environments for Ant and Barkour. This resulted in a total of 160 trials for the car and quadrotor experiments and 100 trials for the Ant and Barkour, from which we provide statistics. The overviews of the experiments are shown in Fig. \ref{fig:comparison_simulator_2d3d} for the 2D car and quadrotor, Fig. \ref{fig:ant_experiment_overview} for Ant, and Fig. \ref{fig:barkour_experiment_overview} for Barkour.

\paragraph{Performance:}
Table. \ref{tab:comprehensive_results} shows the statistics of the results. We have two categories of results based on the constraint violation. 
The first group is feasible trajectories. This group includes trials where the system successfully reached the goal region (defined by an $L_2$ norm less than $0.3$) while maintaining a maximum obstacle constraint violation of less than $10^{-6}$. For this group, we report the {\small\sf{Success rate [\%]}}, the mean time step to steady
state denoted as {\small\sf{Time}}. This metric is defined as the first time step the agent enters and remains in the goal region for at least 10 consecutive steps.  Finally, we report the mean path length denoted as {\small\sf{Path}}. 
The second group relaxes the requirement of constraint satisfaction. This group represents a broader set of trials that reached the goal but allowed for constraint violations. In addition to the three metrics mentioned above, we provide the average constraint violation ({\small\sf{Violation}}).
We have the movies of the 2D car in \href{anc/extension1.mp4}{extension 1}, Quadrotor in \href{anc/extension2.mp4}{extension 2}, Ant in \href{anc/extension3.mp4}{extension 3}, and Barkour in \href{anc/extension4.mp4}{extension 4}. For Barkour, we also provide the results for MPPIs without position-yaw coupling as described in the {\nameref{subsec:param_tune_opt}} section.

The ME-DDP variants consistently outperform standard deterministic DDP, demonstrating that the exploration mechanism helps optimizers explore and find better local minima.

In low-dimensional systems (e.g., 2D car, quadrotor), gradient-based methods exhibit a superior success rate and efficiency through shorter time and path length. Our mechanism using local geometric information, e.g., the gradient and Hessian for optimization and the Hessian for sampling covariance, outperforms purely sampling methods. This suggests that for these systems, local geometric information, specifically the gradient and Hessian, is sufficient to characterize the cost landscape.

In high-dimensional locomotion, the cost landscape is characterized by numerous local minima, making the role of exploration critical. The results demonstrate that while ME-DDP achieves a higher overall success rate, MPPI is capable of steering the system more aggressively, as evidenced by shorter completion times and path lengths. This suggests that while gradient-based methods like ME-DDP often converge to the local basins of their nominal trajectories, MPPIs can discover solutions that DDPs cannot find, such as the yaw-rotation in Barkour (Fig. \ref{fig:barkour_mppi_rotate}) or the jumping gait in the Ant environment presented in extension 3.  Essentially, MPPI can potentially find superior, high-energy modes that DDP cannot, at the cost of a higher failure rate due to the lack of gradient-guided refinement.

Although multimodal policies are designed to capture diverse paths, the results suggest that their efficacy depends heavily on the degree of mode separation and the quality of representative trajectories. In our experiments, we observed that an unimodal policy can sometimes outperform other policies in both DDPs and MPPIs. This is because when the policy fails to separate the trajectories into distinct, meaningful classes, it has a lower local sampling density than the unimodal policy. Furthermore, it fails to provide a broader exploration capability. 
Consequently, they cannot be as efficient as a focused unimodal policy. 
While it remains challenging to draw a universal conclusion on which algorithm consistently outperforms the others across all domains, SV-DDP demonstrates remarkable reliability, consistently maintaining a success rate equal to or greater than 80 \%.

\begin{table*}[!htbp]
\small\sf\centering
\caption{Computational performance benchmark. Comparison of mean execution time and standard deviation [ms] of one time step for the 2D car and quadrotor systems across two look-ahead horizons ($T_{\mr{MPC}}=50, 70$).}
\label{tab:timing_results}
\begin{tabular}{l cc cc}
\toprule
& \multicolumn{2}{c}{2D Car Time [ms]} & \multicolumn{2}{c}{Quadrotor Time [ms]} \\
\cmidrule(lr){2-3} \cmidrule(lr){4-5}
\textbf{Algorithm} & $T_{\mr{mpc}}:$ 50 & $T_{\mr{mpc}}:$ 70 & $T_{\mr{mpc}}:$ 50 & $T_{\mr{mpc}}:$ 70 \\
\midrule
DDP & 19.46 \textpm 3.06 & 25.70 \textpm 3.12 & 21.53 \textpm 2.24 & 29.51 \textpm 2.40 \\
UG-ME-DDP & 24.45 \textpm 1.54 & 32.39 \textpm 1.15 & 26.50 \textpm 1.40 & 35.73 \textpm 1.07 \\
MG-ME-DDP & 23.96 \textpm 1.10 & 32.25 \textpm 1.28 & 25.60 \textpm 0.95 & 35.87 \textpm 0.98 \\
SV-DDP & 25.53 \textpm 3.61 & 33.54 \textpm 3.78 & 26.54 \textpm 1.14 & 37.34 \textpm 1.48 \\
\addlinespace
MPPI & 2.10 \textpm 0.31 & 2.22 \textpm 0.47 & 2.83 \textpm 0.65 & 3.91 \textpm 0.80 \\
\bottomrule
\end{tabular}
\end{table*}

\paragraph{Computational Time:}
We compare computational performance of the DDP, ME-DDP variants, and UG-MPPI using both 2D car and quadrotor dynamics across two look-ahead horizons ($T_{\rm{mpc}}=50$ and $T_{\rm{mpc}}=70$). Both algorithms are implemented in JAX \cite{jax2018github}. Statistics were computed over 200 MPC time steps. As shown in Table \ref{tab:timing_results}, MPPI exhibits significantly faster computational speed compared to DDP-based methods. This is because MPPI bypasses the computation of the gradient and Hessian that are required for DDPs during their backward pass. 
Although the exploration mechanisms in ME-DDP variants introduce additional computational overhead to the normal deterministic DDP, they still meet real-time requirements. Specifically, given the discretization intervals of 20 ms for the 2D car and 10 ms for the quadrotor, the optimization finishes before the next control cycle, ensuring the feasibility of the proposed approach for robotic control. Furthermore, the low execution variance observed across all methods indicates a highly deterministic timing profile, which is critical for maintaining stable control of systems.

For the computational benchmark, we focus on the 2D car and quadrotor to ensure that the timings reflect the algorithmic complexity of the optimizers rather than the overhead of the physics engine. 
Using simple dynamics, we isolate the problem of heavy computation of the propagation of dynamics and other steps in the optimization loop. By doing so, we can provide a clearer assessment of the real-time feasibility of the algorithms.

All timing comparison experiments were conducted on a system equipped with an Intel i9-13900K CPU, a 64Gb of system memory, and an NVIDIA RTX4090 GPU.

\subsection{Hardware Experiments}
To validate the real-time performance and robustness of the proposed ME-DDP framework, we conducted a series of hardware experiments using a quadrotor platform. While simulations provide a controlled environment for algorithmic comparison, physical hardware introduces real-world challenges such as sensor noise and aerodynamic turbulence.
In this section, we provide experimental setup and results.

\subsubsection{Hardware and Experimental Setup:}
\begin{figure}[b]
    \centering
    \includegraphics[width=0.7\linewidth]{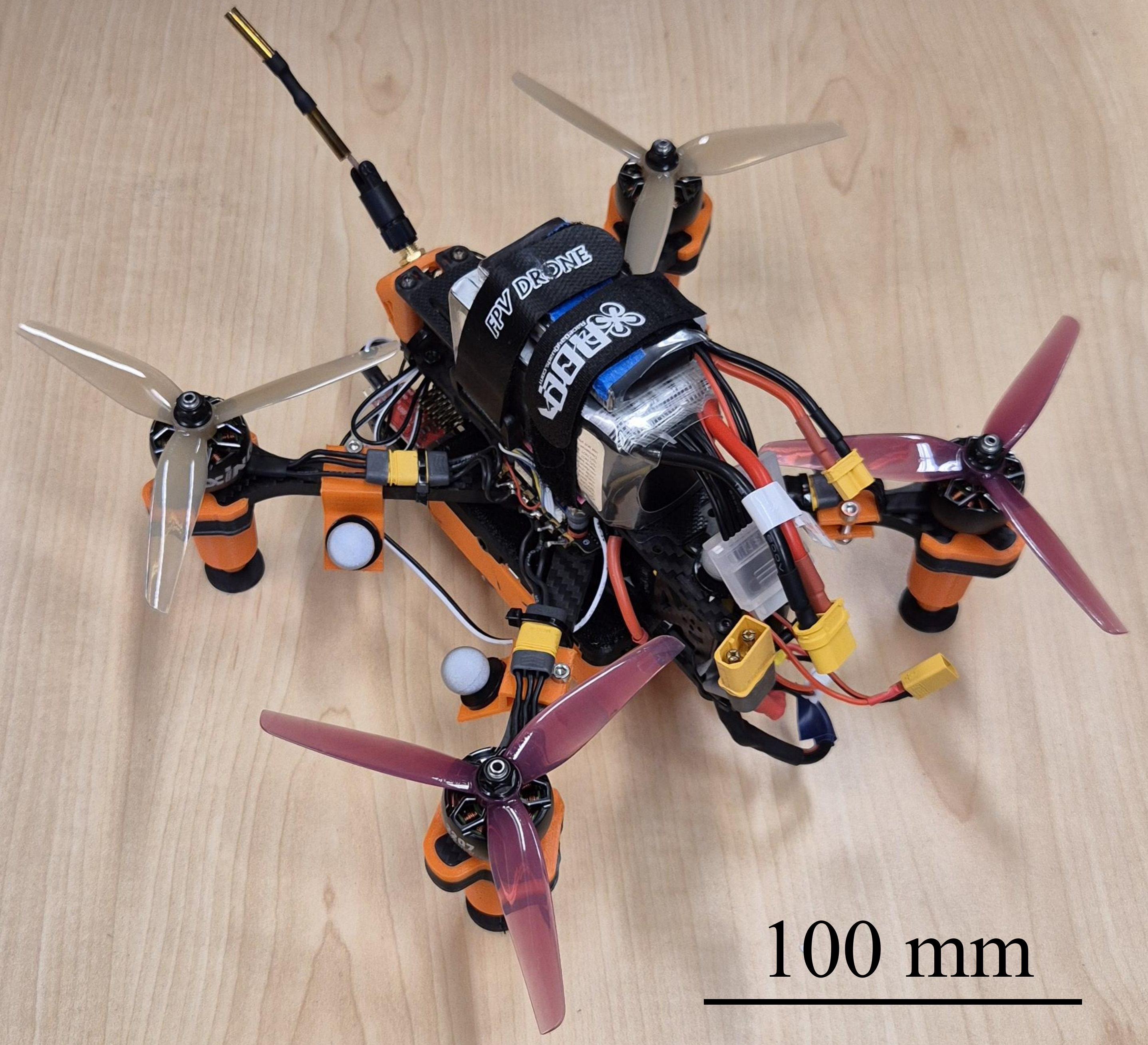}
    \caption{The quadrotor platform used for experimental validation. }
\label{fig:quad_scale_bar}
\end{figure}

The experimental platform is a custom-built quadrotor, as illustrated in Fig. \ref{fig:quad_scale_bar}. The system architecture is divided into three primary components: sensing, computation, and flight control.

\paragraph{State Estimation:} To provide global state estimation, we utilize a VICON motion capture system. The quadrotor's full state is estimated from VICON data using EKF of the PX4-based flight controller \citep{Meier2015px4}.


\paragraph{Ground Station and Optimization:} A ground station receives the state telemetry via a Local 
Area Network. This station serves as the primary computational resource, computing the DDPs in MPC for trajectory optimization. The optimized control sequences are then sent back to the vehicle via a dedicated Wi-Fi link. The control frequency is set to 10 Hz.

\paragraph{Onboard Systems:} The vehicle is equipped with a Raspberry Pi onboard computer, which acts as the bridge between the ground station and the flight hardware. The onboard computer communicates with a flight controller. The controller runs position control based on the optimal state trajectory computed by DDPs. The controller translates the received control commands (positions) into low-level motor signals.

\subsubsection{Results}
\begin{figure}[t]
    \centering
    \includegraphics[width=0.85\linewidth]{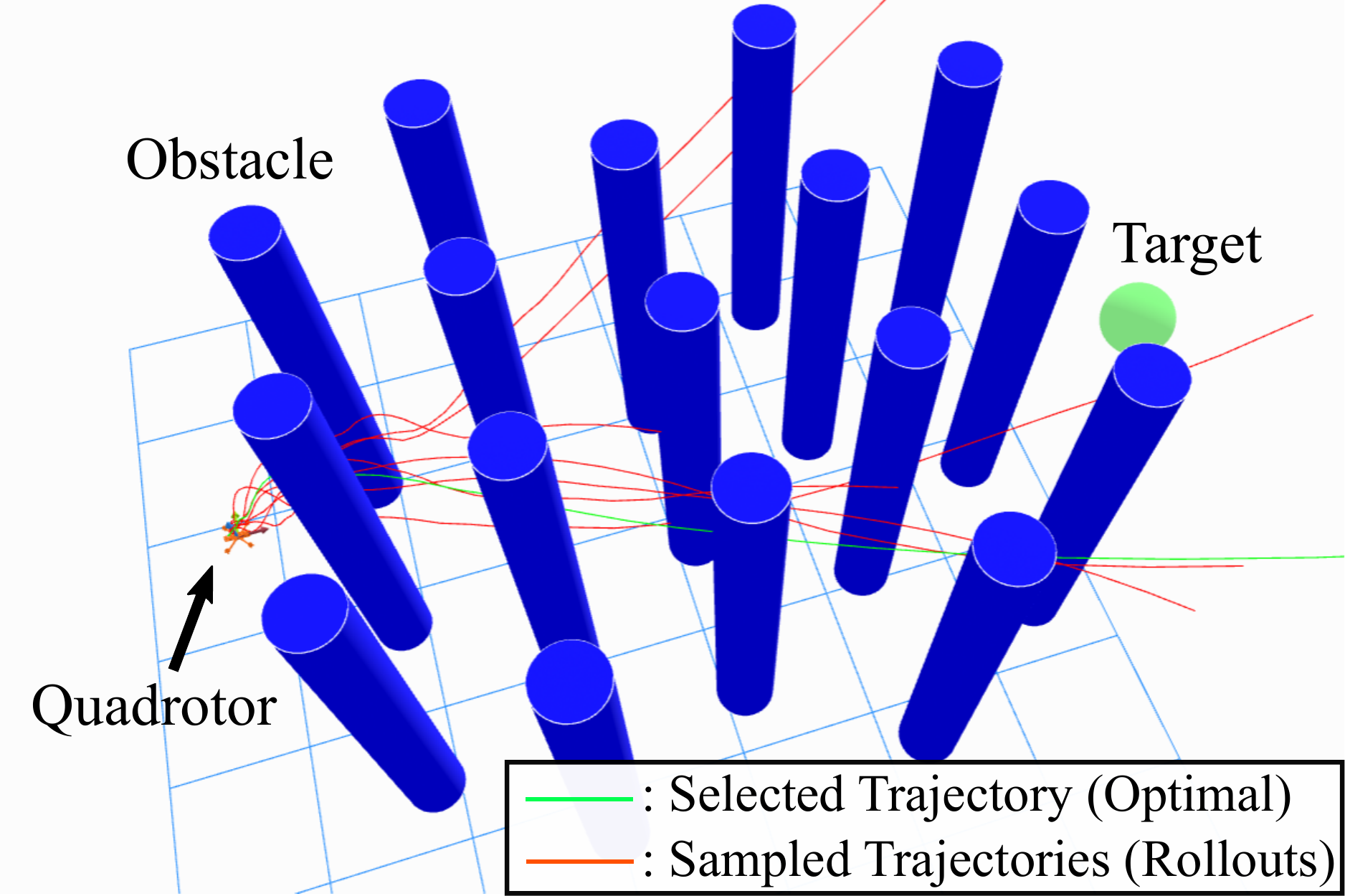}
    \caption{3D visualization of a hardware experiment.}
\label{fig:quad_forest_fox_glove}
\end{figure}

We test the performance of the DDP family, i.e., standard deterministic DDP and three ME variants (UG-ME-DDP, MG-ME-DDP, and SV-DDP), across two environments with different cylindrical obstacle configurations. An overview of the experiment is visualized in Fig. \ref{fig:quad_forest_fox_glove}. To evaluate the robustness of each algorithm and characterize the stochastic behavior inherent in the ME-DDP variants, we conduct multiple trials for each algorithm across different environments.

Instead of physical obstacles, we used virtual representations. For qualitative evaluation, these virtual obstacles are overlaid on a top-down camera feed of the experimental arena. In \href{anc/extension5.mp4}{extension 5} and \href{anc/extension6.mp4}{extension 6}, we present a synchronized side-by-side visualization, with the overlaid real-world video on one side and the corresponding ROS visualization on the other.

\begin{figure*}[!htbp]
    \centering
    \begin{subfigure}[b]{\linewidth}
        \centering
        \includegraphics[width=\linewidth]
    {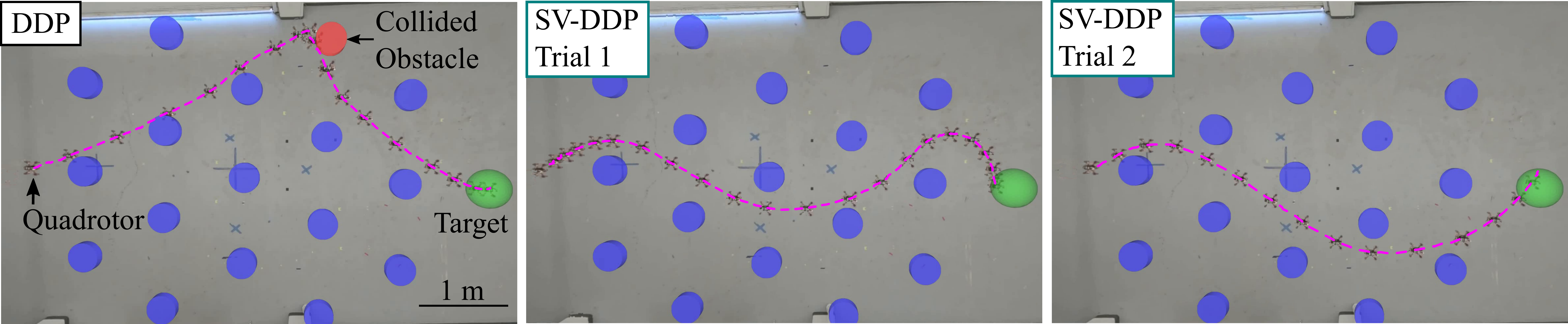}
        \caption{Environment 1: Forest of obstacles}
    \end{subfigure}
    
    \centering
    \begin{subfigure}[b]{\linewidth}
        \centering
        \includegraphics[width=\linewidth]
    {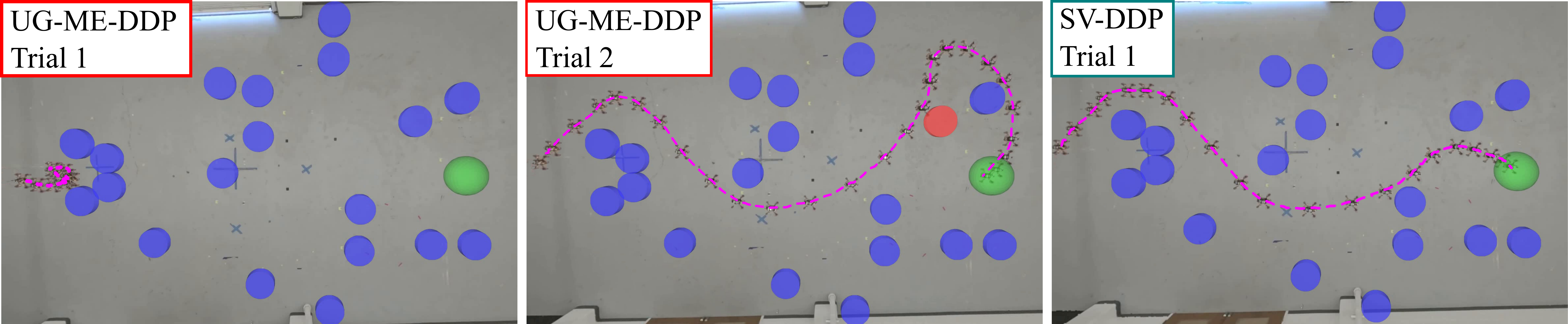}
        \caption{Environment 2: Forest with poor local minima/deadlocks}
    \end{subfigure}
    \caption{Representative resulting trajectories from hardware experiments in two different environments. The drone positions were captured at 0.5 s intervals. Obstacles are colored based on their states. Blue indicates a collision-free, while red denotes a collision verified by ground truth position. In the left panel of (b), the quadrotor is captured at a deadlock.
    }
\label{fig:quad_overlaid_two_env}
\end{figure*}
The results of the experiments are provided in Table. \ref{tab:hardware_results}, and the trajectories of the quadrotor and overlaid obstacle are shown in Fig \ref{fig:quad_overlaid_two_env}. Environment 1 shown in Fig \ref{fig:quad_overlaid_two_env} (a) is a forest of obstacles. As illustrated by the overlaid trajectories, without exploration, the algorithm finds a simple straight path, which acts as a detour to avoid the obstacles. With exploration, on the other hand, it actively explores the cost landscape and discovers shorter, more direct passages through the forest. 

Environment 2 in Fig \ref{fig:quad_overlaid_two_env}  (b) has some specific concave configurations designed to induce multiple poor local minima that can cause deadlocks. In addition to these, the environment incorporates forest-type configurations, as in environment 1. This environment tests the capability of the proposed sampling-based policies to escape these local minima. Furthermore, it simultaneously examines the exploration capability demonstrated in environment 1. 

In Table \ref{tab:hardware_results}, we classify each trial into four categories. 
First, a trial is classified as \textit{\small\sf{Safe Success}}, when the quadrotor reaches the target without violating constraints. This represents the ideal outcomes. Second, the category \textit{\small\sf{Goal with Coll.}} is for the cases where the quadrotor reaches the target but hits obstacles. For these two categories, we provide the average time to reach the target. Finally, \textit{\small\sf{Failure to Goal}} represents cases where the quadrotor fails to hit the target, which we further divide into two groups based on the failure modes. \textit{\small\sf{Deadlock without Coll.}} occurs when the quadrotor is captured by a poor local minimum and spends too much time there.
\textit{\small\sf{Coll.}} is for cases where the quadrotor can escape from the deadlock, but subsequently collides with an obstacle and times out.  We distinguish between these two modes to provide a more granular analysis of constraint satisfaction.


Overall, the sampling mechanism facilitates exploration, and since the constraints are encoded as part of the cost, it also facilitates constraint satisfaction.

The hardware results confirm the trends observed in simulation, where the multimodal policy fails to provide better solutions compared to the standard unimodal one. We observed that without an explicit separation mechanism, the multimodal policy essentially becomes a unimodal one with fewer trajectories.

In this case, the multimodal policy has a lower local sampling density than the unimodal policy. Furthermore, it fails to provide a broader exploration capability. Consequently, it may act as a middle ground of poor performance: it is not as efficient as a focused unimodal policy, nor as diverse as a SV policy.

This phenomenon explains the higher deadlock rates observed for the MG-ME-DDP. The policy wasted its exploration budget without successfully identifying alternative routes. In contrast, the unimodal UG-ME-DDP prioritizes a high-resolution search around the most viable candidate.

\begin{center}
\begin{table*}[h]
\small\sf\centering
\caption{Hardware Performance Analysis. Outcomes are categorized into successful runs and failures with collision. Failures are further distinguished by whether the quadrotor reached the target (Goal with Coll.) or spent too much time on a poor local solution and cannot reach the target with or without collision (Deadlock without Coll./Coll.). The sum of the four outcome categories for each method equals 100 \%.}
\label{tab:hardware_results}
\begin{tabular}{l c cc cc cc}
\toprule
& & \multicolumn{2}{c}{\textbf{Safe Success}} & \multicolumn{2}{c}{\textbf{Goal with Coll.}} & \multicolumn{2}{c}{\textbf{Failure to Goal}} \\
\cmidrule(lr){3-4} \cmidrule(lr){5-6} \cmidrule(lr){7-8}
\textbf{Algorithm} & runs & rate[\%] & time[s] & rate[\%] & time[s] & Deadlock without Coll.[\%] & Coll.[\%] \\
\midrule
DDP       & 9  & 11.1           &  11.71  & 44.4    & 14.06     & 33.3  & 11.1 \\
UG-ME-DDP & 9  & 66.7           &  13.56  & 22.2    & 12.30     & 11.1  &  0.0 \\ 
MG-ME-DDP & 11 & 54.5           &  14.17  & 27.3    & 11.92     & 18.2  &  0.0 \\ 
SV-DDP    & 12 & \textbf{83.3}  &  13.37  &  8.3    & 11.05     &  8.3  &  0.0 \\
\bottomrule
\end{tabular}
\end{table*}
\end{center}


\section{Conclusion}

This work builds upon the Maximum Entropy (ME) framework to provide a unified perspective on the hybrid control mechanism that combines gradient- and sampling-based optimization.

While the core objectives of the ME framework have been previously established, we provide an analytical foundation that clarifies the role of the control Hessian $Q_{uu}$. Specifically, we provide a rigorous way of understanding the physical and mathematical interpretations of sampling from the inverse of $Q_{uu}$, revealing how this mechanism interacts with the non-convex cost landscapes inherent in robotics.

Our benchmarking across four systems in simulation reveals a distinct trade-off between structured optimization and stochastic exploration, advocating for a hybrid optimization approach that moves beyond pure sampling:
\begin{enumerate}
\item \textbf{Low-Dimensional Systems (2D Car, Quadrotor):} In these systems, ME-DDP variants demonstrate the benefits of leveraging local information for optimization and exploration. We found that the local gradient and Hessian provided by the DDP backward pass achieve superior success rate, and more efficient path lengths compared to MPPIs that rely on pure sampling.

\item 
\textbf{High-Dimensional Systems (Ant, Barkour):} In complex systems, ME-DDP remains highly reliable, maintaining a high overall success rate due to its structured refinement. However, we observed that MPPI is capable of discovering global modes that DDP-based methods may miss. While MPPI can struggle with success rates due to the lack of gradient-based refinement, its exploratory nature offers a unique advantage in discovering non-local solutions.
\end{enumerate}

Regarding practical implementation, we demonstrate that while MPPI is computationally efficient per iteration, our DDP-based variants are sufficiently fast for real-time quadrotor control. We validate the efficacy of the algorithms through hardware experiments with a quadrotor navigating cluttered environments, proving that the proposed hybrid approach is robust for real-world deployment.

We conclude that for most robotics applications, providing a rigorous local structure through the ME-DDP framework offers a reliable path to high-performance control while remaining computationally feasible for hardware deployment.

Future research directions include implementing the full feedback structure of the optimal controller on hardware, exploring alternative mechanisms for constraint satisfaction beyond relaxed-$\log$ barrier functions, and establishing theoretical convergence guarantees and conditions.


\section*{Author Contributions}
\textbf{Yuichiro Aoyama}: Conceptualization, Formal analysis, Investigation, Software, Methodology, Visualization, Writing-original draft.
\textbf{Minchan Jung}: Conceptualization, Formal analysis, Investigation, Software, Methodology, Visualization, Validation, Writing-review \& editing.
\textbf{Akash Ratheesh}: Investigation, Software, Methodology, Visualization, Writing-review \& editing. 
\textbf{Evangelos A. Theodorou}: Conceptualization, Supervision, Project administration, Writing-review \& editing.
\section*{Statements and Declarations} 
\subsection*{Ethical considerations}
This article does not contain any studies with human or animal participants.
\subsection*{Consent to participate}
This article does not contain any studies with human or animal participants.
\subsection*{Consent for publication}
Not applicable.
\begin{dci}
The author(s) declared no potential conflicts of interest with respect to the research, authorship, and/or publication of this article.
\end{dci}
\begin{funding}
The author(s) disclosed receipt of the following financial support for the research, authorship, and/or publication of this article:  


Akash Ratheesh and Evangelos A. Theodorou were supported by the National Aeronautics and Space Administration under University Leadership Initiative [grant number 80NSSC22M0070] and the Army Research Office [grant number W911NF2010151].

Minchan Jung was supported by Korea Institute for Advancement of Technology (KIAT) grant funded by the Korea Government (MOTIE), Human Resource Development Program for Industrial Innovation (Global) [grant number RS-2024-00435406].

\end{funding}











\appendix

\renewcommand{\thealgocf}{\thesection.\arabic{algocf}}
\renewcommand{\theequation}{\thesection.\arabic{equation}}

\section*{Appendix}
\section{Index to Multimedia Extensions}

\begin{table}[H]
\centering\sf\small
\setlength{\tabcolsep}{4pt}
\begin{tabularx}{\columnwidth}{ccX} 
\toprule
Extension & Media type & Description \\
\midrule
1 & Video & Algorithmic comparison: 2D Car \\
2 & Video & Algorithmic comparison: Quadrotor\\
3 & Video & Algorithmic comparison: Ant \\
4 & Video & Algorithmic comparison: Barkour \\
5 & Video & Hardware experiments for environment 1\\
6 & Video & Hardware experiments for environment 2\\
\bottomrule
\end{tabularx}
\end{table}

\setcounter{equation}{0}
\setcounter{algocf}{0}
\setcounter{figure}{0}

\renewcommand{\theequation}{\thesection.\arabic{equation}}
\renewcommand{\thealgocf}{\thesection.\arabic{algocf}}
\renewcommand{\thefigure}{\thesection.\arabic{figure}}
\section{Maximum Entropy DDP}
This section provides a derivation of Maximum Entropy DDP (ME-DDP). Starting with the entropic regularized cost and corresponding optimal policy, we derive unimodal and multimodal policies.


\subsection{Optimal Policy}
To solve \eqref{eq:max_entropy_with_cnst},
we set up a Lagrangian $\cL$ with a Lagrangian multiplier $\lambda \in \mathbb{R}$ as:
\begin{align*}
\mathcal{L} &= \mathbb{E}_{u \sim \pi}[l(x,u) + \regV(x')] - \tau \cH[\pi(\cdot | x)] \\
& \hspace{40mm}+ \lambda 
 \Big(1-\int \pi(u|x)\mathrm{d}u \Big)\\
&= \mathbb{E}_{u \sim \pi}[l(x,u) + \regV(x') + \tau \ln \pi(u | x) - \lambda] + \lambda \\
&= \int_{\mathbb{R}^{n_{u}}}\pi(u|x) [\regQ(x,u) + \tau \ln \pi(u|x) - \lambda]\mathrm{d}u + \lambda,
\end{align*}
where we dropped time instance $t$ and denote $x_{t+1}$ as $x'$ for readability. $\pi(\cdot | x)$ means that it originally was a function of $u$, but $u$ vanishes after expectation was taken over $u$. We used the definition of $\regQ(x,u) = l(x, u) + \regV(x')$ in the last equality. 
From the first-order optimality condition, taking the functional derivative of $\mathcal{L}$ with respect to $\pi$ and setting it zero, we have 
\begin{align*}
\regQ(x,u) + \tau \ln \pi(u|x) - \lambda + \tau = 0. 
\end{align*}
Thus, the optimal policy $\pi^{\ast}$ is obtained as
\begin{align}\label{eq:MEDDP_pi_star_with_lam}
\pi^{\ast}(u|x) = \exp \Big[ -\frac{1}{\tau}(\regQ(x,u) - 
\lambda) - 1\Big].
\end{align}
From the constraint in \eqref{eq:max_entropy_with_cnst}, we eliminate $\lambda$ from the above equation. First, we plug \eqref{eq:MEDDP_pi_star_with_lam} into the constraint, having
\begin{align}\label{eq:ME_DDP_Z_inv}
\notag
1 = \int \pi^{\ast}(u|x) \mathrm{d}u =
\int \exp \Big[ -\frac{1}{\tau} \regQ(x,u) \Big]
\exp\Big[ 
\frac{\lambda}{\tau}-1
\Big] \mathrm{d}u \\
\Leftrightarrow
\exp \Big[ \frac{\lambda}{\tau} -1 \Big] = \Big( \int 
\exp\Big[ \frac{1}{\tau}\regQ(x,u) \mathrm{d}u\Big]\Big)^{-1} = Z(x)^{-1},
\end{align}
where $Z(x)$ is the partition function of $\pi$:
\begin{align}\label{eq:MEDDP_Z}
Z(x) = \int \exp\Big[ 
-\frac{1}{\tau}\regQ(x,u)
\Big] \mathrm{d}u.
\end{align}
By substituting \eqref{eq:ME_DDP_Z_inv} into  \eqref{eq:MEDDP_pi_star_with_lam}, the optimal policy is obtained as 
\begin{align}\label{eq:sup_pi_star}
\notag
\pi^{\ast}(u|x) &= \exp \Big[ -\frac{1}{\tau}\regQ(x,u)\Big]
\exp \Big[ \frac{\lambda}{\tau}-1 \Big] \\
&= Z(x)^{-1}\exp\Big[ -\regQ(x,u)/\tau\Big]. 
\end{align}

\paragraph{Value function and Partition function}
To derive the relationship between the value and partition functions in 
\eqref{eq:me_ddp_pi_star_uni}, we transform \eqref{eq:sup_pi_star} to get
\begin{align*}
\regQ(x,u) &= -\tau \ln(Z(x) \pi^{\ast}(u|x)) \\
&= -\tau (\ln Z(x) + \ln \pi^{\ast}(u|x)).
\end{align*}

In \eqref{eq:max_entropy_with_cnst}, since the entropy $\cH[\pi(\cdot|x)]$ is not a function of $u$, the term can be inside of the expectation:
\begin{align*}
\tilde{V}(x)
&= \min_{\pi}\big \{ \mathbb{E}_{u\sim\pi}\big[l(x, u) + \tilde{V}(x') - \tau  \cH[\pi(\cdot|x)] \big] \big\},
\end{align*}
From the above two equations, by taking the expectation with respect to the optimal policy, we get the following relationship.
\begin{align}\label{eq:sup_V_islog_Z}
\tilde{V}(x) &= \mathbb{E}_{\pi^{\ast}}[\regQ(x,u)-\tau \cH[\pi^{\ast}]] \\\notag
& = \int {\pi^{\ast}(u|x)} \big[-\tau \big(\ln{Z(x)} + \ln\pi^{\ast}(u|x) \big) \notag\\
\notag
&\hspace{40mm}+\tau \ln \pi^{\ast}(u|x)\big] \mathrm{d}u \\\notag
&= -\tau \ln Z(x) \int \pi^{\ast}(u|x)\mathrm{d}u \\\notag
&= -\tau \ln Z(x).
\end{align}

\paragraph{Normal and Regularized Value Functions}
We proceed to establish the relationship between the regularized and normal value functions.
Assume that the regularized value function $\regV_{t}(x_{t})$ is obtained by adding a constant $c_{t}$ to the normal value function $V_{t}(x_{t})$:
\begin{equation}\label{eq:V_reg_assumption}
    \regV_{t}(x_{t}) = V_t(x_{t}) + c_t
\end{equation}
At the terminal time step $t=T$, the value function is defined by the terminal cost $\Phi(x_{T})$ and since the entropy $\cH[\pi(\cdot|x_{t})]$ is defined only up to $t=T-1$, we have
\begin{align*}
V_{T}(x_{T}) = \regV_{T}(x_{T}) = \Phi(x_{T}), 
\end{align*}
which implies that the hypothesis holds with $c_{T}=0$.

Next, assume the hypothesis \eqref{eq:V_reg_assumption} holds for time step $t+1$: $\regV_{t+1}(x_{t+1}) = V_{t+1}(x_{t+1}) + c_{t+1}$. We must show it holds for time step $t$.
From the definition of the regularized $Q$ function $\regQ$, we have
\begin{align}\label{eq:regQ_Q_plus_c}
\regQ_t(x_t, u_t) &= \rlap{$\overbrace{\phantom{ l(x_t, u_t) + V_{t+1}(x_{t+1})}}^{\regQ(x_{t}, u_{t})}$} l(x_t, u_t) + \underbrace{V_{t+1}(x_{t+1}) + c_{t+1}}_{\regV_{t+1}(x_{t+1})} \\\notag
&= Q(x_{t}, u_{t}) + c_{t+1}.
\end{align}
To compute integral of $\regQ$ to obtain the partition function $Z(x)$, we consider a quadratic approximation of $Q$ around a pair of nominal trajectories as in \eqref{eq:Q_quad_expand}. We complete the square of the deviation term with respect to $\delta u$:
\begin{align}\label{eq:sup_MEDDP_expanded_Q}
Q(x,u) &\approx \underbrace{l(\bar{x}, \bar{u}) + V(\bar{x}')}_{Q(\bar{x}, \bar{u})} + \delta Q(\delta x, \delta u),\\\notag
\delta Q (\delta x, \delta u)
&= \frac{1}{2} \delta u^{\tr}Q_{uu}\delta u + \big[Q_{u} + Q_{ux}\delta x  \big]^{\tr}\delta u \\\notag 
& \hspace{30mm}+ \Big( Q_{x}^{\tr}
+ \frac{1}{2}\delta x ^{\tr}Q_{xx} \Big)\delta x\\\notag
&= \frac{1}{2}(\delta u - \delta u^{\ast})^{\tr}Q_{uu}(\delta u - \delta u^{\ast}) \\ \notag 
&\hspace{10mm} \underbrace{-\frac{1}{2} \delta u^{\ast \tr}Q_{uu}\delta u^{\ast} + \Big( Q_{x}^{\tr}
+ \frac{1}{2}\delta x ^{\tr}Q_{xx} \Big) \delta x}_{\delta Q(\delta x, \delta u^{\ast})},
\end{align}
where the optimal control $\delta u^{\ast}$ is given in \eqref{eq:delta-u-star}. 
This procedure can be seen as computing the minimizer ($\delta u^{\ast}$) by completing the square of the quadratic function. Furthermore, since $V_{t}(x_{t}) = Q_{t}(x_{t}, u_{t}^{\ast})\approx Q_{t}(\bar{x}_{t}, \bar{u}_{t}) + \delta Q_{t}(\delta x_{t}, \delta u^{\ast}_{t})$, we have
\begin{equation*}
Q(x_{t}, u_{t}) = V(x_{t}) + \frac{1}{2}(\delta u - \delta u^{\ast})^{\tr}Q_{uu}(\delta u - \delta u^{\ast}),
\end{equation*}
and with \eqref{eq:regQ_Q_plus_c},
\begin{equation}\label{eq:regQ_V_c_quadraticu}
\regQ_{t}(x_{t}, u_{t}) = V(x_{t}) + c_{t+1} + \frac{1}{2}v^{\tr}Q_{uu}v,
\end{equation}
where we put $v =\delta u - \delta u^{\ast}$.
Assuming that $Q_{uu}$ is PD, the quadratic approximation also allows us to compute $Z$ in \eqref{eq:MEDDP_Z} using Gaussian integral, that is, an integral has a form of 
\begin{align*}
\int \exp \Big[ \frac{1}{2}y^{\tr}A^{-1} y\Big] \mathrm{d}y = [(2\pi)^{n_{y}}|A^{-1}|]^{1/2}, 
\end{align*}
with $y\in \bR^{n_{y}}$, and a symmetric PD matrix $A\in\bR^{n_{y} \times n_{y}}$. Substituting \eqref{eq:regQ_V_c_quadraticu} into \eqref{eq:MEDDP_Z}, we have
\begin{align*}
Z(x)
&= \int \exp\Big[ -\frac{V(x)+c_{t+1}}{\tau}\Big]  \exp\Big[
-\frac{1}{2}v^{\tr}\frac{Q_{uu}}{\tau}v
\Big] \mathrm{d}u \\
&= \exp\Big[ -\frac{V(x)+c_{t+1}}{\tau}\Big][(2\pi)^{n_{u}}|\tau Q_{uu}^{-1}|]^{1/2}.
\end{align*}
When $Q_{uu}$ is not PD, it needs to be regularized as in \eqref{eq:DDP_regularization}. We revisit the regularization later when computing the unimodal Gaussian policy.
Substituting this $Z$ back into \eqref{eq:sup_V_islog_Z}, the value function is obtained as the sum of value function of normal DDP and a term induced by the Gaussian policy which we call $V_{H}$:
\begin{align}\label{eq:me_ddp_value_function_lnz}
\tilde{V}(x) &= -\tau \ln Z(x) \\\notag
&=  -\tau \big[ 
-\frac{V(x)+c_{t+1}}{\tau} + \frac{1}{2}\ln((2\pi)^{n_{u}}|\tau Q_{uu}^{-1}|)) \big]\\\notag
&= V(x) + c_{t+1} +\underbrace{\frac{\tau}{2}\big[
\ln |Q_{uu}| -n_{u} \ln(2\pi \tau)
\big]}_{V_{H}} \\\notag
&=V(x) + \underbrace{c_{t+1} + V_{H}}_{c_{t}}.
\end{align}

The term $c_t$ is purely a function of $c_{t+1}$ and the Hessian $Q_{uu}$. Since $Q_{uu}$ is evaluated along the nominal trajectory $(\bar{x}_{t}, \bar{u}_{t})$, and not along $(x_{t}, u_{t})$, $c_t$ remains state-independent.

From \eqref{eq:V_reg_assumption} and \eqref{eq:regQ_Q_plus_c} and \eqref{eq:me_ddp_value_function_lnz},  the normal value functions $(V, Q)$ and the regularized ones $(\regV, \regQ)$ are identical up to the constant, therefore, the Ricatti recursion used to propagate derivatives back in time (\eqref{eq:Qexpanded_derivative}, \eqref{eq:value_update_with_gain}) are the same among them, and therefore the derivatives are the same, for instance:
\begin{equation}\label{eq:regQuu_Quu_arethesame}
Q_{uu} = \regQ_{uu}.
\end{equation}
The difference shows up in  zeroth-order terms:
\begin{align*}
\regV_{t}(\bar{x}_{t}) &= V_{t}(\bar{x}_t) + c_{t}, \quad
\regQ_{t} = Q_{t} + c_{t+1} \\
\quad c_{t} &= c_{t+1} + V_{H,t}, \quad\quad c_{T} = 0,
\end{align*}
with $t=1,\dots, T-1$, where $c_{t}$ captures the effect of the entropy term at time step $t$.

\subsection{Unimodal policy}

Plugging \eqref{eq:regQ_Q_plus_c} into the optimal policy in  \eqref{eq:sup_pi_star}, and using \eqref{eq:regQuu_Quu_arethesame}, we obtain
\begin{align*}
\pi^{\ast}(u|x) &\propto \exp \big[ -  \regQ(x,u)/{\tau}\big] \\ &\propto \exp \Big[- {\frac{1}{2\tau}}
(\delta u - \delta u^{\ast})^{\tr}Q_{uu}(\delta u - \delta u^{\ast})\big] \\
&\propto
\exp \big[-{\frac{1}{2}}
(\delta u - \delta u^{\ast})^{\tr}\Sigma^{-1}(\delta u - \delta u^{\ast})\big], 
\end{align*}
with $\Sigma = \tau Q_{uu}^{-1}$.
This equation indicates that the optimal policy is normally distributed with mean $\delta u^{\ast}$ and covariance $\tau Q_{uu}^{-1}$. When $Q$ is not locally convex, we add a regularization term to $\delta Q$ as in \eqref{eq:DDP_regularization}. In this case, the covariance is also regularized as 
\begin{align*}
&\delta Q + \frac{1}{2}\mu_{\mr{reg}}\delta u^{\tr}\delta u \\
=&\frac{1}{2} (\delta u-\delta u^{\ast})^{\tr}Q_{uu}^{\rm{reg}}(\delta u - \delta u^{\ast}) + (\text{terms without $\delta u$}),
\end{align*}
which yields 
\begin{align*}
\pi^{\ast}(u|x) \propto
\exp \big[-{\frac{1}{2\tau}}
(\delta u - \delta u^{\ast})^{\tr}[Q_{uu}^{\rm{reg}}]^{-1}(\delta u - \delta u^{\ast})\big], 
\end{align*}
with $\delta u^{\ast}$ computed with $Q_{uu}^{\rm{reg}}$. In fact, the requirement of $Q_{uu}$ being PD serves a dual purpose: it ensures the existence of a unique minimizer for the DDP sub-problem while simultaneously defining a valid covariance for the Gaussian policy. Hereafter, we treat $Q_{uu}$ as a PD matrix. In cases where it is not PD, $Q_{uu}$ implicitly represents the regularized one used to satisfy the requirements of both the DDP backward pass and the covariance.

\subsection{Multimodal policy}

In this section, we consider approximating the value function using multiple trajectories. We first introduce the additive structure of the value functions. Then, we show how to compose the multimodal policy from the unimodal policies associated with each trajectory.

\paragraph{Additive structure:} Instead of performing a single quadratic approximation of the cost around a pair of nominal trajectories, we consider the combined approximation. We start from the terminal cost $\Phi(x_{T})$ with $N$ sets of trajectories:
$
X^{(n)} = [x_
{1}^{(n)}, \dots, x_{T}^{(n)}], \quad 
U^{(n)} = [u_
{1}^{(n)}, \dots, u_{T-1}^{(n)}],
$
with $n=1 ,\dots, N$. Using the LogSumExp operation and the terminal condition \eqref{eq:value_terminal_cnd}, we first approximate the value function at terminal timestep $T$ over all $N$ trajectories as 
\begin{align}\label{eq:MEDDP_combined_terminal}
\regV(x_{T}) 
= -\tau \ln \sum_{n=1}^{N} \exp\Big[
-\frac{1}{\tau}\Phi^{(n)}(x_{T})
\Big],
\end{align}
where $\Phi^{(n)}(x)$ denotes the terminal cost $\Phi(\cdot)$ evaluated for $x^{(n)}$. To proceed, we introduce the exponential transformation
$z_{t} = \mathcal{E}_{\tau }[\regV(x_{t})]$, with $\mathcal{E}_{\tau}(y) = \exp[
-y/{\tau}]$, which transforms the cost and value functions into reward and desirability functions, respectively: 
\begin{align*}
r_{t} = \mathcal{E}_{\tau }[l_{t}], \quad  
r_{T} = \mathcal{E}_{\tau }[\Phi(x_{T})], \quad  
z_{t} = \mathcal{E}_{\tau }[\regV(x_{t})].
\end{align*}
From \eqref{eq:MEDDP_combined_terminal} and \eqref{eq:value_terminal_cnd},
at the terminal time step, $z$ (the approximation by $z^{(n)}$s) is written as a sum of $z^{(n)}$s due to the property of the exponential function as below.
\begin{align*}
z_{T}(x_{T}) &= 
\exp \Big[
-\frac{1}{\tau}\Big(-\tau \ln \sum_{n=1}^{N}\exp \Big[ -\frac{1}{\tau}\Phi^{(n)}(x)\Big]\Big)
\Big]\\\notag 
&=\sum_{n=1}^{N} \exp\Big[
-\frac{1}{\tau}\Phi^{(n)}(x) 
\Big]
= \sum_{n=1}^{N}z_{T}^{(n)}(x_{T}).
\end{align*}
This additive structure holds not only at the terminal time step, which can be proved by the following induction. 
Before starting the induction, we establish the following relation:
\begin{align}\label{eq:sup_z_equal_Z}
z(x) = \mathcal{E}_{\tau}[\regV(x)] = \exp \Big[
-\frac{1}{\tau}(-\tau \ln Z(x))
\Big] = Z(x),
\end{align}
and 
\begin{align}\label{eq:sup_Z_rz}
Z(x) 
&=\int \exp \Big[
-\frac{1}{\tau}(\underbrace{l(x,u)+ \regV(x')}_{\regQ(x,u)})\Big] \mathrm{d}u\\\notag
&=\int \exp \Big[
-\frac{1}{\tau}l(x,u)\Big]
\exp \Big[ 
-\frac{1}{\tau}\regV(x')
\Big] \mathrm{d}u\\\notag
&= \int r(x,u) z(x')\mathrm{d}u,
\end{align}
and therefore,
\begin{equation}\label{eq:z_int_r_z(x')}
z(x) = \int r(x,u) z(x') \mr{d}u,
\end{equation}
for the global approximation. In addition, the relation also holds for the local trajectories:
\begin{equation}
z^{(n)}(x) = \int r^{(n)}(x,u) z^{(n)}(x') \mr{d}u,
\end{equation}
which can be confirmed by performing \eqref{eq:sup_z_equal_Z} and \eqref{eq:sup_Z_rz} with the $n$-th trajectory.

Assume the additive structure holds at time instance $t+1$:
\begin{align*}
z(x') = \sum_{n=1}^{N}z^{(n)}(x'). 
\end{align*}
Substitution of the equation above into \eqref{eq:z_int_r_z(x')}, at time $t$, we get 
\begin{align}\label{eq:sup_additive_z}
\notag
z(x) &= \int \sum_{n=1}^{N}z^{(n)}(x')r(x,u)\mathrm{d}u \\
\notag
&= \sum_{n=1}^{N} \int z^{(n)}(x')r(x,u)\mathrm{d}u \\
&\approx \sum_{n=1}^{N} \int z^{(n)}(x')r^{(n)}(x,u)\mathrm{d}u \\
\notag
&= \sum_{n=1}^{N} z^{(n)}(x).
\end{align}
This implies that the additive structure holds at all time steps by induction.

The approximation in the third line is justified by the local support of the desirability function $z^{(n)}(x)$. When multiplied by $z^{(n)}$, the global reward $r(x,u)$ is effectively filtered, contributing only its local component $r^{(n)}(x,u)$. We note that, if $r(x,u)$ were formally constructed as a sum of local components, this simple additive property would not hold due to cross-terms.  This local-filtering property is the key to the derivation.

\paragraph{Computing optimal control and weights:} In order to use the additive structure in the optimal control distribution, we first write $\pi^{\ast}$ in terms of $r$ and $z$ as
\begin{align*}
\pi^{\ast}(u|x) &= 
Z^{-1}\exp\big[ -\frac{1}{\tau}\regQ(x,u)\big] \\
&= 
Z^{-1}\exp\big[ -\frac{1}{\tau}
\big( l(x,u) + \regV(x')\big)
\big] \\
&= [z(x)]^{-1}r(x,u) z(f(x,u)).
\end{align*}
The $n$-th local policy takes the form of 
\begin{align*}
\pi^{\ast(n)}(u|x) = [z^{(n)}(x)]^{-1}[r^{(n)}(x,u)] z^{(n)}(x').
\end{align*}
Using the additive structure of $z$, and from the two equations above, $\pi^{\ast}$ is written as a weighted sum of $\pi^{\ast (n)}$ s as
\begin{align*}
\pi^{\ast}(u|x) &= [z(x)]^{-1}\Big[ \sum_{n=1}^{N}z^{(n)}(x') \Big] r(x,u)\\
&= \sum_{n=1}^{N} \frac{z^{(n)}(x)}{z(x)}[z^{(n)}(x)]^{-1}z^{(n)}(x')r(x,u)\\
&\approx \sum_{n=1}^{N} \frac{z^{(n)}(x)}{z(x)}[z^{(n)}(x)]^{-1}z^{(n)}(x')[r^{(n)}(x,u)]\\
&= \sum_{n=1}^{N} \omega^{(n)}(x) \pi^{\ast(n)}(u|x),
\end{align*}
with the weight
\begin{align}\label{eq:sup_weight}
\omega^{(n)} &= [z(x)]^{-1}z^{(n)}(x), \quad \text{with}\ 
\sum_{n=1}^{N}\omega^{(n)} =1.
\end{align}

Each policy for the $n$ th trajectory is given by Gaussian distribution as we saw in the unimodal case:
\begin{align*}
\pi^{(n)}(\delta u^{(n)}|\delta x^{(n)}) = \mathcal{N}(\delta u^{(n)}; \delta u^{\ast (n)}, \tau (Q_{uu}^{n})^{-1}).
\end{align*}
Since both $z$ and $\pi$ are a weighted sum of $z^{(n)}$ and $\pi^{(n)}$, the backward pass of ME-DDP is achieved by performing the backward pass in $N$ different nominal trajectories and composing them with weights. To compute the weights, we first represent the desirability function $z$ with the value function $\regV$. From \eqref{eq:sup_V_islog_Z} for each trajectories, we have  
\begin{equation}\label{eq:sup_z^n_with_V^n}
z^{(n)}(x) = \exp \Big[-\frac{1}{\tau}\regV^{(n)}(x)\Big].
\end{equation}
Plugging this $z^{(n)}$ in the right-hand side of  \eqref{eq:sup_additive_z}, and $Z$ in \eqref{eq:sup_V_islog_Z} in the left-hand side of the same equation, we have a similar LogSumExp structure in value function: 
\begin{align*}
\regV(x) = -\tau \ln \sum_{n=1}^{N} \exp \Big[
-\frac{1}{\tau}\regV^{(n)}(x)
\Big].
\end{align*}
Plugging the equation above into \eqref{eq:sup_V_islog_Z} gives
\begin{equation*}
z(x) = \exp \Big[ -\frac{1}{\tau}\regV(x)\Big] = \sum_{n=1}^{N} \exp \Big[
-\frac{1}{\tau}\regV^{(n)}(x)\Big].
\end{equation*}
From the above equation, \eqref{eq:sup_z^n_with_V^n}, and \eqref{eq:sup_weight}, we get
\begin{align*}
\omega_{t}^{(n)} &= z_{t}(x_{t})^{-1}z_{t}^{(n)}(x_{t})=\frac{\exp \big[-\frac{1}{\tau}
(\regV_{t}^{(n)}(x_{t}))
\big]}
{
\sum_{n=1}^{N}\exp \big[
-\frac{1}{\tau}
\regV_{t}^{(n)}(x_{t})
\big]
},
\end{align*}
where we explicitly write $V$ as a function of state and time by recovering the time index in the notation. The value function $\regV_{t}(x_{t})$ can be approximated by the following quadratic approximation and recursion:
\begin{align*}
\tilde{V}_{t}^{(n)}(x) &\approx V_{t}^{(n)}(\bar{x}^{(n)})
+ c_{t}+V^{(n)}_{x,t}(\bar{x}^{(n)})^{\tr} \delta x^{(n)} \\
& \hspace{20mm}  +\frac{1}{2}\delta x^{(n)\tr} V_{xx,t}^{(n)}(\bar{x}^{(n)})\delta x^{(n)},\\
c_{t} &= c_{t+1} + V_{H,t}^{(n)}, 
\end{align*}
with $c_{T}=0, \ t= 1,\dots, T-1$.
that uses derivatives from the backward pass of normal deterministic DDP and $\delta x$ from the forward pass. Although the policy is defined at every time step, independent sampling at each time step $t$ can lead to physically inconsistent control jitter, due to the mode switching. To ensure temporal coherence,  we therefore pick the mode to sample based on the weight at time $t=1$ ($w_{1}$) and keep sampling from the same mode throughout the time horizon.


\subsection{Pseudocode}
We provide the exploration mechanisms for UG-ME-DDP and MG-ME-DDP in Algorithms \ref{alg:UG_Exploration} and \ref{alg:MG_Exploration}, respectively. These exploration mechanisms can be used in the main algorithm in Algorithm \ref{alg:SVDDP_Main}. We note that in the formulation proposed in \cite{so2022maximum}, the number of trajectories was fixed ($N=2$) in UG-ME-DDP. To isolate the sampling density and the property of the policy, here we allow $N \geq 2$.

In UG-ME-DDP, the policy is composed base on the best trajectory. 
In MG-ME-DDP, we need $\batch{V}_{H}=\{V_{H,1:T-1}\}_{n=1}^{N}$ that accounts for the effect of entropy to compose the multimodal policy.  
\begin{algorithm}[hbt!]
\DontPrintSemicolon
\SetAlgoLined
\SetCommentSty{textnormal} 
\caption{Exploration by Unimodal Gaussian Policy}\label{alg:UG_Exploration}
\KwIn{
  \batch{U},
  \batch{X},
  $\batch{Q}_{uu}$,
  \batch{K}, $b$
  }
\KwOut{Updated trajectories \batch{X}, \batch{U}} 
\For{$n = 1$ \KwTo $N, n \neq b$ \textbf{in parallel}}{
        $X^{(n)}, U^{(n)} \gets \text{SampleFromUG}(X^{(b)}, U^{(b)}, \batch{Q}_{uu}^{(b)}, \batch{K}^{(b)}$)\;
    }
\Return $\{\batch{X}, \batch{U}\}$\;
\end{algorithm}

\begin{algorithm}[hbt!]
\DontPrintSemicolon
\SetAlgoLined
\SetCommentSty{textnormal} 
\caption{Exploration by Multimodal Gaussian Policy}\label{alg:MG_Exploration}
\KwIn{
  \batch{U},
  \batch{X},
  $\batch{Q}_{uu}$,
  \batch{K},
  $\batch{V}_{H}$, $b$
  }
\KwOut{Updated trajectories \batch{X}, \batch{U}} 
\For{$n = 1$ \KwTo $N, n \neq b$ \textbf{in parallel}}{
        $X^{(n)}, U^{(n)} \gets \text{SampleFromMG}(\batch{X}, \batch{U}, \batch{Q}_{uu}, \batch{K}, \batch{V}_{H})$\;
    }
\Return $\{\batch{X}, \batch{U}\}$\;
\end{algorithm}

\setcounter{equation}{0}
\setcounter{algocf}{0}
\setcounter{figure}{0}

\renewcommand{\theequation}{\thesection.\arabic{equation}}
\renewcommand{\thealgocf}{\thesection.\arabic{algocf}}
\renewcommand{\thefigure}{\thesection.\arabic{figure}}

\section{Positive Definiteness of Control Hessian}
This section is to provide additional explanation for \nameref{sec:geom_analysis} section.
In this section, we show how $Q_{uu,t}$ can be PD given the following conditions: 
\begin{enumerate}
\item First-order dynamics: The transition dynamics are locally linear, $\delta x_{t+1} = f_{x}\delta x_t + f_{u} \delta u_t$. meaning second-order derivatives of the dynamics ($f_{xx}, f_{uu}, f_{xu}$) are zero. 
\item PD Terminal and running cost: The Hessian of the terminal cost and therefore that of the value function is PD $V_{xx,T} = \Phi_{xx,T} \succ 0$. The Hessians of the running cost are PD  $l_{uu,t}, l_{xx,t} \succ 0$ for $t=1, \dots T-1$.
\item Running cost is decoupled: $l_{xu}=O_{n_{x},n_{u}}$.
\end{enumerate}
We note that this is not a restrictive assumption. Standard design choices, such as quadratic cost or PD Hessian approximation, satisfy these requirements. We note that this is not a restrictive assumption. Standard design choices, such as quadratic or cost with PD Hessian approximation, can satisfy these requirements.

With conditions above, we can show that at time step $T-1$, $Q_{uu, T-1}$ is PD:
\begin{equation} \label{eq:pd_quu_Tm1}
Q_{uu, T-1} = \underbrace{l_{uu, T-1}}_{\succ 0} + \underbrace{f_{u,T-1}^{\tr} V_{xx, T} f_{u,T-1}}_{\succeq 0} \succ 0,
\end{equation}
where the term $f_{u}^{\tr} V_{xx, T} f_{u}$ is at least positive semi-definite (PSD) because $V_{xx,T} \succ 0$.

Assume the Hessian of the value function at time step $t+1$ satisfies $V_{xx, t+1} \succ 0$. We now show by induction that $Q_{uu, t}$ is PD and $V_{xx, t}$ remains PD for all $t$. By this assumption and the condition $l_{uu,t} \succ 0$, the matrix $Q_{uu,t}$ is PD which follows same argument as in~\eqref{eq:pd_quu_Tm1}.


The Riccati recursion for the value function Hessian is:
$$V_{xx,t} = Q_{xx,t} - Q_{xu,t} Q_{uu,t}^{-1} Q_{ux,t},$$
which is obtained by substituting into~\eqref{eq:value_update_with_gain} where $Q_{uu}$  terms in gains cancel out.

A symmetric block matrix $M$ is positive definite if and only if its blocks satisfy certain conditions:
\begin{align*}
&M = \begin{bmatrix} Q_{xx} & Q_{xu} \\ Q_{ux} & Q_{uu} \end{bmatrix} \succ 0 \\
&\iff Q_{uu} \succ 0 \ \
\text{and} \ \ Q_{xx} - Q_{ux}^\top Q_{uu}^{-1} Q_{ux} \succ 0.
\end{align*}
The term $V_{xx} = Q_{xx} - Q_{ux}^\top Q_{uu}^{-1} Q_{ux}$
is the Schur Complement of $Q_{uu}$ in the matrix $M$.
Therefore, to prove $V_{xx,t} \succ 0$,  it is enough to show that joint Hessian of the $Q$-function $M$ is PD.

By constructions in~\eqref{eq:Qexpanded_derivative}, we can decompose $M$ into: $$M = \begin{bmatrix} l_{xx} & l_{xu} \\ l_{ux} & l_{uu} \end{bmatrix} + \begin{bmatrix} f_x \\ f_u \end{bmatrix}^\top V_{xx,t+1} \begin{bmatrix} f_x & f_u \end{bmatrix} \succ 0.$$
From the assumption that $l_{xu} = O_{n_{x},n_{u}}$, $M$ is a sum of PD and PSD matrices and therefore $M \succ 0$.
Therefore, it follows by backward induction over time that $Q_{uu,t}$ and $V_{xx,t}$ are PD for all time steps.

\setcounter{equation}{0}
\setcounter{algocf}{0}
\setcounter{figure}{0}

\renewcommand{\theequation}{\thesection.\arabic{equation}}
\renewcommand{\thealgocf}{\thesection.\arabic{algocf}}
\renewcommand{\thefigure}{\thesection.\arabic{figure}}
\section{Block-diagonal approximation in SVNM}

In this section, we examine the intuition of the approximation employed in SVNM. We first consider the size of the Hessian. Here, We follow the same notation as in Section \nameref{sec:SV dyn_opt}. 

Suppose that we concatenate the particles to get
\begin{align*}
X =[
x_{1}^{\tr}, x_{2}^{\tr},  \dots,  x_{N}]^{\tr}
 \in \mathbb{R}^{Nd}.
\end{align*}
Then, the full Hessian of the KL function is 
\begin{equation*}
H = \nabla^{2}_{X} D_{\rm{KL}} (q||p) \in \mathbb{R}^{Nd \times Nd}.
\end{equation*}
In a block view, it can be written as 
\begin{align*}
H = \begin{bmatrix}
H^{1,1} & H^{1,2} & \cdots &H^{1,N} \\
H^{2,1} & H^{2,2} & \cdots &H^{2,N} \\
\vdots & \vdots & \ddots & \vdots \\
H^{N,1} & H^{N,2} & \cdots & H^{N,N}
\end{bmatrix}, 
\end{align*}
with
\begin{align*}H^{s,n} = \frac{\partial^{2} {D_{\rm{KL}}}}{\partial x_{s} \partial x_{n}}  \in \mathbb{R}^{d\times d}
\end{align*}
and $s, n=1 \dots N$. Here, each block $H^{s,n}$ encodes how the update of  particle $s$ depends on the particle $n$. The diagonal blocks are the local curvature of the KL functional associated with particle $x_{s}$, ignoring interactions with other particles.
There are $N^{2}$ blocks each of which has $d^{2}$ entries.

In expressions such as \eqref{eq:SV_Newton_hessian}, the term 
$h_{ij}(x_{s}, x_{n})$ is $i,j$ element of block $H^{s,l}$:
\begin{equation*}
H^{s,n}
_{i,j} = h_{ij}(x_{s}, x_{n}), \quad i,j = 1, \cdots, d. 
\end{equation*}

To understand the effect of the block-diagonal approximation intuitively, we consider a simple case with $N=2$. Then, the equation for the Newton direction is:
\begin{equation*}
\begin{bmatrix}
H^{1,1} & H^{1,2} \\ H^{2,1} & H^{2,2}
\end{bmatrix}
\begin{bmatrix}
\beta^{1}\\
\beta^{2}
\end{bmatrix}
= -
\begin{bmatrix}
\nabla \hat{J}[\bm{0}](x_{1}) \\
\nabla \hat{J}[\bm{0}](x_{2})
\end{bmatrix}.
\end{equation*}
Now, by focusing on the resulting two equations, we can see that they can be written as 
\begin{align*}
\sum_{n=1}^{N}H^{s,n} \beta^{n} = - \nabla \hat{J}[\bm{0}](x_{s}) ,\quad s= 1, \dots, N.
\end{align*}
The update direction $\beta^{s}$ of particle $s$ depends on the update of all other particles $\beta^{n}$. The only way to solve the system of equations is to invert the large $H$ matrix, which is expensive. An alternative performed in the approximation is to keep the terms only with $n=s$ in the summation, which yields
\begin{equation*}
H^{s,s}\beta^{s} = - \nabla \hat{J}[\bm{0}](x_{s}), \quad s = 1 ,\dots, N,
\end{equation*}
in the previous simple example with $N=2$, the system becomes
\begin{equation*}
\begin{bmatrix}
H^{1,1} & O_{d}\\ O_{d} & H^{2,2}  
\end{bmatrix}
\begin{bmatrix}
\beta^{1}\\
\beta^{2}
\end{bmatrix}
= -\begin{bmatrix}
\nabla \hat{J}[\bm{0}](x_{1}) \\
\nabla \hat{J}[\bm{0}](x_{2})
\end{bmatrix}\\
\end{equation*}
Under this approximation, instead of inverting $H\in\mathbb{R}^{Nd\times Nd}$, we only need to invert the $N$ diagonal blocks $H^{s,s} \in \mathbb{R}^{d\times d}$ independently, allowing the $\beta^{s}$ to be computed in parallel. With these, we recover the direction $w(x_{s})$ by 
\begin{equation*}
w(x_{s}) = \sum_{n=1}^{N}\beta ^{s} k(x_{n}, x_{s}),
\end{equation*}
which coincides with \eqref{eq:SVNM_diag_direction}.

\setcounter{equation}{0}
\setcounter{algocf}{0}
\setcounter{figure}{0}

\renewcommand{\theequation}{\thesection.\arabic{equation}}
\renewcommand{\thealgocf}{\thesection.\arabic{algocf}}
\renewcommand{\thefigure}{\thesection.\arabic{figure}}
\section{DDP and Barrier Function}
This section provides how the relaxed-$log$ barrier function modifies DDP. Here, we use the concatenated state and control as $y_{t}=[x_{t}^{\tr}, u_{t}^{\tr}]^{\tr}$. Since the barrier term is a part of cost, it modifies the $Q$ function of the standard DDP, yielding $\hat{Q}$, whose derivatives are given as:
\begin{align*}
    \hat{Q}_{y} = Q_{y} + \nabla_{y} \cB_{\mr{r}}(g(y)), \quad 
    \hat{Q}_{yy} = Q_{yy} + \nabla_{y}^{2}\cB_{\mr{r}}(g(y)).
\end{align*}
The Hessian of the barrier function is given as follows: 
\begin{align*}
&\nabla_y^2 \mathcal{B}_{\mr{r},i}(g_i; \mu)\\ =& 
\begin{cases} 
\dfrac{\mu}{g_i^2} \nabla g_i(\nabla g_i)^\tr - \dfrac{\mu}{g_i} \nabla^2 g_i, & g_i \leq -\delta_{\mu}, \\
\dfrac{\mu}{\delta_{\mu}^2} \nabla g_i(\nabla g_i)^\tr + \dfrac{\mu (g_i + 2\delta_{\mu})}{\delta_{\mu}^2} \nabla^2 g_i, & g_i > -\delta_{\mu}.
\end{cases}
\end{align*}
To keep the Hessian well-conditioned, we drop the second-order terms ($\nabla^{2} g_{i}$), which can be negative definite.

\setcounter{equation}{0}
\setcounter{algocf}{0}
\setcounter{figure}{0}

\renewcommand{\theequation}{\thesection.\arabic{equation}}
\renewcommand{\thealgocf}{\thesection.\arabic{algocf}}
\renewcommand{\thefigure}{\thesection.\arabic{figure}}

\section{Numerical Sensitivity of Relaxed-Barrier Function}
This section provides a detailed analysis of the numerical limitations of the relaxed-barrier formulation introduced in Section \nameref{sec:geom_analysis}. We identify two primary sources of sensitivity: the inherent scaling of the $\log$ barrier gradient and the hard branching mechanism introduced by the relaxation.

\paragraph{Scaling of Gradient}
The numerical sensitivity arises from the inherent scaling of the gradient near the boundary of the constraints $g(x) \approx 0$ in the exact $\log$ barrier and a small switching point $\delta_{\mu}$ in the relaxed one. 
Consider the gradient of the relaxed barrier function $\cB_{\rm{r}}(x)$ provided in \eqref{eq:def_relaxed_log}:

\[
\nabla \cB_{\rm{r}}(x) = 
\begin{cases}
-\;\dfrac{\nabla g(x)}{g(x)}, \quad & g(x) \leq -\delta_{\rm{\mu}},\\
\left( \dfrac{g(x) + 2\delta_{\rm{\mu}}}{\delta_{\mu}^{2}} \right) \nabla g(x), \quad & g(x) > -\delta_{\rm{\mu}}.
\end{cases}
\]
This scaling amplifies small numerical discrepancies. Suppose that we execute the algorithm across different computational environments (e.g., CPU vs. GPU or different hardware architectures). We denote values computed on different devices with superscripts $a$ and $b$. Small differences in $\nabla g(x)$, from device-specific optimizations or floating-point rounding errors, are amplified. For two devices with slightly different gradients $\nabla g^a(x)$ and $\nabla g^b(x) = \nabla g^a(x) + \epsilon$, the discrepancy in the gradient around the boundary $g(x) \approx -\delta_{\mu}$ satisfies:
\begin{align*}
&\|\nabla \cB_{\rm{r}}^{a}(x) - \nabla \cB_{\rm{r}}^{b}(x)\| \\
\approx&
\begin{cases}
\dfrac{\|\nabla g^{a}(x) - \nabla g^{b}(x)\|}{|g(x)|} = \dfrac{\epsilon}{|g(x)|}\approx \dfrac{\epsilon}{\delta_{\mu}}, & g(x) \leq -\delta_{\mu},\\\\
\dfrac{g(x) + 2\delta_{\mu}}{\delta_{\mu}^{2}} \norm{\nabla g_{1}(x)-\nabla g_{2}(x) } \approx \dfrac{\epsilon}{\delta_{\mu}}, & g(x) > -\delta_{\mu}.\\
\end{cases}
\end{align*}
For example, if $\epsilon = 10^{-11}$ and the relaxation parameter is small (e.g., $\delta_{\mu} = 10^{-6}$), the discrepancy is amplified to the order of $10^{-5}$. These errors accumulate through the forward and backward passes of the DDP, ultimately showing up as difference in trajectories. While a small $\delta_{\mu}$ is often necessary for strict constraint satisfaction, especially in MPC, it introduces a source of device-dependent sensitivity.
We note that this is not unique to the relaxed formulation. The exact barrier also has the same issue near the boundary of the constraints.

\paragraph{Hard Branching (Unique to the relaxed formulation.)}
Relaxed $\log$-barrier functions switch from the standard $\log$ to a quadratic approximation at $g(x)=-\delta_{\mu}$. This introduces a hard branch, that is, the code must choose which branch to evaluate for each state. Even if the formulation is $\mathcal{C}^2$ continuous at the switching point, a small floating-point difference from, for example, order of operations, can make one evaluation select the $\log$ branch while another selects the quadratic branch. This results in slightly different derivatives that propagate and accumulate through DDP, as before. The effect is more pronounced for very small $\delta_{\mu}$, where even small deviations can create a large difference around $\delta_{\mu}$.

\bibliographystyle{SageH_no_url}

\begin{thebibliography}{87}
\providecommand{\natexlab}[1]{#1}
\providecommand{\url}[1]{\texttt{#1}}
\providecommand{\urlprefix}{URL }
\expandafter\ifx\csname urlstyle\endcsname\relax
  \providecommand{\doi}[1]{DOI:\discretionary{}{}{}#1}\else
  \providecommand{\doi}{DOI:\discretionary{}{}{}\begingroup \urlstyle{rm}\Url}\fi

\bibitem[{Aguiar et~al.(2017)Aguiar, Bayer, Hauser, H{\"a}usler, Notarstefano, Pascoal, Rucco and Saccon}]{Aguiar2017PRONTO}
Aguiar AP, Bayer FA, Hauser J, H{\"a}usler AJ, Notarstefano G, Pascoal AM, Rucco A and Saccon A (2017) \emph{Constrained Optimal Motion Planning for Autonomous Vehicles Using PRONTO}.
\newblock Cham: Springer International Publishing.
\newblock ISBN 978-3-319-55372-6, pp. 207--226.
\newblock \doi{10.1007/978-3-319-55372-6_10}.

\bibitem[{Aoyama et~al.(2025)Aoyama, Lehmann and Theodorou}]{Aoyama2025SVDDP}
Aoyama Y, Lehmann P and Theodorou EA (2025) Second-order stein variational dynamic optimization.
\newblock In: \emph{2025 IEEE International Conference on Robotics and Automation (ICRA)}. pp. 7209--7216.
\newblock \doi{10.1109/ICRA55743.2025.11128381}.

\bibitem[{Aoyama and Theodorou(2024)}]{aoyama2024tsallis_me_ddp}
Aoyama Y and Theodorou EA (2024) Generalized maximum entropy differential dynamic programming.
\newblock In: \emph{2024 IEEE 63rd Conference on Decision and Control (CDC)}. pp. 8825--8831.
\newblock \doi{10.1109/CDC56724.2024.10886519}.

\bibitem[{Barcelos et~al.(2024)Barcelos, Lai, Oliveira, Borges and Ramos}]{Barcelos2024Pathsignature}
Barcelos L, Lai T, Oliveira R, Borges P and Ramos F (2024) Path signatures for diversity in probabilistic trajectory optimisation.
\newblock \emph{The International Journal of Robotics Research} 43(11): 1693--1710.
\newblock \doi{10.1177/02783649241233300}.

\bibitem[{Betts(2001)}]{Betts2001OptCtrl}
Betts JT (2001) \emph{Practical Methods for Optimal Control Using Nonlinear Programming}.
\newblock Philadelphia, PA: SIAM.

\bibitem[{Boutselis and Theodorou(2021)}]{Boutselis2021DDPLiegroup}
Boutselis GI and Theodorou E (2021) Discrete-time differential dynamic programming on lie groups: Derivation, convergence analysis, and numerical results.
\newblock \emph{IEEE Transactions on Automatic Control} 66(10): 4636--4651.
\newblock \doi{10.1109/TAC.2020.3034206}.

\bibitem[{Bradbury et~al.(2018)Bradbury, Frostig, Hawkins, Johnson, Katariya, Leary, Maclaurin, Necula, Paszke, Vander{P}las, Wanderman-{M}ilne and Zhang}]{jax2018github}
Bradbury J, Frostig R, Hawkins P, Johnson MJ, Katariya Y, Leary C, Maclaurin D, Necula G, Paszke A, Vander{P}las J, Wanderman-{M}ilne S and Zhang Q (2018) {JAX}: composable transformations of {P}ython+{N}um{P}y programs.

\bibitem[{Caluwaerts et~al.(2023)Caluwaerts, Iscen, Kew, Yu, Zhang, Freeman, Lee, Lee, Saliceti, Zhuang, Batchelor, Bohez, Casarini, Chen, Cortes, Coumans, Dostmohamed, Dulac-Arnold, Escontrela, Frey, Hafner, Jain, Jyenis, Kuang, Lee, Luu, Nachum, Oslund, Powell, Reyes, Romano, Sadeghi, Sloat, Tabanpour, Zheng, Neunert, Hadsell, Heess, Nori, Seto, Parada, Sindhwani, Vanhoucke and Tan}]{google2023barkour}
Caluwaerts K, Iscen A, Kew JC, Yu W, Zhang T, Freeman D, Lee KH, Lee L, Saliceti S, Zhuang V, Batchelor N, Bohez S, Casarini F, Chen JE, Cortes O, Coumans E, Dostmohamed A, Dulac-Arnold G, Escontrela A, Frey E, Hafner R, Jain D, Jyenis B, Kuang Y, Lee E, Luu L, Nachum O, Oslund K, Powell J, Reyes D, Romano F, Sadeghi F, Sloat R, Tabanpour B, Zheng D, Neunert M, Hadsell R, Heess N, Nori F, Seto J, Parada C, Sindhwani V, Vanhoucke V and Tan J (2023) Barkour: Benchmarking animal-level agility with quadruped robots.

\bibitem[{Carvalho et~al.(2023)Carvalho, Le, Baierl, Koert and Peters}]{Carvalho2023Diffusionrobot}
Carvalho J, Le AT, Baierl M, Koert D and Peters J (2023) Motion planning diffusion: Learning and planning of robot motions with diffusion models.
\newblock In: \emph{2023 IEEE/RSJ International Conference on Intelligent Robots and Systems (IROS)}. pp. 1916--1923.
\newblock \doi{10.1109/IROS55552.2023.10342382}.

\bibitem[{Chen and Ghattas(2020)}]{Chen2020ProjectedSVGD}
Chen P and Ghattas O (2020) Projected stein variational gradient descent.
\newblock In: Larochelle H, Ranzato M, Hadsell R, Balcan M and Lin H (eds.) \emph{Advances in Neural Information Processing Systems}, volume~33. Curran Associates, Inc., pp. 1947--1958.

\bibitem[{Chen et~al.(2019)Chen, Wu, Chen, O\textquotesingle Leary-Roseberry and Ghattas}]{Chen2019ProjectedSVNewton}
Chen P, Wu K, Chen J, O\textquotesingle Leary-Roseberry T and Ghattas O (2019) Projected stein variational newton: A fast and scalable bayesian inference method in high dimensions.
\newblock In: Wallach H, Larochelle H, Beygelzimer A, d\textquotesingle Alch\'{e}-Buc F, Fox E and Garnett R (eds.) \emph{Advances in Neural Information Processing Systems}, volume~32. Curran Associates, Inc.

\bibitem[{Cho(2005)}]{CHO2005BarrierStoch}
Cho GM (2005) Log-barrier method for two-stage quadratic stochastic programming.
\newblock \emph{Applied Mathematics and Computation} 164(1): 45--69.
\newblock \doi{https://doi.org/10.1016/j.amc.2004.04.095}.

\bibitem[{Deits and Tedrake(2015)}]{deits2015computing}
Deits R and Tedrake R (2015) Computing large convex regions of obstacle-free space for robotic navigation.
\newblock \emph{The International Journal of Robotics Research} 34(7): 955--977.

\bibitem[{Detommaso et~al.(2018)Detommaso, Cui, Marzouk, Spantini and Scheichl}]{detommaso2018steinNewton}
Detommaso G, Cui T, Marzouk Y, Spantini A and Scheichl R (2018) A stein variational newton method.
\newblock In: Bengio S, Wallach H, Larochelle H, Grauman K, Cesa-Bianchi N and Garnett R (eds.) \emph{Advances in Neural Information Processing Systems}, volume~31. Curran Associates, Inc.

\bibitem[{Dong and Tong(2021)}]{Dong2021Replicaexchange}
Dong J and Tong XT (2021) Replica exchange for non-convex optimization.
\newblock \emph{J. Mach. Learn. Res.} 22(1).

\bibitem[{Feller and Ebenbauer(2017{\natexlab{a}})}]{Feller2017BarrierLinSys}
Feller C and Ebenbauer C (2017{\natexlab{a}}) Relaxed logarithmic barrier function based model predictive control of linear systems.
\newblock \emph{IEEE Transactions on Automatic Control} 62(3): 1223--1238.
\newblock \doi{10.1109/TAC.2016.2582040}.

\bibitem[{Feller and Ebenbauer(2017{\natexlab{b}})}]{Feller2017BarrierLinSysStable}
Feller C and Ebenbauer C (2017{\natexlab{b}}) A stabilizing iteration scheme for model predictive control based on relaxed barrier functions.
\newblock \emph{Automatica} 80: 328--339.
\newblock \doi{https://doi.org/10.1016/j.automatica.2017.02.001}.

\bibitem[{Fiacco and McCormick(1968)}]{fiacco1968nonlinearbarrier}
Fiacco AV and McCormick GP (1968) \emph{Nonlinear Programming: Sequential Unconstrained Minimization Techniques}.
\newblock John Wiley and Sons, New York.

\bibitem[{Freeman et~al.(2021)Freeman, Frey, Raichuk, Girgin, Mordatch and Bachem}]{brax2021github}
Freeman CD, Frey E, Raichuk A, Girgin S, Mordatch I and Bachem O (2021) Brax - a differentiable physics engine for large scale rigid body simulation.

\bibitem[{Gandhi et~al.(2021)Gandhi, Vlahov, Gibson, Williams and Theodorou}]{RobustMPPI}
Gandhi MS, Vlahov B, Gibson J, Williams G and Theodorou EA (2021) Robust model predictive path integral control: Analysis and performance guarantees.
\newblock \emph{IEEE Robotics and Automation Letters} 6(2): 1423--1430.
\newblock \doi{10.1109/LRA.2021.3057563}.

\bibitem[{Garreau et~al.(2018)Garreau, Jitkrittum and Kanagawa}]{garreau2018large_median_heuristic}
Garreau D, Jitkrittum W and Kanagawa M (2018) Large sample analysis of the median heuristic.
\newblock \emph{arXIv preprint arXiv:1707.07269} .

\bibitem[{Gill et~al.(2000)Gill, Jay, Leonard, Petzold and Sharma}]{Gill2000SQPdynamical}
Gill PE, Jay LO, Leonard MW, Petzold LR and Sharma V (2000) An sqp method for the optimal control of large-scale dynamical systems.
\newblock \emph{Journal of Computational and Applied Mathematics} 120(1): 197--213.
\newblock \doi{https://doi.org/10.1016/S0377-0427(00)00310-1}.

\bibitem[{Gill et~al.(2002)Gill, Murray and Saunders}]{Gill2002SNOPT}
Gill PE, Murray W and Saunders MA (2002) Snopt: An sqp algorithm for large-scale constrained optimization.
\newblock \emph{SIAM Journal on Optimization} 12(4): 979--1006.
\newblock \doi{10.1137/S1052623499350013}.

\bibitem[{Gorham and Mackey(2017)}]{Gorham2017IQkernel}
Gorham J and Mackey L (2017) Measuring sample quality with kernels.
\newblock In: \emph{Proceedings of the 34th International Conference on Machine Learning}, volume~70. pp. 1292--1301.

\bibitem[{Grandia et~al.(2019)Grandia, Farshidian, Ranftl and Hutter}]{Grandia2019BarrierDDP}
Grandia R, Farshidian F, Ranftl R and Hutter M (2019) Feedback mpc for torque-controlled legged robots.
\newblock In: \emph{2019 IEEE/RSJ International Conference on Intelligent Robots and Systems (IROS)}. IEEE Press, p. 4730–4737.
\newblock \doi{10.1109/IROS40897.2019.8968251}.

\bibitem[{Harford(2016)}]{harford2016messy}
Harford T (2016) \emph{Messy: How to Be Creative and Resilient in a Tidy-Minded World}.
\newblock Little, Brown Book Group.
\newblock ISBN 9781408706770.

\bibitem[{Hauser(2002)}]{HAUSER2002prototypePRONTO}
Hauser J (2002) A projection operator approach to the optimization of trajectory functionals.
\newblock \emph{IFAC Proceedings Volumes} 35(1): 377--382.
\newblock \doi{https://doi.org/10.3182/20020721-6-ES-1901.00312}.
\newblock 15th IFAC World Congress.

\bibitem[{Hauser and Saccon(2006)}]{Hauser2006CDCbarrierfunc}
Hauser J and Saccon A (2006) A barrier function method for the optimization of trajectory functionals with constraints.
\newblock In: \emph{Proceedings of the 45th IEEE Conference on Decision and Control}. pp. 864--869.
\newblock \doi{10.1109/CDC.2006.377331}.

\bibitem[{Hestenes(1969)}]{Hestenes1969MultiplierAG}
Hestenes MR (1969) Multiplier and gradient methods.
\newblock \emph{Journal of Optimization Theory and Applications} 4: 303--320.
\newblock \doi{10.1007/BF00927673}.

\bibitem[{Huang et~al.(2024)Huang, Sundaralingam, Mousavian, Murali, Goldberg and Fox}]{huang2024diffusionseeder}
Huang H, Sundaralingam B, Mousavian A, Murali A, Goldberg K and Fox D (2024) Diffusionseeder: Seeding motion optimization with diffusion for rapid motion planning.
\newblock In: \emph{Proceedings of the 8th Annual Conference on Robot Learning (CoRL)}. Munich, Germany.

\bibitem[{Jacobson and Mayne(1970)}]{Jacobson1970ddp}
Jacobson DH and Mayne DQ (1970) \emph{Differential dynamic programming}.
\newblock Elsevier.
\newblock ISBN 0-444-00070-4.

\bibitem[{Jr. and Ho(1975)}]{BrysonHo1975AppliedOptCtrl}
Jr AEB and Ho YC (1975) \emph{Applied Optimal Control: Optimization, Estimation, and Control}.
\newblock Washington, DC: Hemisphere Publishing Corporation.

\bibitem[{Kalakrishnan et~al.(2011)Kalakrishnan, Chitta, Theodorou, Pastor and Schaal}]{kalakrishnan2011stomp}
Kalakrishnan M, Chitta S, Theodorou E, Pastor P and Schaal S (2011) Stomp: Stochastic trajectory optimization for motion planning.
\newblock In: \emph{2011 IEEE international conference on robotics and automation}. IEEE, pp. 4569--4574.

\bibitem[{Kaufman(1964)}]{Pontryagin1964mathematical}
Kaufman H (1964) The mathematical theory of optimal processes, by l. s. pontryagin, v. g. boltyanskii, r. v. gamkrelidze, and e. f. mishchenko. authorized translation from the russian. translator: K. n. trirogoff, editor: L.. w. neustadt. interscience publishers (division of john wiley and sons, inc. , new york) 1962. viii + 360 pages.
\newblock \emph{Canadian Mathematical Bulletin} 7(3): 500–500.
\newblock \doi{10.1017/S0008439500032112}.

\bibitem[{Kim et~al.(2025)Kim, Jung, Hong and Kim}]{10960717}
Kim MG, Jung M, Hong J and Kim KKK (2025) Mppi-ipddp: A hybrid method of collision-free smooth trajectory generation for autonomous robots.
\newblock \emph{IEEE Transactions on Industrial Informatics} 21(7): 5037--5046.
\newblock \doi{10.1109/TII.2024.3507940}.

\bibitem[{Kuindersma et~al.(2016)Kuindersma, Deits, Fallon, Valenzuela, Dai, Permenter, Cooley and Tedrake}]{Kuindersma2016OptAtlas}
Kuindersma S, Deits RF, Fallon M, Valenzuela A, Dai H, Permenter F, Cooley T and Tedrake R (2016) Optimization-based locomotion planning, estimation, and control design for the atlas humanoid robot.
\newblock \emph{Autonomous Robots} 40(3): 429--455.

\bibitem[{Lambert and Boots(2021)}]{lambert2021entropyregularizedmotionplanning}
Lambert A and Boots B (2021) Entropy regularized motion planning via stein variational inference.

\bibitem[{Lambert et~al.(2021)Lambert, Ramos, Boots, Fox and Fishman}]{lambdert2021SVMPC}
Lambert A, Ramos F, Boots B, Fox D and Fishman A (2021) Stein variational model predictive control.
\newblock In: Kober J, Ramos F and Tomlin C (eds.) \emph{Proceedings of the 2020 Conference on Robot Learning}, \emph{Proceedings of Machine Learning Research}, volume 155. PMLR, pp. 1278--1297.

\bibitem[{Le et~al.(2023)Le, Chalvatzaki, Biess and Peters}]{le2023acceleratingOT}
Le AT, Chalvatzaki G, Biess A and Peters J (2023) Accelerating motion planning via optimal transport.
\newblock In: \emph{Advances in Neural Information Processing Systems (NeurIPS)}.

\bibitem[{Lee et~al.(2024)Lee, Ramos, Lerch and Abraham}]{Lee2024SVErgodic}
Lee D, Ramos F, Lerch C and Abraham I (2024) Stein variational ergodic search.
\newblock In: \emph{Proceedings of Robotics: Science and Systems (RSS)}. Delft, Netherlands.

\bibitem[{Lee et~al.(2025)Lee, Li, Huang, Heiden, Jatavallabhula, Damken, Smith, Nowrouzezahrai, Ramos and Shkurti}]{Lee2025SVGD_STAMP}
Lee Y, Li AZ, Huang P, Heiden E, Jatavallabhula KM, Damken F, Smith K, Nowrouzezahrai D, Ramos F and Shkurti F (2025) Stamp: Differentiable task and motion planning via stein variational gradient descent.
\newblock \emph{IEEE Robotics and Automation Letters} 10(6): 6007--6014.
\newblock \doi{10.1109/LRA.2025.3561575}.

\bibitem[{Levine(2018)}]{levine2018reinforcement}
Levine S (2018) Reinforcement learning and control as probabilistic inference: Tutorial and review.
\newblock \emph{arXiv preprint arXiv:1805.00909} .

\bibitem[{Levine and Koltun(2013)}]{Levine2016GuidedPolicy}
Levine S and Koltun V (2013) Guided policy search.
\newblock In: Dasgupta S and McAllester D (eds.) \emph{Proceedings of the 30th International Conference on Machine Learning}, \emph{Proceedings of Machine Learning Research}, volume~28. Atlanta, Georgia, USA: PMLR, pp. 1--9.

\bibitem[{Li and Todorov(2004)}]{Li2004iLQR}
Li W and Todorov E (2004) Iterative linear quadratic regulator design for nonlinear biological movement systems.
\newblock In: \emph{Proceedings of the 1st International Conference on Informatics in Control, Automation and Robotics (ICINCO)}, volume~1. Set{\'u}bal, Portugal, pp. 222--229.

\bibitem[{Liao(1996)}]{liao1996global}
Liao LZ (1996) Global convergence of differential dynamic programming and newton's method for discrete-time optimal control.
\newblock Technical report, Department of Mathematics, Hong Kong Baptist University, Kowloon, Hong Kong.

\bibitem[{Liao and Shoemaker(1991)}]{liao1991convergence}
Liao LZ and Shoemaker C (1991) Convergence in unconstrained discrete-time differential dynamic programming.
\newblock \emph{IEEE Transactions on Automatic Control} 36(6): 692--706.
\newblock \doi{10.1109/9.86943}.

\bibitem[{Liu(2017)}]{Liu2017SteinGDFLOW}
Liu Q (2017) Stein variational gradient descent as gradient flow.
\newblock In: Guyon I, Luxburg UV, Bengio S, Wallach H, Fergus R, Vishwanathan S and Garnett R (eds.) \emph{Advances in Neural Information Processing Systems}, volume~30. Curran Associates, Inc.

\bibitem[{Liu and Wang(2016)}]{Liu2016SVGD}
Liu Q and Wang D (2016) Stein variational gradient descent: A general purpose variational inference algorithm.
\newblock In: \emph{Advances in Neural Information Processing Systems (NeurIPS)}, volume~29.

\bibitem[{Liu et~al.(2017)Liu, Ramachandran, Liu and Peng}]{Yang2017SVpolicygrad}
Liu Y, Ramachandran P, Liu Q and Peng J (2017) Stein variational policy gradient.
\newblock In: Elidan G, Kersting K and Ihler A (eds.) \emph{Proceedings of the Thirty-Third Conference on Uncertainty in Artificial Intelligence, {UAI} 2017, Sydney, Australia, August 11-15, 2017}. {AUAI} Press.

\bibitem[{Luukkonen(2011)}]{Luukkonen2011}
Luukkonen T (2011) Modelling and control of quadcopter.
\newblock Technical report, Aalto University, Independent Research Project in Applied Mathematics, Espoo, Finland.

\bibitem[{Meier et~al.(2015)Meier, Honegger and Pollefeys}]{Meier2015px4}
Meier L, Honegger D and Pollefeys M (2015) Px4: A node-based multithreaded open source robotics framework for deeply embedded platforms.
\newblock In: \emph{2015 IEEE International Conference on Robotics and Automation (ICRA)}. pp. 6235--6240.
\newblock \doi{10.1109/ICRA.2015.7140074}.

\bibitem[{{Microsoft}(2021)}]{nni2021}
{Microsoft} (2021) {Neural Network Intelligence}.

\bibitem[{Miura et~al.(2024)Miura, Akai, Honda and Hara}]{Miura2024SVMPPISpline}
Miura T, Akai N, Honda K and Hara S (2024) Spline-interpolated model predictive path integral control with stein variational inference for reactive navigation.
\newblock In: \emph{2024 IEEE International Conference on Robotics and Automation (ICRA)}. pp. 13171--13177.
\newblock \doi{10.1109/ICRA57147.2024.10610501}.

\bibitem[{Murray and Wright(1994)}]{MurrayLinesearch1994Barrier}
Murray W and Wright MH (1994) Line search procedures for the logarithmic barrier function.
\newblock \emph{SIAM Journal on Optimization} 4(2): 229--246.
\newblock \doi{10.1137/0804013}.

\bibitem[{Nocedal and Wright(2006)}]{Nocedal2006numerical}
Nocedal J and Wright S (2006) \emph{Numerical optimization}.
\newblock Springer Science \& Business Media.

\bibitem[{Osa(2020)}]{Osa2020MultiModal}
Osa T (2020) Multimodal trajectory optimization for motion planning.
\newblock \emph{The International Journal of Robotics Research} 39: 027836492091829.
\newblock \doi{10.1177/0278364920918296}.

\bibitem[{O’Neill and Wright(2020)}]{ONeil2020NewtonCGBarrier}
O’Neill M and Wright SJ (2020) A log-barrier newton-cg method for bound constrained optimization with complexity guarantees.
\newblock \emph{IMA Journal of Numerical Analysis} 41(1): 84--121.
\newblock \doi{10.1093/imanum/drz074}.

\bibitem[{Pan et~al.(2024)Pan, Yi, Shi and Qu}]{pan2024mbd}
Pan C, Yi Z, Shi G and Qu G (2024) Model-based diffusion for trajectory optimization.
\newblock In: \emph{Advances in Neural Information Processing Systems}, volume~37.

\bibitem[{Powell(1969)}]{powell1969method}
Powell MJD (1969) A method for nonlinear constraints in minimization problems.
\newblock In: Fletcher R (ed.) \emph{Optimization}. New York, NY: Academic Press, pp. 283--298.

\bibitem[{Power and Berenson(2024)}]{power2024constSV}
Power T and Berenson D (2024) Constrained stein variational trajectory optimization.
\newblock \emph{IEEE Transactions on Robotics} 40: 3602--3619.
\newblock \doi{10.1109/TRO.2024.3428428}.

\bibitem[{Puterman(1994)}]{puterman1994markov}
Puterman ML (1994) \emph{Markov Decision Processes: Discrete Stochastic Dynamic Programming}.
\newblock Wiley Series in Probability and Statistics. New York, NY, USA: John Wiley \& Sons, Inc.
\newblock ISBN 978-0471619772.

\bibitem[{Rao et~al.(1998)Rao, Wright and Rawlings}]{Rao1998IPMPC}
Rao C, Wright S and Rawlings J (1998) Application of interior-point methods to model predictive control.
\newblock \emph{Journal of Optimization Theory and Applications} 99(3): 723--757.
\newblock \doi{10.1023/A:1021711402723}.

\bibitem[{Rockafellar(1973)}]{rockafellar1973multiplier}
Rockafellar RT (1973) The multiplier method of {H}estenes and {P}owell applied to convex programming.
\newblock \emph{Journal of Optimization Theory and Applications} 12(6): 555--562.

\bibitem[{Schulman et~al.(2014)Schulman, Duan, Ho, Lee, Awwal, Bradlow, Pan, Patil, Goldberg and Abbeel}]{Schulman2014motion}
Schulman J, Duan Y, Ho J, Lee AX, Awwal I, Bradlow H, Pan J, Patil S, Goldberg K and Abbeel P (2014) Motion planning with sequential convex optimization and convex collision checking.
\newblock \emph{The International Journal of Robotics Research} 33(9): 1251--1270.
\newblock \doi{10.1177/0278364914528132}.

\bibitem[{Sleiman et~al.(2021)Sleiman, Farshidian and Hutter}]{Sleiman2021ETHDDPstyle}
Sleiman JP, Farshidian F and Hutter M (2021) Constraint handling in continuous-time {DDP}-based model predictive control.
\newblock In: \emph{2021 IEEE International Conference on Robotics and Automation (ICRA)}. pp. 8209--8215.
\newblock \doi{10.1109/ICRA48506.2021.9560795}.

\bibitem[{So et~al.(2022)So, Wang and Theodorou}]{so2022maximum}
So O, Wang Z and Theodorou EA (2022) Maximum entropy differential dynamic programming.
\newblock In: \emph{2022 International Conference on Robotics and Automation (ICRA)}. pp. 3422--3428.
\newblock \doi{10.1109/ICRA46639.2022.9812228}.

\bibitem[{Sundaralingam et~al.(2023)Sundaralingam, Hari, Fishman, Garrett, Van~Wyk, Blukis, Millane, Oleynikova, Handa, Ramos, Ratliff and Fox}]{Sundaralingam2023curobo}
Sundaralingam B, Hari SKS, Fishman A, Garrett C, Van~Wyk K, Blukis V, Millane A, Oleynikova H, Handa A, Ramos F, Ratliff N and Fox D (2023) Curobo: Parallelized collision-free robot motion generation.
\newblock In: \emph{2023 IEEE International Conference on Robotics and Automation (ICRA)}. pp. 8112--8119.
\newblock \doi{10.1109/ICRA48891.2023.10160765}.

\bibitem[{Tassa et~al.(2012)Tassa, Erez and Todorov}]{TassaDDP2012}
Tassa Y, Erez T and Todorov E (2012) Synthesis and stabilization of complex behaviors through online trajectory optimization.
\newblock In: \emph{2012 IEEE/RSJ International Conference on Intelligent Robots and Systems}. IEEE, pp. 4906--4913.
\newblock \doi{10.1109/IROS.2012.6386025}.

\bibitem[{Tassa et~al.(2014)Tassa, Mansard and Todorov}]{Tassa2014controllimited}
Tassa Y, Mansard N and Todorov E (2014) Control-limited differential dynamic programming.
\newblock In: \emph{2014 IEEE International Conference on Robotics and Automation (ICRA)}. IEEE, pp. 1168--1175.
\newblock \doi{10.1109/ICRA.2014.6907001}.

\bibitem[{Trivedi et~al.(2014)Trivedi, Wang, Kpotufe and Shakhnarovich}]{Trivedi2014expectoutergradient}
Trivedi S, Wang J, Kpotufe S and Shakhnarovich G (2014) A consistent estimator of the expected gradient outerproduct.
\newblock In: \emph{Proceedings of the Thirtieth Conference on Uncertainty in Artificial Intelligence}, UAI'14. Arlington, Virginia, USA: AUAI Press.
\newblock ISBN 9780974903910, p. 819–828.

\bibitem[{Urain et~al.(2022)Urain, Le, Lambert, Chalvatzaki, Boots and Peters}]{urain2022learaningimplicitprior}
Urain J, Le AT, Lambert A, Chalvatzaki G, Boots B and Peters J (2022) Learning implicit priors for motion optimization.
\newblock In: \emph{2022 IEEE/RSJ International Conference on Intelligent Robots and Systems (IROS)}. pp. 7672--7679.
\newblock \doi{10.1109/IROS47612.2022.9981264}.

\bibitem[{Vlahov et~al.(2024)Vlahov, Gibson, Fan, Spieler, Agha-mohammadi and Theodorou}]{Vlahov2024ColoredMPPI}
Vlahov B, Gibson J, Fan DD, Spieler P, Agha-mohammadi Aa and Theodorou EA (2024) Low frequency sampling in model predictive path integral control.
\newblock \emph{IEEE Robotics and Automation Letters} 9(5): 4543--4550.
\newblock \doi{10.1109/LRA.2024.3382530}.

\bibitem[{W{\"a}chter and Biegler(2006)}]{Wachter2006IPOPT}
W{\"a}chter A and Biegler LT (2006) On the implementation of an interior-point filter line-search algorithm for large-scale nonlinear programming.
\newblock \emph{Mathematical Programming} 106(1): 25--57.
\newblock \doi{10.1007/s10107-004-0559-y}.

\bibitem[{Wang et~al.(2021)Wang, So, Gibson, Vlahov, Gandhi, Liu and Theodorou}]{Wang2021Tsallis}
Wang Z, So O, Gibson J, Vlahov B, Gandhi MS, Liu GH and Theodorou EA (2021) Variational inference mpc using tsallis divergence.
\newblock In: \emph{Robotics: Science and Systems (RSS)}. p.~73.
\newblock Held virtually, July 12--16, 2021.

\bibitem[{Williams et~al.(2017)Williams, Aldrich and Theodorou}]{williams2017model}
Williams G, Aldrich A and Theodorou EA (2017) Model predictive path integral control: From theory to parallel computation.
\newblock \emph{Journal of Guidance, Control, and Dynamics} 40(2): 344--357.

\bibitem[{Williams et~al.(2016)Williams, Drews, Goldfain, Rehg and Theodorou}]{Williams2016MPPI}
Williams G, Drews P, Goldfain B, Rehg JM and Theodorou EA (2016) Aggressive driving with model predictive path integral control.
\newblock In: \emph{2016 IEEE International Conference on Robotics and Automation (ICRA)}. pp. 1433--1440.
\newblock \doi{10.1109/ICRA.2016.7487277}.

\bibitem[{Williams et~al.(2018{\natexlab{a}})Williams, Drews, Goldfain, Rehg and Theodorou}]{Williams2018MPPI}
Williams G, Drews P, Goldfain B, Rehg JM and Theodorou EA (2018{\natexlab{a}}) Information-theoretic model predictive control: Theory and applications to autonomous driving.
\newblock \emph{IEEE Transactions on Robotics} 34(6): 1603--1622.
\newblock \doi{10.1109/TRO.2018.2865891}.

\bibitem[{Williams et~al.(2018{\natexlab{b}})Williams, Goldfain, Drews, Saigol, Rehg and Theodorou}]{Williams-RSS-18}
Williams G, Goldfain B, Drews P, Saigol K, Rehg J and Theodorou E (2018{\natexlab{b}}) Robust sampling based model predictive control with sparse objective information.
\newblock In: \emph{Proceedings of Robotics: Science and Systems}. Pittsburgh, Pennsylvania.
\newblock \doi{10.15607/RSS.2018.XIV.042}.

\bibitem[{Wills and Heath(2004)}]{WILLS2004BarrierMPC}
Wills AG and Heath WP (2004) Barrier function based model predictive control.
\newblock \emph{Automatica} 40(8): 1415--1422.
\newblock \doi{https://doi.org/10.1016/j.automatica.2004.03.002}.

\bibitem[{Wilson(1963)}]{Wilson1963simplicialSQP}
Wilson RB (1963) \emph{A simplicial algorithm for concave programming}.
\newblock PhD Thesis, Graduate School of Business Administration, Harvard University, Cambridge, MA, USA.

\bibitem[{Xue et~al.(2025)Xue, Pan, Yi, Qu and Shi}]{Haoru2025DialMPC}
Xue H, Pan C, Yi Z, Qu G and Shi G (2025) Full-order sampling-based mpc for torque-level locomotion control via diffusion-style annealing.
\newblock In: \emph{2025 IEEE International Conference on Robotics and Automation (ICRA)}. pp. 4974--4981.
\newblock \doi{10.1109/ICRA55743.2025.11127320}.

\bibitem[{Yi et~al.(2024)Yi, Pan, He, Qu and Shi}]{yi2024covompc}
Yi Z, Pan C, He G, Qu G and Shi G (2024) {CoVO-MPC}: Theoretical analysis of sampling-based mpc and optimal covariance design.
\newblock \emph{arXiv preprint arXiv:2401.07369} .

\bibitem[{Yin et~al.(2025)Yin, Lai, Barcelos, Jacob, Li and Ramos}]{Yin2025SVDiffusion}
Yin Z, Lai T, Barcelos L, Jacob J, Li Y and Ramos F (2025) Diverse motion planning with stein diffusion trajectory inference.
\newblock In: \emph{2025 IEEE International Conference on Robotics and Automation (ICRA)}. pp. 15610--15616.
\newblock \doi{10.1109/ICRA55743.2025.11127953}.

\bibitem[{Zhang et~al.(2025)Zhang, Zhang, Zhu, Yan, Brox and Boedecker}]{Zhang2025constrainedBarrierRL}
Zhang B, Zhang Y, Zhu H, Yan S, Brox T and Boedecker J (2025) Constrained reinforcement learning with smoothed log barrier function.
\newblock \emph{Transactions on Machine Learning Research} .

\bibitem[{Zhuo et~al.(2018)Zhuo, Liu, Shi, Zhu, Chen and Zhang}]{zhuo2018messageSV}
Zhuo J, Liu C, Shi J, Zhu J, Chen N and Zhang B (2018) Message passing stein variational gradient descent.
\newblock In: Dy J and Krause A (eds.) \emph{Proceedings of the 35th International Conference on Machine Learning}, \emph{Proceedings of Machine Learning Research}, volume~80. PMLR, pp. 6018--6027.

\bibitem[{Ziebart(2010)}]{ziebart2010modeling}
Ziebart BD (2010) \emph{Modeling Purposeful Adaptive Behavior with the Principle of Maximum Causal Entropy}.
\newblock PhD Thesis, Carnegie Mellon University, Pittsburgh, PA, USA.
\newblock CMU-CS-10-110.

\bibitem[{Zucker et~al.(2013)Zucker, Ratliff, Dragan, Pivtoraiko, Klingensmith, Dellin, Bagnell and Srinivasa}]{Zucker2013HamiltonianChomp}
Zucker M, Ratliff N, Dragan AD, Pivtoraiko M, Klingensmith M, Dellin CM, Bagnell JA and Srinivasa SS (2013) Chomp: Covariant hamiltonian optimization for motion planning.
\newblock \emph{The International Journal of Robotics Research} 32(9-10): 1164--1193.
\newblock \doi{10.1177/0278364913488805}.

\end{thebibliography}

\end{document}